\documentclass[journal]{IEEEtran}
\ifCLASSINFOpdf
\else
\fi
\usepackage{amsmath,amsfonts,amssymb,mathtools}
\usepackage{multirow}
\usepackage[dvipsnames]{xcolor}
\usepackage{graphicx}
\usepackage{hyperref}
\usepackage{enumitem}
\setlist{nolistsep}
\usepackage[caption=false,font=footnotesize]{subfig}
\usepackage{algpseudocode}

\begin{document}
%
\title{Study of Subjective and Objective Quality Assessment of Mobile Cloud Gaming Videos}
%
%
%
\author{Avinab Saha, Yu-Chih Chen, Chase Davis, Bo Qiu, Xiaoming Wang, Rahul Gowda, Ioannis Katsavounidis, Alan C. Bovik,~\IEEEmembership{Fellow,~IEEE}
\thanks{This work was supported by Meta Platforms, Inc. A.C. Bovik was supported in part by the National Science Foundation AI Institute for Foundations of Machine Learning (IFML) under Grant 2019844. (Corresponding author: Avinab Saha.)}
\thanks{This work involved human subjects or animals in its research. Approval of all ethical and experimental procedures and protocols was granted by the Institutional Review Board (IRB), University of Texas, Austin, under FWA No. 00002030 and Protocol No. 2007-11-0066.}
\thanks{Avinab Saha, Yu-Chih Chen, Alan C Bovik are with the
Department of Electrical and Computer Engineering, The University of Texas at Austin, TX 78712 USA (e-mail: avinab.saha@utexas.edu, berriechen@utexas.edu, bovik@ece.utexas.edu). Chase Davis, Bo Qiu, Xiaoming Wang, Rahul Gowda, Ioannis Katsavounidis are with Meta Platforms Inc., Menlo Park, CA 94025, USA
(e-mail: chased@fb.com, qiub@fb.com, xmwang@fb.com, rahulgowda@fb.com, ikatsavounidis@fb.com).}
}
%
%

\markboth{IEEE TRANSACTIONS ON IMAGE PROCESSING, 2023 (Pre-print)}%
{Shell \MakeLowercase{\textit{et al.}}: Bare Demo of IEEEtran.cls for IEEE Journals}
%



\maketitle

\begin{abstract}
We present the outcomes of a recent large-scale subjective study of Mobile Cloud Gaming Video Quality Assessment (MCG-VQA) on a diverse set of gaming videos. Rapid advancements in cloud services, faster video encoding technologies, and increased access to high-speed, low-latency wireless internet have all contributed to the exponential growth of the Mobile Cloud Gaming industry. Consequently, the development of methods to assess the quality of real-time video feeds to end-users of cloud gaming platforms has become increasingly important. However, due to the lack of a large-scale public Mobile Cloud Gaming Video dataset containing a diverse set of distorted videos with corresponding subjective scores, there has been limited work on the development of MCG-VQA models. Towards accelerating progress towards these goals, we created a new dataset, named the LIVE-Meta Mobile Cloud Gaming (LIVE-Meta-MCG) video quality database, composed of 600 landscape and portrait gaming videos, on which we collected 14,400 subjective quality ratings from an in-lab subjective study. Additionally, to demonstrate the usefulness of the new resource, we benchmarked multiple state-of-the-art VQA algorithms on the database. The new database will be made publicly available on our website: \url{https://live.ece.utexas.edu/research/LIVE-Meta-Mobile-Cloud-Gaming/index.html}
\end{abstract}


%
\IEEEpeerreviewmaketitle

\section{Introduction}
\begin{figure}[]
  \centering
  \centerline{\includegraphics[width=6.0cm]{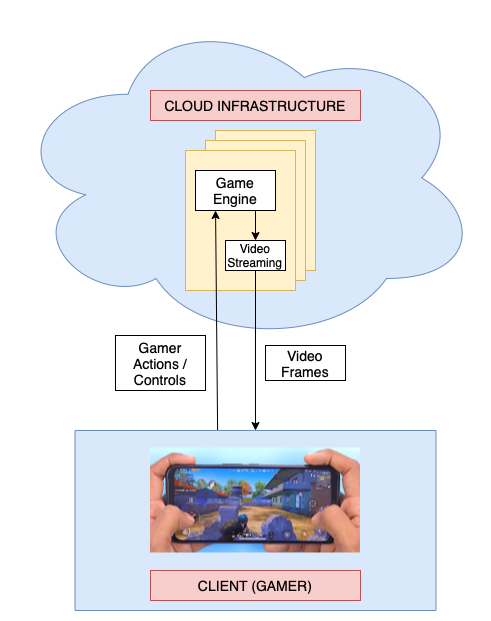}}
  \caption{Exemplar Mobile Cloud Gaming system. Video games scenes are rendered in the Cloud servers of service providers, then the gaming video frames are sent over the Internet to end-users' Mobile devices. The game players' interactions are sent back to the Cloud server over the same network.} \label{fig:mcg}
\end{figure}

%
%
%
%
\IEEEPARstart{T}{he} last decade has witnessed the growth of cloud gaming services as an emergent technology in the digital gaming industry, and many major technology companies such as Meta, Google, Apple, NVIDIA and Microsoft have aggressively invested in building cloud gaming infrastructure. According to a survey by Allied Market Research \cite{Allied}, the cloud gaming industry is projected to grow at a compounded annual growth rate of $57.2\%$ from 2021 to 2030. This astronomical growth may be attributed to multiple factors. Cloud gaming services are a cost-effective alternative to traditional physical gaming consoles and PC (personal computer) based digital video games, a critical factor contributing to their rapid growth. Cloud gaming subscribers are able to access large and diverse libraries of games playable on any device anywhere without downloading or installing them. Cloud gaming aims to provide high-quality gaming experiences to users by executing complex game software on powerful cloud gaming servers, and streaming the computed game scenes over the internet in real-time, as depicted in Fig \ref{fig:mcg}. Gamers use lightweight software that can be executed on any device to view real-time video game streams while interacting with the games. Cloud gaming services also facilitate rapid video game development processes by eliminating support requirements on multiple user systems, leading to lower overall production costs. This alleviates the need to upgrade consoles and PCs to maintain the gaming experiences of the end-users, as newer and more complex games are made available. Other notable factors contributing to the growth of cloud gaming services include the development of hardware-accelerated video compression methods, access to inexpensive high-speed, lower latency wireless internet services facilitated by the introduction of global 5G services, and the availability of more efficient and affordable cloud platform infrastructures like AWS, Google Cloud, and Microsoft Azure. Another significant contributor to the acceleration of the cloud gaming market since 2019 has been COVID-19 induced restrictions and lockdowns. Indeed, the amount of time spent playing video games increased by more than $71\%$ during the COVID-19 lockdown, \textcolor{black}{as reported in \cite{Allied}.}\\ 
\indent Recent trends suggest that smartphones have begun to  dominate the global cloud gaming industry, and this uptrend is expected to continue. Mobile Cloud Gaming differs from generic Cloud Gaming in various important ways. First, Mobile Cloud Gaming services generally render video game scenes at 720p resolution and 30 frames per second (fps) to accommodate the current gamut of mobile devices, while helping to stabilize delivery and ensuring smoother connections. By comparison, non-mobile Cloud Gaming applications, which are typically played on PCs and televisions, are usually rendered at 1080p/4K resolution and 30-120 fps. Second, Mobile Gaming experiences support gameplay in both portrait and landscape orientations on mobile devices, unlike PCs and television games, which are only playable in landscape mode. Third, Mobile Cloud Gaming services allow users to play over the wireless internet, and must contend with variable internet connections and transmission speeds, unlike cloud gaming services played on PCs and televisions having stable, high-bandwidth wired internet access. This raises significant technical challenges that must be met to deliver acceptable levels of perceived game video quality. \\
\indent In a cloud gaming setup, video artifacts can severely impair the perceptual quality of delivered gaming videos. Because of this, there is heightened interest in developing perceptual Video Quality Assessment (VQA) models for gaming videos. However, there have been limited advancements in this direction for two reasons. First, VQA algorithms that have been trained on generic VQA databases generally do not perform well on content-specific gaming videos, which exhibit different appearances and statistical properties than naturalistic camera-captured videos. \\
\indent Second, building those models inevitably requires the construction of psychometric VQA databases containing large numbers of representative gaming videos that have been labeled with human-annotated scores. Unfortunately, there are very few VQA databases dedicated to Cloud Gaming VQA research, and none are public databases focused on MCG-VQA. Towards advancing progress in this domain, we created a new resource that we call the LIVE-Meta Mobile Cloud Gaming (LIVE-Meta MCG) database, composed of 600 landscape and portrait gaming videos, and targeted explicitly towards mobile cloud gaming. The new database contains 600 videos drawn from 30 source sequences obtained from 16 different games, impaired by varying degrees of video compression and resizing distortions. We then conducted a sizeable human subjective study on these videos. To demonstrate the usefulness of the new database, we also performed a rigorous evaluation of current state-of-the-art VQA models on it, and compared their performance. \\ 
\indent The remaining parts of the paper are organized as follows. Section \ref{sec:relwork} presents prior work relevant to our mobile cloud gaming video quality. In Section \ref{sec:relevance}, we discuss the relevance of the new mobile gaming VQA dataset and highlight the novelty and significance of our work. Section \ref{sec:details} explains the data acquisition process and the design of the human study protocol. Section \ref{sec:bench} compares the performances of various state-of-the-art (SOTA) No-Reference VQA models on the LIVE-Meta Mobile Cloud Gaming (LIVE-Meta MCG) database. \textcolor{black}{Section \ref{sec:proxymos} studies the performances of popular Full Reference VQA algorithms originally developed for natural videos, from the perspective of their possibly being used as proxy-MOS or pre-training targets in the development of deep-learning based NR-VQA models for Mobile Cloud Gaming.} We conclude in Section \ref{sec:conclusion} by summarizing the paper and discussing possible directions of future work.

\begin{table*}[]
 
\centering
\caption{A Summary of Existing Gaming VQA Databases and the new LIVE-META Mobile Cloud Gaming Database}
\resizebox{\textwidth}{!}{%
\begin{tabular}{|c|c|c|c|c|c|c|c|c|c|c|c|}
\hline
Database &
  \# Videos &
  \begin{tabular}[c]{@{}c@{}}\# Source \\ Sequences\end{tabular} &
  \begin{tabular}[c]{@{}c@{}} Pristine Source \\ Sequences\end{tabular} &
  \begin{tabular}[c]{@{}c@{}}\# Ratings\\ per Video\end{tabular} &
  Public &
  Resolution &
  Distortion Type &
  Duration &
  Display Device &
  Display Orientation &
  Study Type \\ \hline
GamingVideoSET &
  90 &
  6 &
  Yes &
  25 &
  Yes &
  \begin{tabular}[c]{@{}c@{}}480p, 720p, \\ 1080p\end{tabular} &
  H.264 &
  30 sec &
  24'' Monitor &
  Landscape &
  Laboratory \\ \hline
KUGVD &
  90 &
  6 &
  Yes &
  17 &
  Yes &
  \begin{tabular}[c]{@{}c@{}}480p, 720p, \\ 1080p\end{tabular} &
  H.264 &
  30 sec &
  55'' Monitor &
  Landscape &
  Laboratory \\ \hline
CGVDS &
  \begin{tabular}[c]{@{}c@{}}360 +\\ anchor stimuli\end{tabular} &
  15 &
  Yes &
  Unavailable &
  Yes &
  \begin{tabular}[c]{@{}c@{}}480p, 720p, \\ 1080p\end{tabular} &
  H.264 NVENC &
  30 sec &
  24'' Monitor &
  Landscape &
  Laboratory \\ \hline
TGV &
  1293 &
  150 &
  No &
  Unavailable &
  No &
  \begin{tabular}[c]{@{}c@{}}480p, 720p, \\ 1080p\end{tabular} &
  \begin{tabular}[c]{@{}c@{}}H.264, H.265,\\  Tencent codec\end{tabular} &
  5 sec &
  \begin{tabular}[c]{@{}c@{}}Unknown\\ Mobile Device\end{tabular} &
  Landscape &
  Laboratory \\ \hline
\begin{tabular}[c]{@{}c@{}}LIVE-YT\\ -Gaming\end{tabular} &
  600 &
  600 &
  No &
  30 &
  Yes &
  \begin{tabular}[c]{@{}c@{}}360p, 480p, \\ 720p, 1080p\end{tabular} &
  UGC distortions &
  8-9 sec &
  Multiple Devices &
  Landscape &
  Online \\ \hline
\begin{tabular}[c]{@{}c@{}}LIVE-Meta \\ Mobile Cloud Gaming\end{tabular} &
  600 &
  30 &
  Yes &
  24 &
  Yes &
  \begin{tabular}[c]{@{}c@{}}360p, 480p, \\ 540p, 720p\end{tabular} &
  H.264 NVENC &
  20 sec &
  Google Pixel 5 &
  \begin{tabular}[c]{@{}c@{}}Landscape, \\ Portrait\end{tabular} &
  Laboratory \\ \hline
\end{tabular}}
\label{tab:dataset-table}
\end{table*}

\section{Related Work}
\label{sec:relwork}
Video Quality Assessment research over the last decade has been elevated by the availability of large, comprehensive databases containing videos labeled by subjective quality scores obtained by conducting either laboratory or online studies. Given the explosive growth of the digital gaming industry over the last few years, there is an urgent need to develop gaming-specific VQA algorithms that can be used to monitor and control the quality of video gaming streams transmitted throughout the global internet, towards ensuring that millions of users will experience holistic, high-quality gameplay. Consequently, VQA researchers have begun to develop subjective VQA databases that are focused on gaming videos, as tools for the development of Gaming VQA algorithms. Early work has produced the GamingVideoSET \cite{Barman2018GamingVideoSETAD} and the Kingston University Gaming Video Dataset (KUGVD) \cite{8727887}. However, these databases are quite limited in the number of videos having associated subjective quality ratings and in the variety of source content. Both databases \cite{Barman2018GamingVideoSETAD}, \cite{8727887} were built on only six source sequences, each used to create 15 resolution-bitrate distortion pairs, yielding a total of only 90 videos rated by human subjects. These data limitations are a bottleneck to the development of reliable and flexible VQA models. Towards bridging this gap, a more extensive Cloud Gaming Video Dataset (CGVDS) dataset was introduced in \cite{10.1145/3339825.3391872}. This dataset includes subjective quality ratings on more than 360 gaming videos obtained from 15 source sequences, collected in a laboratory human study. However, all of the videos in the CGVDS dataset were rendered in landscape mode; hence training a VQA model on them could result in unreliable performance on portrait gaming videos. The other two datasets in the Gaming VQA domain are the Tencent Gaming Video (TGV) dataset \cite{Wen2021SubjectiveAO} and the LIVE-YT-Gaming dataset \cite{yu2022subjective}. The TGV dataset contains 1293 landscape gaming videos drawn from 150 source sequences. However, this dataset is not available in the public domain. The LIVE-YT-Gaming video dataset contains 600 original user-generated content (UGC) gaming videos harvested from the internet. Since these UGC videos were obtained by downloading after-the-fact user-generated gameplay videos from a variety of websites, they are not good candidates for training Cloud Gaming VQA algorithms. Instead, it is desirable to be able to train MCG-VQA models on multiple distorted versions of high-quality source videos, so that they can be used to choose optimal streaming settings for given network conditions, to deliver the best possible viewing experiences to gaming end-users. \\ 
\indent Other than the LIVE-YT-Gaming dataset, the source videos in gaming databases are of very high pristine quality. They have generally been played using powerful hardware devices, under high-quality game settings and recorded with professional-grade software. The source sequences are then typically processed with resizing and video compression operations to generate a corpus of the distorted videos. We summarize the characteristics of existing gaming VQA databases along with the new LIVE-Meta Mobile Cloud Gaming video quality database in Table \ref{tab:dataset-table}. \\
\indent Along with the development of Gaming Video Quality databases, several methods have been proposed for Gaming VQA tasks. NR-GVQM \cite{Zadtootaghaj2018NRGVQMAN} trains an SVR model to evaluate the quality of gaming content videos by extracting $9$ frame-level features, using VMAF \cite{vmaf} scores as proxy ground-truth labels. In \cite{inproceedingsnofu}, the authors introduced ``nofu", a lightweight model that uses only a center crop of each frame, to speed up the computation of $12$ frame-based features, followed by model training and temporal pooling. Recent gaming VQA models based on deep learning include NDNet-Gaming \cite{NDNetgaming}, DEMI \cite{9287080}, and GAMIVAL \cite{gamival}. Both NDNet-Gaming and DEMI use Densenet-121 \cite{DBLP:journals/corr/HuangLW16a} deep learning backbones. Because of the limited amount of subjective scores available to train deep-learning backbones, the Densenet-121 in NDNet-Gaming is pre-trained on VMAF scores that serve as proxy ground truth labels, then fine-tuned using MOS scores. A temporal pooling algorithm is finally used to compute video quality predictions. DEMI uses a CNN architecture similar to NDNet-Gaming, while addressing artifacts that include blockiness, blur, and jerkiness. GAMIVAL combines features computed under distorted natural scene statistics model with features computed by the pre-trained CNN backbone used in NDNet-Gaming, to predict gaming video quality. The ITU-T G.1072 \cite{recogaming} planning model determines gaming video quality based on using objective (non-perceptual) video parameters such as bitrate, framerate, encoding resolution, game complexity, and network parameters.

\begin{figure*}[htbp]
    \centering
    \subfloat[Asphalt]{{\includegraphics[width=0.23\textwidth]{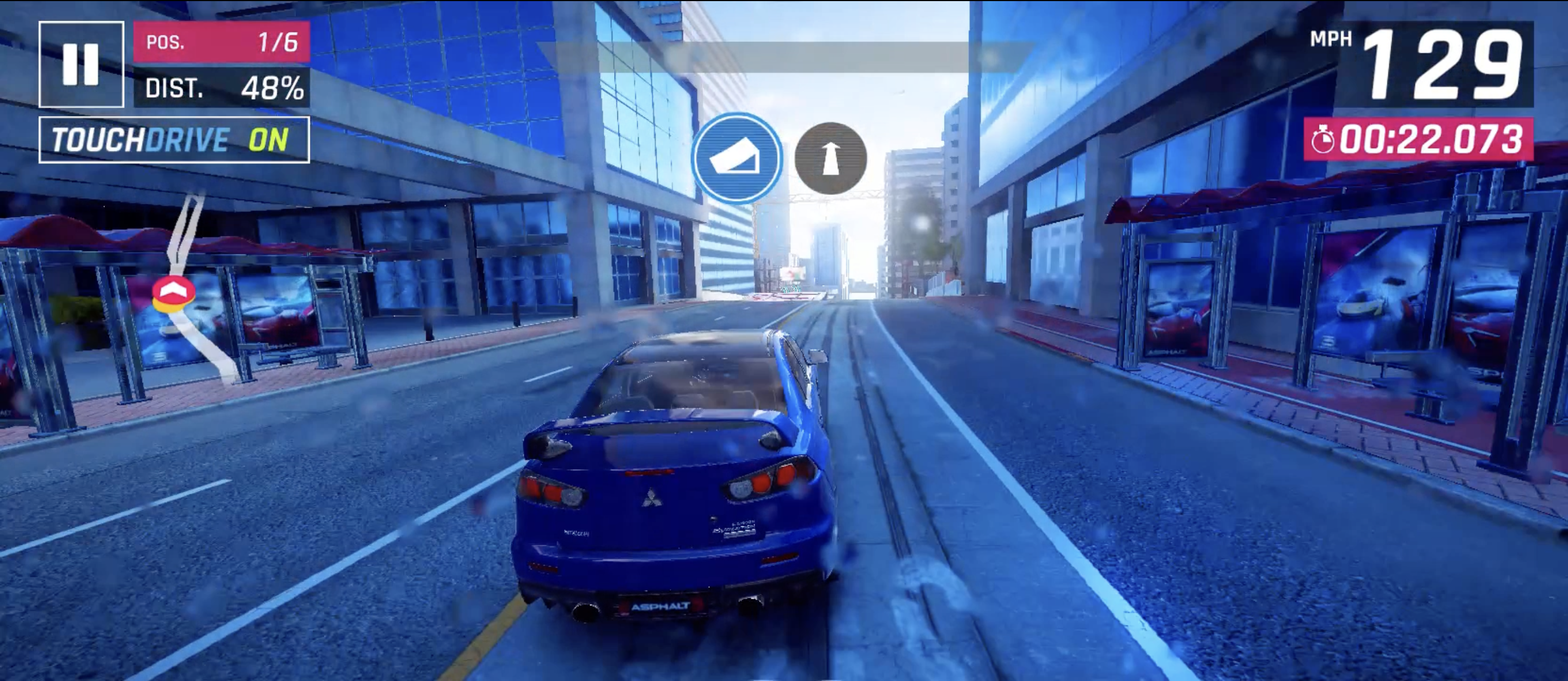}}}
    \hfill
    \subfloat[Design Island]{{\includegraphics[width=0.23\textwidth]{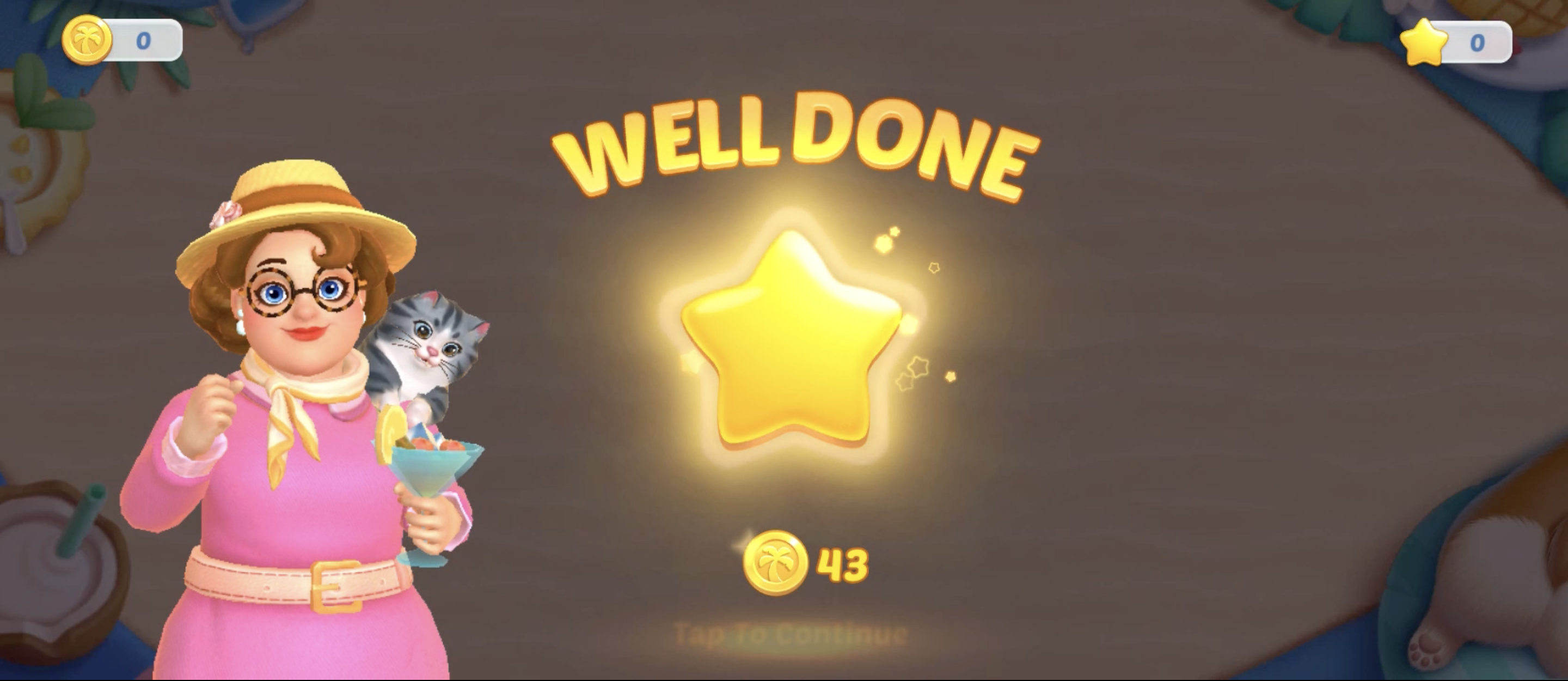}}}
    \hfill
    \subfloat[Dragon Mania Legends]{{\includegraphics[width=0.23\textwidth]{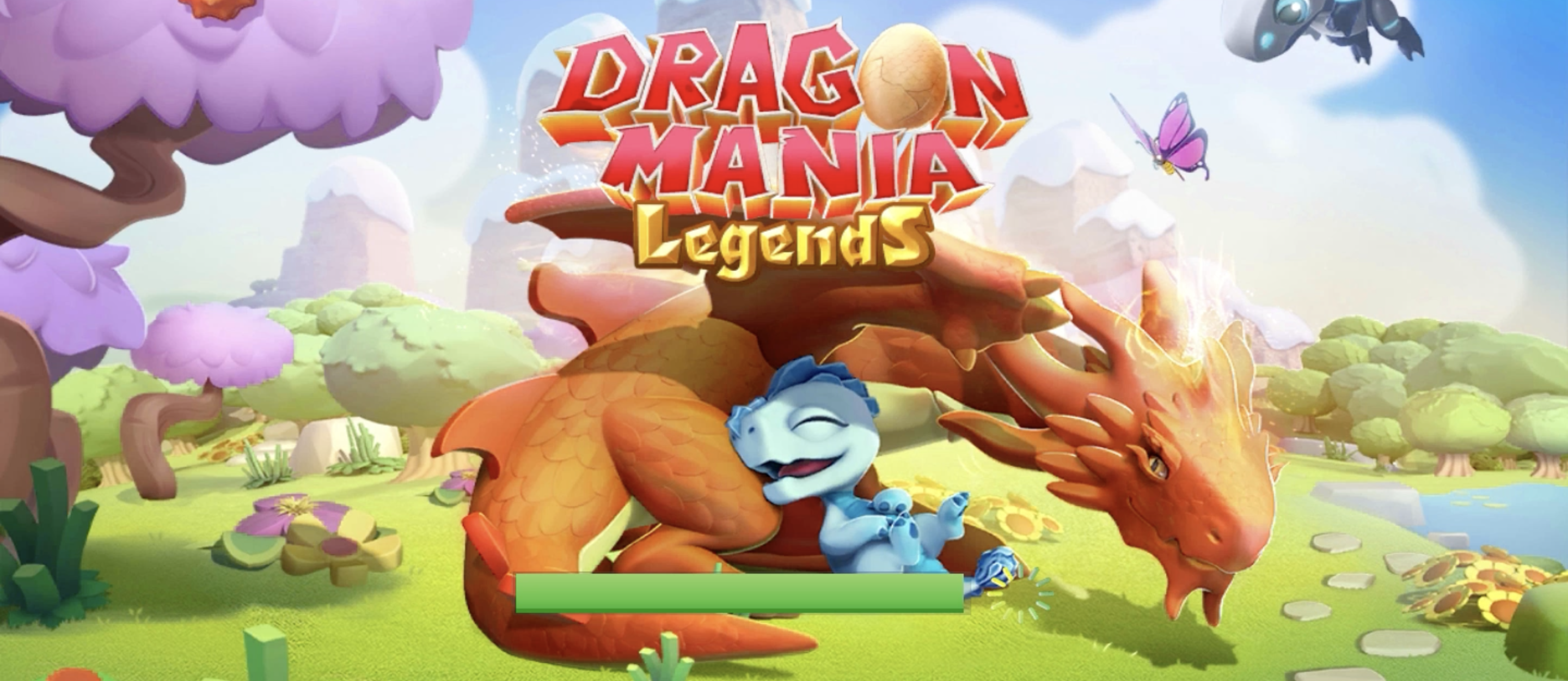}}}
    \hfill
    \subfloat[Hungrydragon]{{\includegraphics[width=0.23\textwidth]{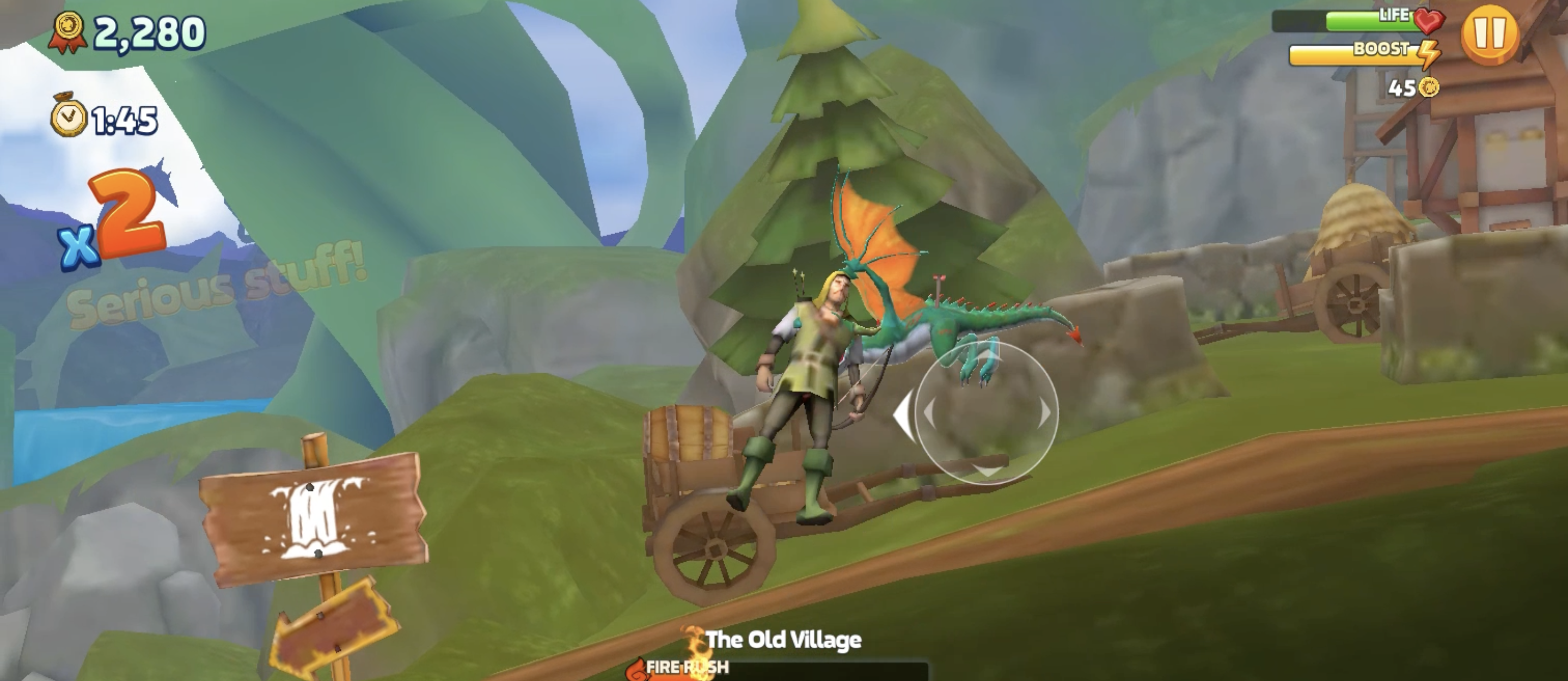}}}

    \subfloat[Mobile Legends Adventure]{{\includegraphics[width=0.23\textwidth]{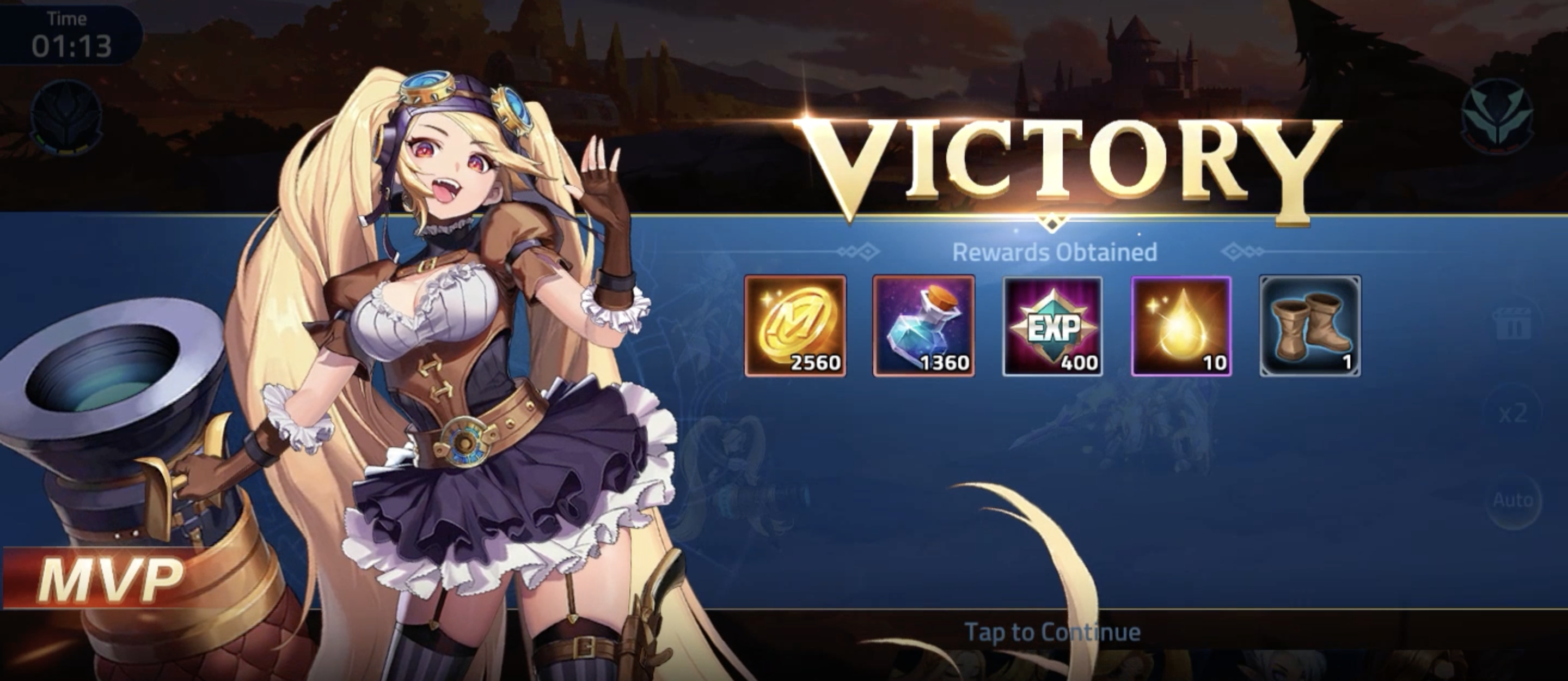}}}
    \hfill
    \subfloat[Mystery Manor]{{\includegraphics[width=0.23\textwidth]{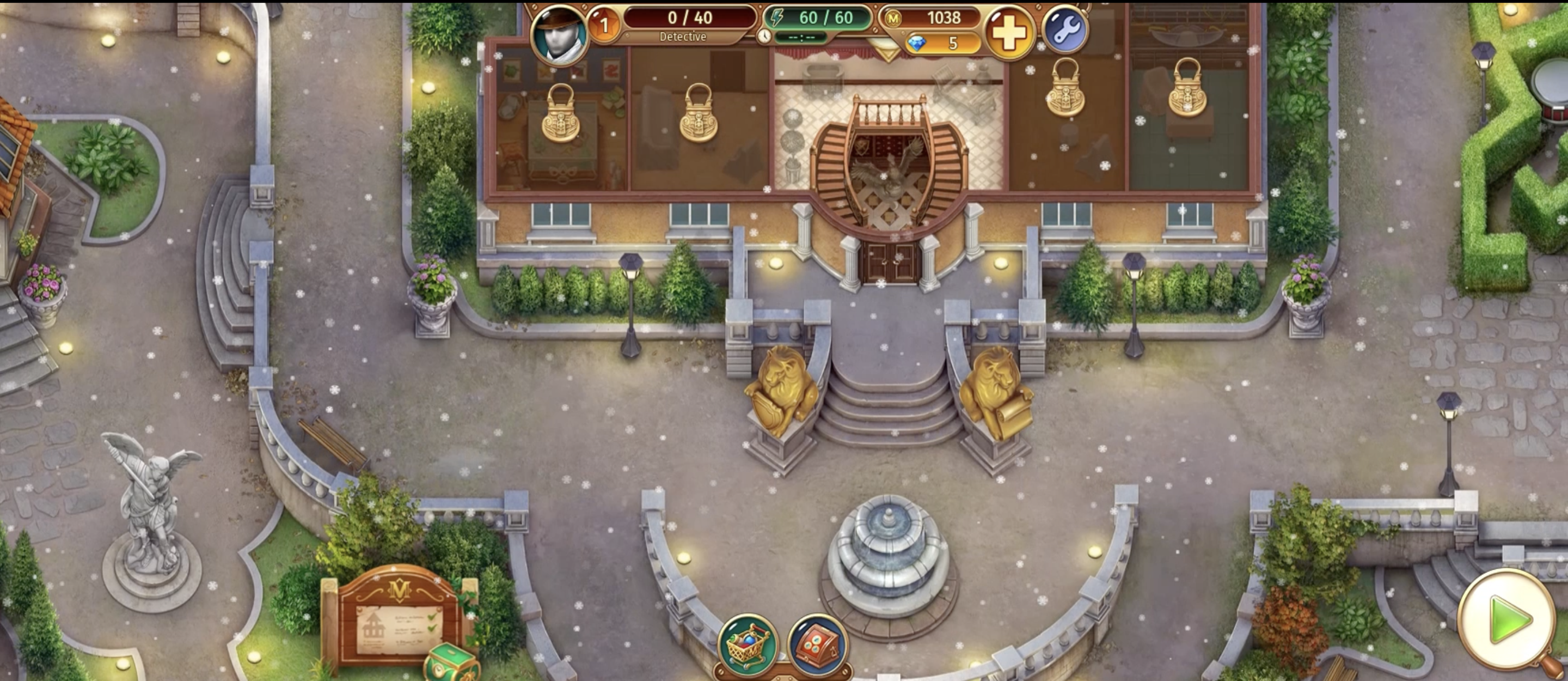}}}
    \hfill
    \subfloat[Plants vs Zombies]{{\includegraphics[width=0.23\textwidth]{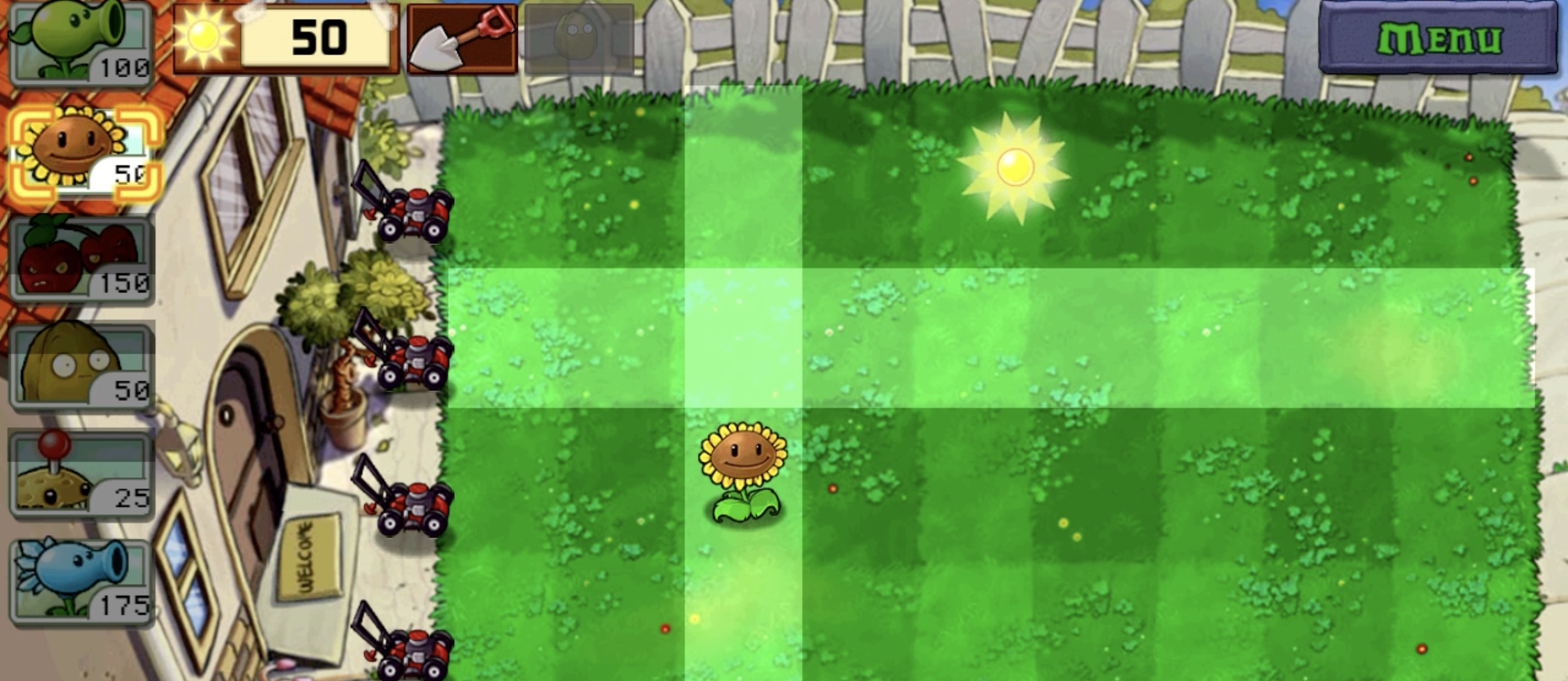}}}
    \hfill
    \subfloat[State of Survival]{{\includegraphics[width=0.23\textwidth]{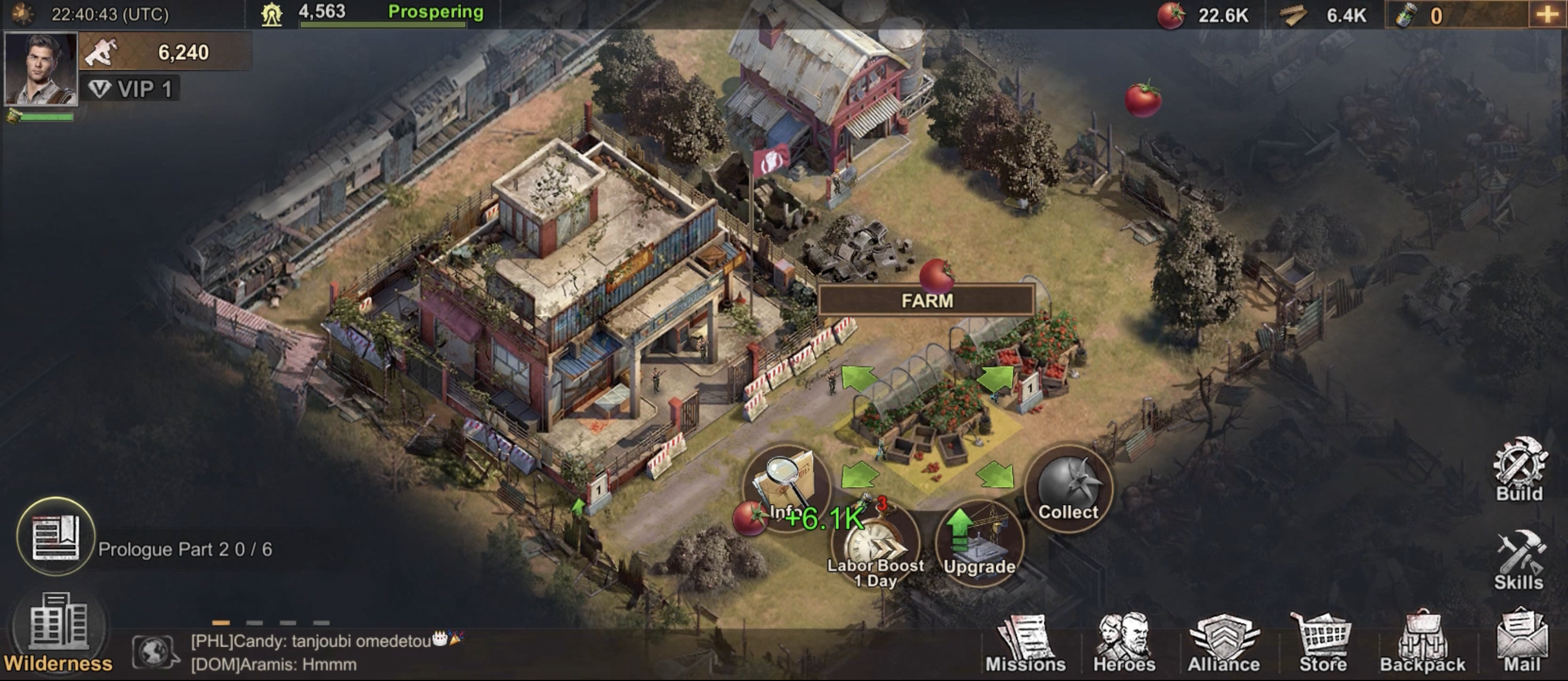}}}    
    \captionsetup{justification=centering}
    \caption{Sample frames of landscape gaming videos in the LIVE-Meta Mobile Cloud Gaming Database.}%
    \label{fig:land}
\end{figure*}

\begin{figure*}[htbp]
    \centering
    \subfloat[Bejewelled]{{\includegraphics[width=0.15\textwidth]{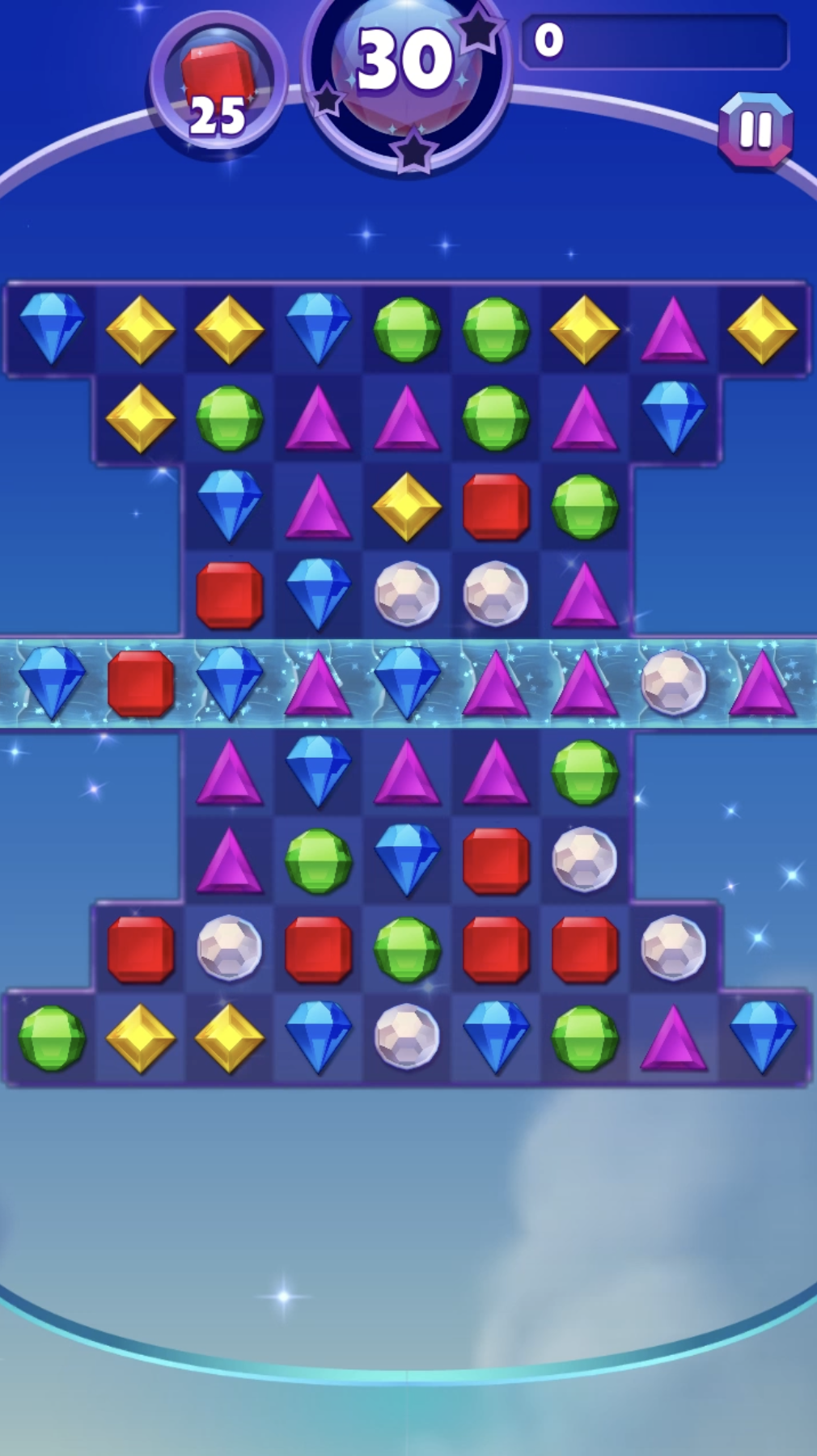}}}
    \hfill
    \subfloat[Bowling Club]{{\includegraphics[width=0.15\textwidth]{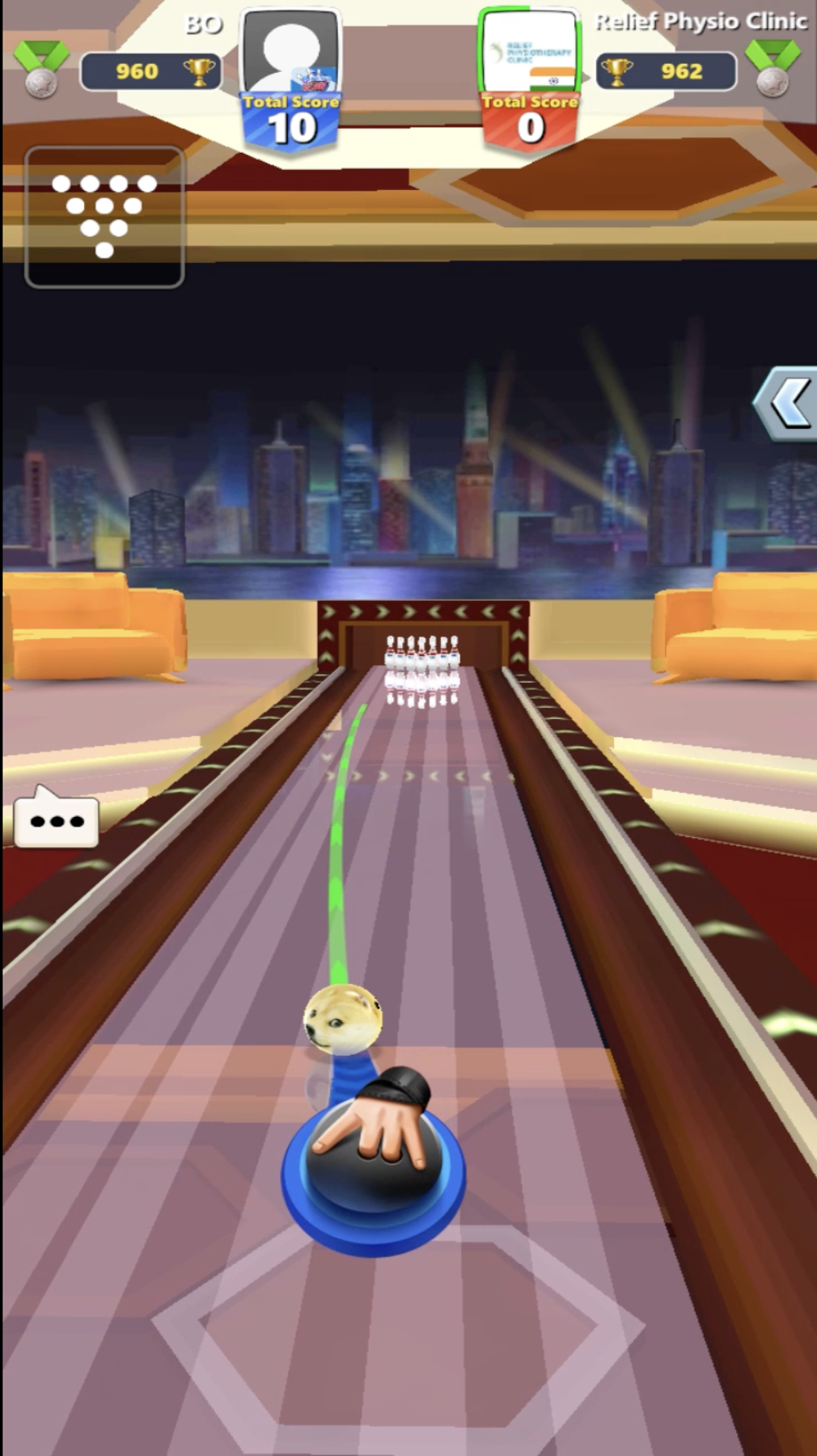}}}
    \hfill
    \subfloat[Dirtbike]{{\includegraphics[width=0.15\textwidth]{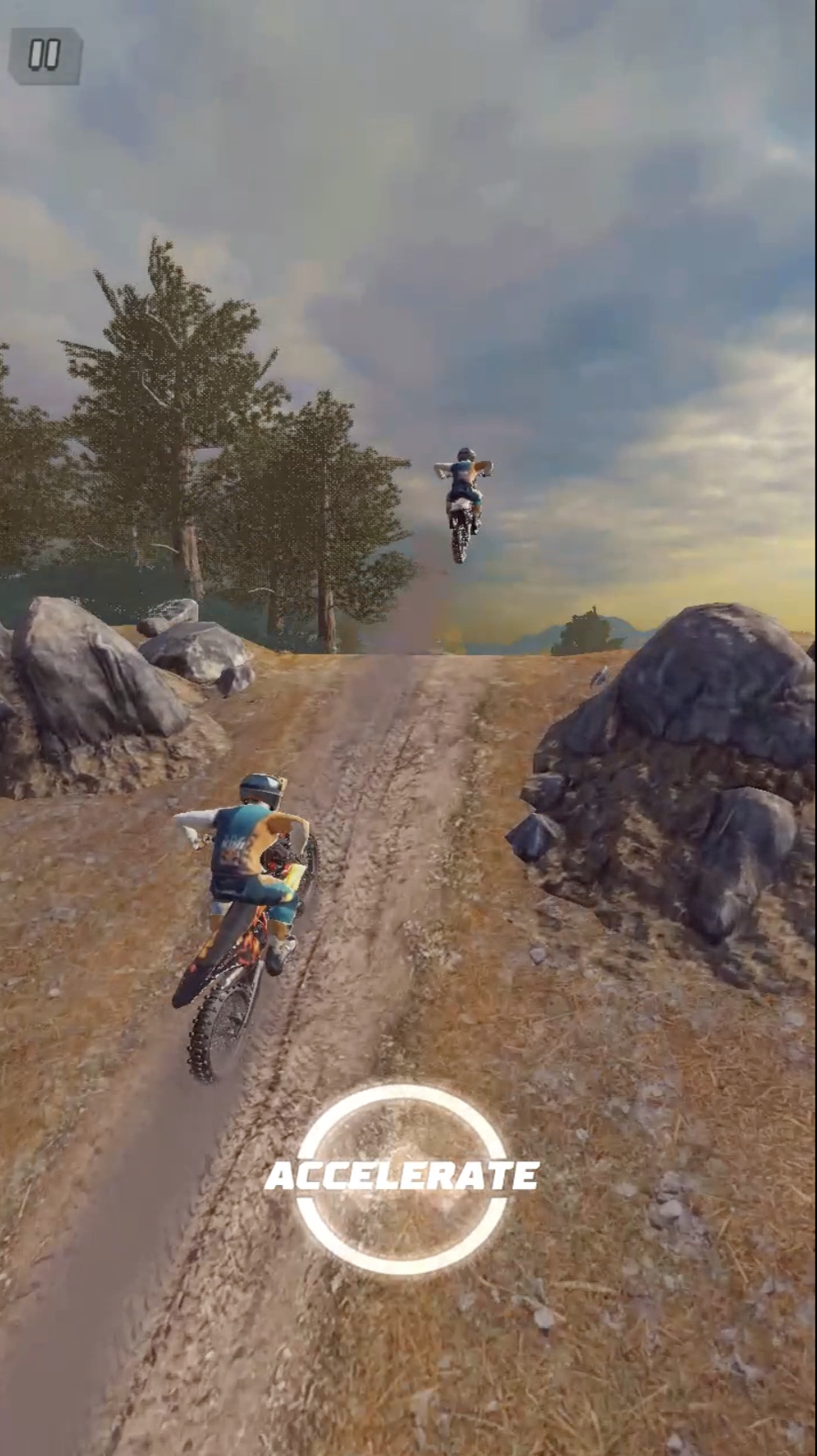}}}
    \hfill
    \subfloat[PGA Golf Tour]{{\includegraphics[width=0.15\textwidth]{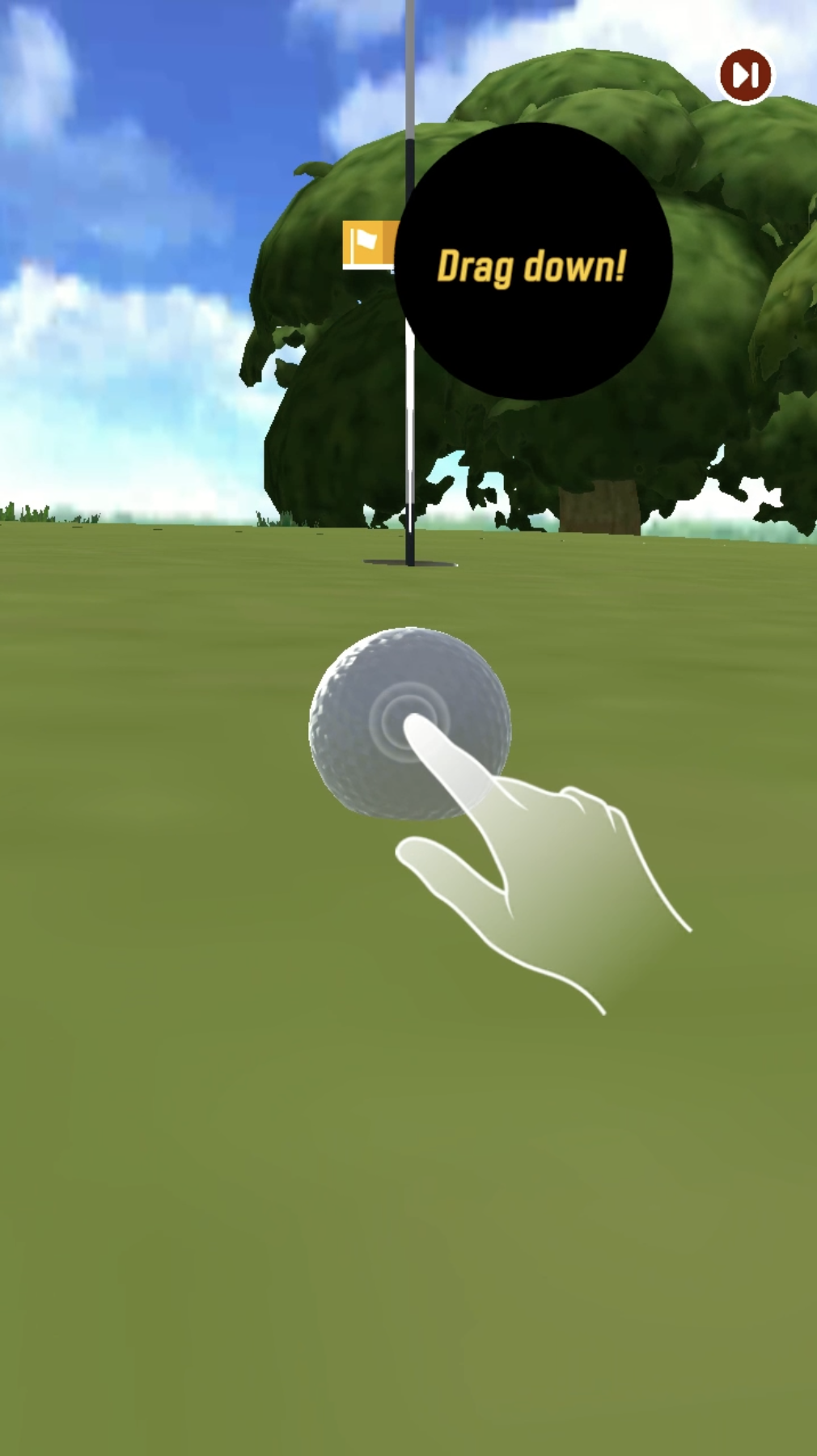}}}
    \hfill
    \subfloat[Sonic]{{\includegraphics[width=0.15\textwidth]{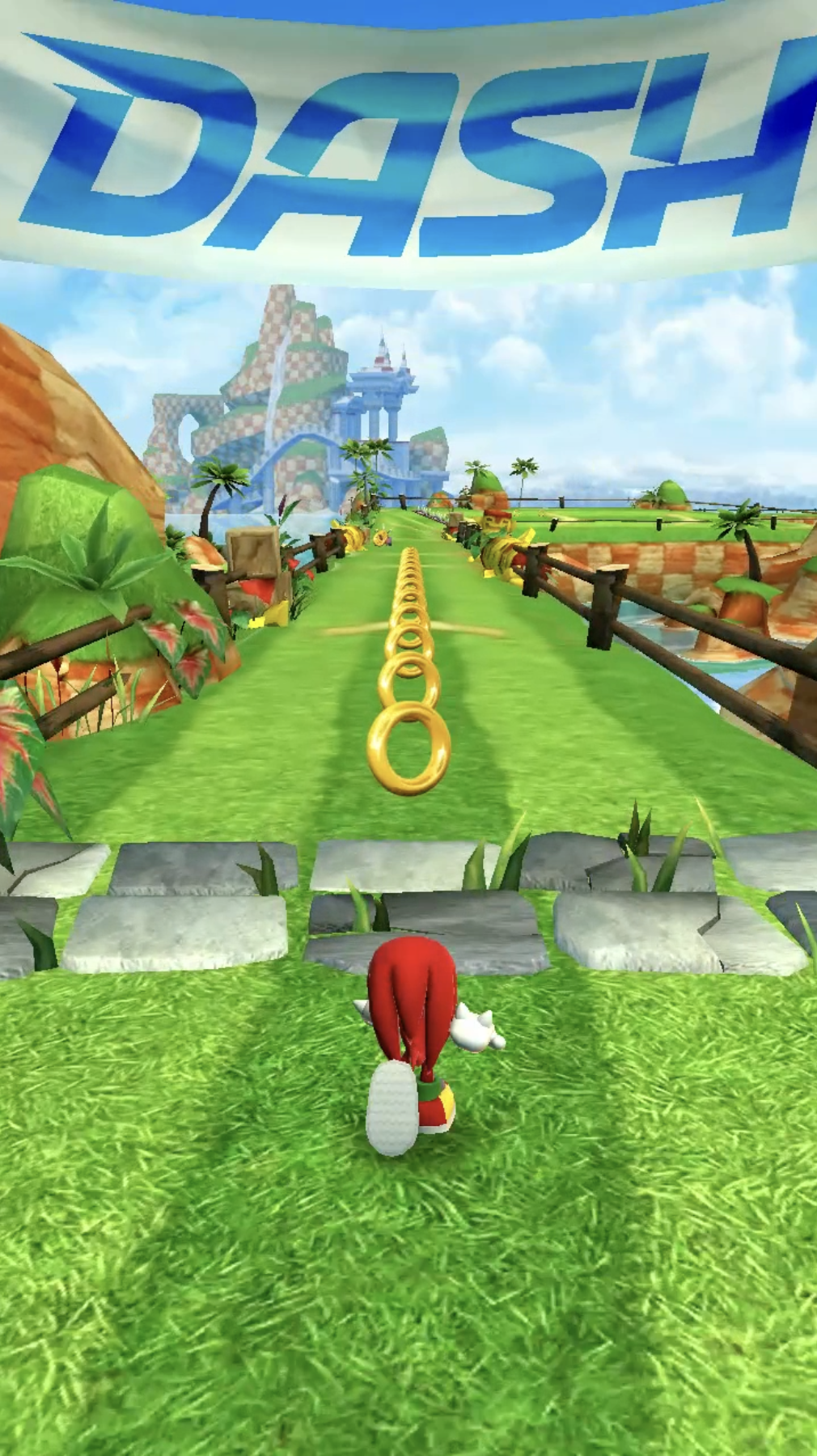}}}
    \hfill
    \subfloat[WWE]{{\includegraphics[width=0.15\textwidth]{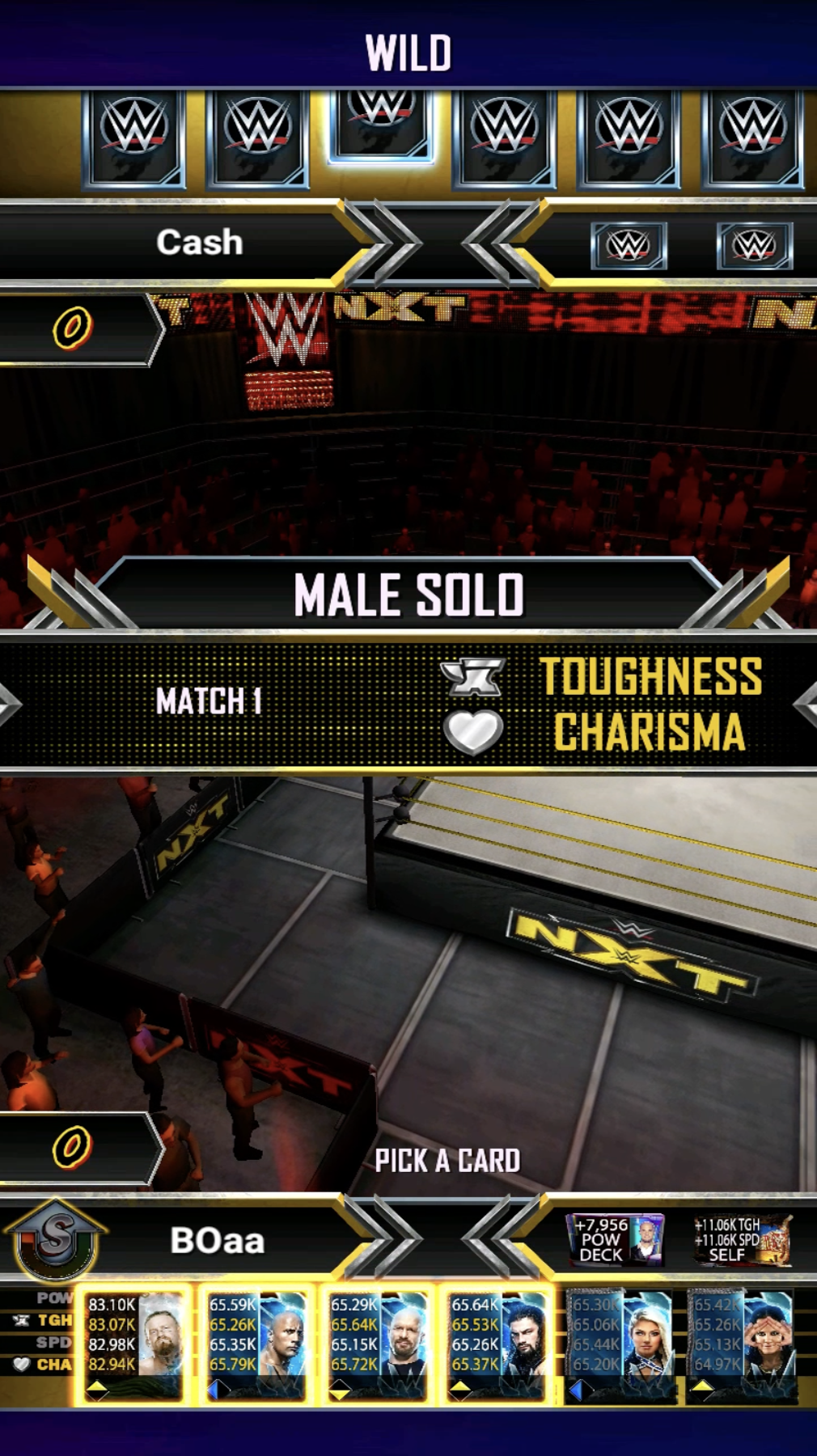}}}
    \caption{Sample frames of portrait gaming videos in the LIVE-Meta Mobile Cloud Gaming Database.}%
    \label{fig:portrait}
\end{figure*}

\section{Relevance and Novelty of LIVE-Meta Mobile Cloud Gaming Database}
\label{sec:relevance}
The new psychometric data resource that we describe here has multiple unique attributes that address most of the shortcomings of existing gaming databases. \\
\indent First, it includes the largest number of unique source sequences of any non-UGC public gaming VQA database. While the LIVE-YT-Gaming dataset does contain more unique contents, it is directed towards a different problem - VQA of low-quality, user-generated, user-recorded gaming videos. The TGV dataset \cite{Wen2021SubjectiveAO} also has more source sequences, but none of the data is publicly available, making it impossible to independently verify the integrity and modeling value of the videos. Moreover, the video durations are only 5 seconds, heightening the possibility that the subjective quality ratings on the gaming videos, which often contain much longer gameplay scenes, might be less reliable, as explained in \cite{8463417}. The videos that comprise the LIVE-Meta MCG dataset include a wide range of gameplay and game-lobby video shots. The level of activity in the videos include low, medium, and high motion scenes, a diversity not present in other public gaming databases. \\ 
\indent Second, the new data resource can be used to design reliable and robust VQA algorithms, suitable for analyzing high-quality gaming videos subjected to wide ranges and combinations of resizing and compression distortions characteristic of modern streaming workflows. A salient feature of the dataset is that we include videos for all possible resolution-bitrate pairs that are currently relevant to mobile cloud gaming. We believe that VQA tools designed on this data will enable better decision making when selecting streaming settings to deliver perceptually optimized viewing experiences. \\
\indent Third, not only does the corpus of videos that we assembled target the mobile device scenario, we also conducted the human study using a modern mobile device, unlike any other gaming VQA resource. \\
\indent Lastly, another unique and differentiating aspect of the new LIVE-Meta MCG is that it includes gaming videos presented in both portrait and landscape orientations. A summary of unique attributes of the new dataset with comparisons against existing gaming VQA datasets is given in Table \ref{tab:dataset-table}.

\section{Details of subjective study}
\label{sec:details}
The LIVE-Meta MCG Database contains 600 video sequences generated from 30 high-quality (pristine) reference source videos by compressing each video using 20 different resolution-bitrate protocols. These videos served as the stimuli that were quality-rated by the humans who participated in our laboratory subjective experiments. Sample frames of landscape and portrait mode gaming video contents in the database are shown in Figs. \ref{fig:land} and \ref{fig:portrait}, respectively.

\begin{figure*}[htbp]
    \centering
    \subfloat[Contrast vs Brightness]{{\includegraphics[width=0.32\textwidth]{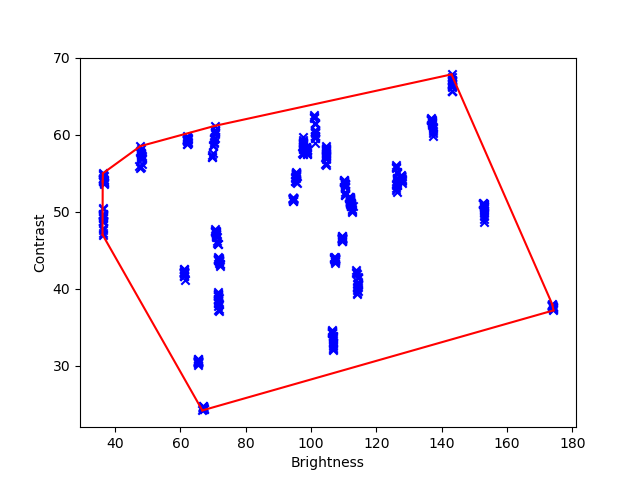}}}
    \hfill
    \subfloat[Sharpness vs Colourfulness]{{\includegraphics[width=0.32\textwidth]{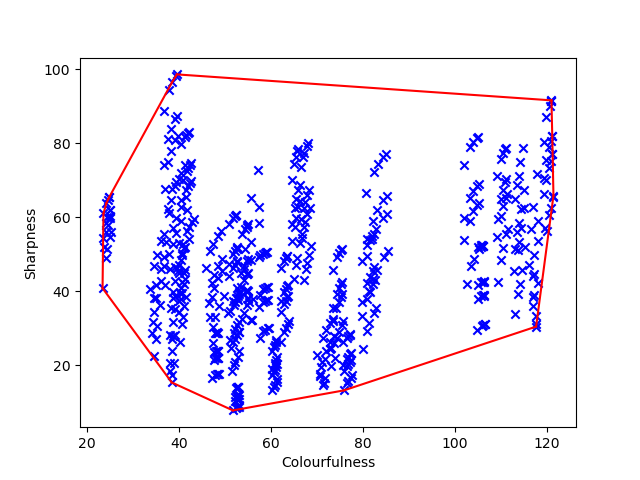}}}
    \hfill
    \subfloat[Temporal Information vs Spatial Information]{{\includegraphics[width=0.32\textwidth]{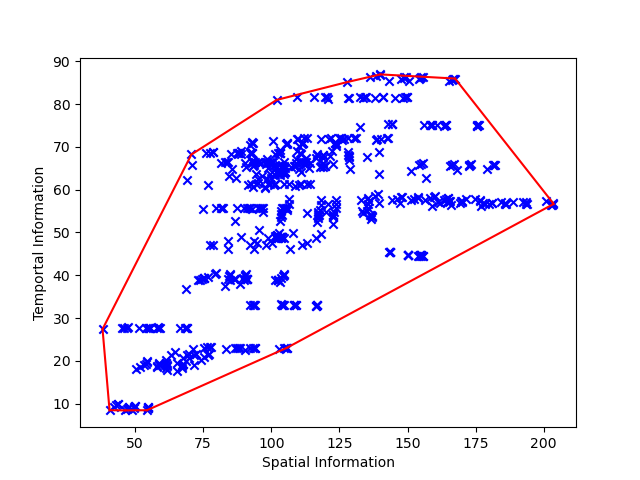}}}

    \caption{Source content (blue ‘x’) distribution in paired feature space with
corresponding convex hulls (red boundaries). Left column: Contrast x Brightness,
middle column: Sharpness x Colourfulness, right column: Temporal Information vs Spatial Information.}%
    \label{fig:objective_scores}
\end{figure*}

\subsection{Source Sequences}
\label{subsection:source}
We collected $16$ uncompressed, high-quality source gameplay videos from the Facebook Cloud Gaming servers. We recorded the raw YUV $4:2:0$ video game streams, which were rendered at the cloud servers without any impairments, i.e., before the cloud gaming application pipeline distorted the video stream during gameplay sessions. All of the obtained videos were of original 720p resolution and framerate $30$ frames per second, in raw YUV $4:2:0$ format, with their audio components removed. Since, we included both portrait and landscape games in the dataset, by 720p resolution we mean that either the width or the height is $720$ pixels, with the other dimension being at least $1280$ pixels and often larger. The video contents include $16$ different games encompassing diverse contents. Section \ref{appendix:gamenames} details the games present in the dataset along with their original resolutions as rendered by the Cloud Game engine. \\
\indent The original $16$ reference videos we collected ranged from $58$ seconds to $3$ minutes which were clipped to lengths that were practical for the human study. Deciding the clip durations presents decisions that depend on several factors. For example, using videos of varying lengths could lead to biases in the subjective ratings provided by the human volunteers. Using longer videos could limit the data diversity in human studies of necessarily limited participant duration. Moreover, long videos often exhibit distortion changes over time. While it would be worthwhile to investigate time varying distortions of gaming videos, that topic falls outside the scope of the current study, being more appropriate for ``Quality of Experience" (QoE) studies similar to  those presented in \cite{8093636}, \cite{7987076}, \cite{https://doi.org/10.48550/arxiv.1808.03898}. \\
\indent \textcolor{black}{The goal of our study is to conduct a passive viewing test that will enable us to annotate the video quality of gaming videos.} The results from the study \cite{8463417} illustrated that no significant differences were observed in video quality ratings obtained on the viewing of interactive and passive games that were of 90 seconds duration. However, passive tests of duration 10 seconds yielded significantly higher quality ratings on videos than longer passive tests, indicating that time-varying QoE factors play little role in short-duration tests. The ITU-T P.809 \cite{itut} standard recommends using 30-second videos when conducting passive human evaluation of gaming video quality. \textcolor{black}{However, we conducted a trial study involving 20 human participants, each of whom were shown gaming videos of durations ranging from 5 to 35 seconds and asked to provide subjective video quality ratings. The human participants' feedback led us to conclude that gaming videos of durations no more than 15-20 seconds were needed in order to comfortably provide subjective quality ratings. The feedback received generally indicated that it was sometimes difficult to comfortably rate videos that were 10 seconds or shorter, especially on those containing significant motion typical of gaming videos. On the other hand, videos that were 25 seconds or longer were reported to feel too lengthy, and that quality could have been accurately assessed within the initial 15-20 seconds. Moreover, some participants observed the video quality to change over the course of the 25-35 seconds, making it challenging to assign a single quality score.  Since the focus of the current study is not to study the time varying (QoE) effects sometimes observed on longer duration videos, we selected between one and three clips from each reference video, each of 20 seconds duration, yielding a total of 30 video clips drawn from the 16 reference videos, all of 720p resolution.} We took care that each clip did not include annoying disruptions of otherwise interesting gameplay, and also that clips from the same game presented different scenarios. By distorting the 30 video clips as described in Section \ref{ssubsection:disitor}, we obtained 600 videos. \\
\indent To illustrate the diversity of the video contents in the database, we calculated the following objective features: Brightness, Contrast, Colorfulness \cite{10.1117/12.477378}, Sharpness, Spatial Information and Temporal Information as recommended in \cite{6280595}, \cite{itutww} for all 600 videos in the database. We calculated the first four objective features on each video frame, then averaged them across all frames to obtain the final feature values. For each frame, brightness and contrast were determined as the mean and standard deviation of the pixel luminance values. We calculated the sharpness of each frame  by computing the mean sobel gradient magnitudes at each frame coordinate. We superimposed the convex hulls of the scatter plots of  pairs of these features, illustrating the broad feature coverage of the videos in Fig. \ref{fig:objective_scores}. \textcolor{black}{In Fig. \ref{fig:objfeats}, we compare the coverage of our proposed database against other existing Cloud Gaming databases.}
\begin{table}[]
\centering
\caption{Resolution and Bitrates values of the videos in the LIVE-META Mobile Cloud Gaming Database}
\begin{tabular}{|c|c|}
\hline
Encoding Parameter & Value                                    \\ \hline
Resolution         & 360p, 480p, 540p, 720p                   \\ \hline
Bitrate            & 250kbps, 500kbps, 800kbps, 2mbps, 50mbps \\ \hline
\end{tabular}
\label{tab:encoding-table}
\end{table}

\begin{figure*}[h!]
\centering
\includegraphics[width=0.82\textwidth]{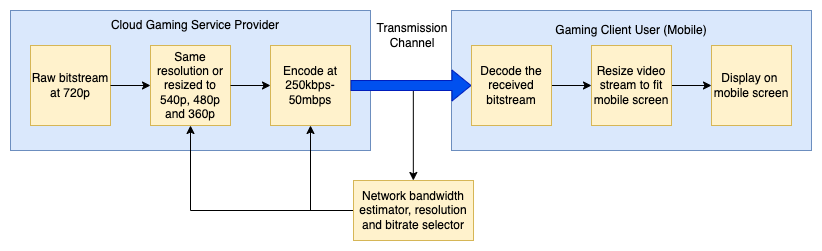}
\label{fig:first}
\caption{High-level flow diagram of the mobile cloud gaming pipeline used in the creation of LIVE-Meta Mobile Cloud Gaming database.}
\label{fig:simcld}
\end{figure*}
\subsection{Mobile Cloud Gaming Pipeline}
\label{ssubsection:disitor}
From each of the $30$ reference sequences, $20$ distorted video sequences were generated using a combination of resizing and compression distortion processes. Fig. \ref{fig:simcld} shows a simplified model of the mobile cloud gaming pipeline. The encoding settings we used are similar to those employed in the CGVDS database \cite{10.1145/3339825.3391872}. We used the Constant Bit Rate (CBR) encoding mode in the hardware accelerated NVIDIA NVENC H.264 encoder \cite{Nvenc}, with preset set to low latency and high quality. The videos were spatially resized using FFMPEG’s default
bicubic interpolation. \\
\indent We processed each of the $30$ reference videos using all $20$ possible combinations of resolutions and bitrates listed in Table \ref{tab:encoding-table}. The bitrates range from $250$ kbps to  \textcolor{black}{$50$} mbps, and the resolutions range from $360$p to $720$p. \textcolor{black}{The reference videos were first spatially resized to 360p, 480p, or 540p or they were maintained at the original 720p resolution, followed by encoding in CBR mode at different bitrates. } The selected combinations broadly emulate generic mobile cloud gaming services and available wireless network bandwidths. Most mobile cloud gaming service providers render games at $720$p resolution and then, depending on network conditions, either downscale the games to resolutions $360$p, $480$p, or $540$p, or maintain the original resolution before encoding the videos at constant bitrates. Based on our experiments, we generally observed that $250$ kbps was the lowest threshold of bandwidth for which acceptable levels of video quality were observed for most of the games in the dataset. We also encoded the videos at higher bitrates typical of common encoding scenarios: $500$ kbps, $800$ kbps, and $2$ mbps, in addition to $250$ kbps. \textcolor{black}{Our choice of bitrates ensured that we observed a wide range of perceptual qualities across these bitrates and contents. } \\
\textcolor{black}{\indent Contemporary subjective video quality databases commonly include reference videos. However, since Android mobile devices cannot play lossless (QP=$0$ encoded) videos, we could not directly incorporate true reference videos in the human study.  As an alternative, we encoded the videos at a very high bitrate of $50$ mbps to produce ``visually lossless" alternatives to uncompressed videos. We will refer to these videos as  ``proxy reference videos." We conducted a thorough visual inspection, comparing each reference video to its proxy reference, and concluded that the $50$ mbps bitrate was sufficiently high to preserve all visual information in the videos and prevent the introduction of visible artifacts, particularly when taking into account the maximum resolution of the videos was 720p.} \textcolor{black}{To further support the conclusions obtained by visual inspection, we also encoded the source videos using QP=$0$ and observed that the average bit rate of those videos across all the contents was less than that of the proxy reference videos ($50$ mbps). This strengthens our earlier claim of preserving the visual information in the proxy reference videos since more bits were allocated in the encoding process than would be required for lossless compression.} \textcolor{black}{We were also unable to include videos with only resizing distortions (i.e., without video compression) because of the same device limitation. However, following our observation that the proxy reference videos were ``visually lossless" when encoded at a bitrate of $50$ mbps, we used the same bitrate to encode the videos with only resizing distortions.}

\subsection{Subjective Testing Environment and Display}
We conducted the large-scale human study in the Subjective Study room in the Laboratory of Image and Video Engineering at The University of Texas at Austin. A Google Pixel $5$, running on the Android $11$ operating system, was used to display all videos using a custom-built android application. \textcolor{black}{We chose the popular and affordable mid-tier Google Pixel 5 mobile phone as a reasonably representative device that Cloud Gaming clients may often use. The device's compatibility with the Android operating system also provided us with great flexibility when developing the interface application for the subjective study. The Pixel 5's high-quality OLED display is renowned for its excellent color accuracy in the brightness range of 60 - 80\% of peak brightness \cite{xda}, making it an excellent choice. } \\
\indent The mobile device was interfaced with a wireless mouse and keyboard to enable the subjects to easily record video quality ratings. The Google Pixel $5$ has a $6$-inch OLED panel with a $19.5:9$ aspect ratio Full HD+ ($2340\times1080$) resolution and up to a 90Hz refresh rate. The adaptive brightness feature of the mobile device was disabled, and the brightness was set to $75\%$ of the maximum to prevent fluctuations during the study sessions. We utilized the mobile device's ability to automatically resize incoming video streams using its hardware scaler during cloud gaming, by up-scaling the videos displayed on the mobile device to fit the mobile screen during playback to the subjects. The Android application was memory and compute optimized to ensure smooth playback during the human study.\\  
\indent We arranged the lighting and environment of the LIVE Subjective Study room to simulate a living room. The room's glass windows were covered with black paper to prevent volunteers from being distracted by any outside activities. To achieve a similar level of illumination as one found in a  typical living room, we used two stand-up incandescent lamps, and also placed two white LED studio lights behind where the viewer was seated. We positioned all the lights so that there were no reflections of the light sources from the display screen visible to the subjects. The incident luminance on the display screen was measured by a lux meter and found to be approximately 200 Lux.  \\
\indent A sturdy smartphone mount similar to those found on car dashboards was deployed to secure the mobile device onto the subjects' desktop. The mount is telescopic, with adjustable viewing angles and heights of the mobile device. The study participants sat comfortably in height-adjustable chairs and were asked to adjust the viewing angle and the height of the mount so they could observe the videos played on the mobile device at approximately arm's length, similar to the experience of typical gameplay sessions. \\
\indent We created a video playlist for each participant. After each video was played, a continuous rating bar appeared with a cursor initialized to the extreme left. With the mouse connected wirelessly to the device, the volunteers could freely move the cursor to finalize the quality ratings they gave. There were five labels on the quality bar indicating Bad, Poor, Fair, Good and Excellent to help guide the participants when making their decisions. The subjects' scores were sampled as integers on $[0, 100]$ based on the final position of the cursor, where $0$ indicated the worst quality and $100$ the best. However, numerical values were not shown to the volunteers. To confirm the final score of each video, the volunteer pressed the NEXT button below the rating bar, and the score was then stored in a text file. The application then played the following video on the playlist. Fig. \ref{fig:appscreen} in the Appendix Section \ref{sec:appendix} demonstrates the steps involved in the video quality rating process in the Android application. 

\begin{table*}[]
\centering
\caption{Illustration of the round-robin approach used to allocate video groups to subject groups. Sessions A, B refer to the two sessions of the human study for every subject. Grid locations marked as X indicate the video group in the column was not rated by the subject group in the row. Each Video Group contained $100$ videos and each Subject Group has $12$ subjects}
\label{tab:protocol_table}
\resizebox{\textwidth}{!}{%
\begin{tabular}{|c|c|c|c|c|c|c|}
\hline
GROUP & Video Group : I & Video Group : II & Video Group : III & Video Group : IV & Video Group : V & Video Group : VI \\ \hline
Subject Group : 1 & Session A & Session B & X         & X         & X         & X         \\ \hline
Subject Group : 2 & X         & Session A & Session B & X         & X         & X         \\ \hline
Subject Group : 3 & Session B & X         & Session A & X         & X         & X         \\ \hline
Subject Group : 4 & X         & X         & X         & Session A & Session B & X         \\ \hline
Subject Group : 5 & X         & X         & X         & X         & Session A & Session B \\ \hline
Subject Group : 6 & X         & X         & X         & Session B & X         & Session A \\ \hline
\end{tabular}%
}
\end{table*}
\subsection{Subjective Testing Protocol}
We followed a single-stimulus (SS) testing protocol in the human study, as described in the ITU-R BT 500.13 recommendation \cite{itutpic}. As explained in Section \ref {ssubsection:disitor}, we could not include the actual reference videos due to limitations of the Mobile device, but we did include $50$ mbps, and $720$p resolution encoded versions of each source video as reasonable proxy reference videos. \\
\indent As explained in Section \ref{ssubsection:disitor}, we generated the $600$ processed videos by combinations of resizing and compression of the $30$ reference videos. The reference (and hence the distorted) videos include equal numbers of portrait and landscape videos. We divided the 30 reference videos into six groups in such a way that groups I, II, III were comprised only of portrait videos while groups IV, V, VI comprised only of landscape videos. In addition, we ensured that no two reference videos in a video group came from the same game. Since we generated 20 distorted versions of each reference video, each video group contained $5 * 20 = 100$ videos. We evenly split the $72$ human participants into six groups. Using a round-robin method, we assigned two video groups to each subject group across two sessions, A and B. The exact allocation of video groups for each subject group can be found in Table \ref{tab:protocol_table}. As shown in the Table \ref{tab:protocol_table}, since two subject groups rated each video group, we obtained $2 * 12 = 24$ ratings per video. We designed the study protocol as shown in Table \ref{tab:protocol_table} in a manner such that all the subjects watched either portrait or landscape orientation in both sessions, and never viewed both portrait and landscape videos. We used this approach to eliminate biases caused by any difference in subject preferences for one or the other orientation by any subject.\\
\indent For the human study, we developed a unique playlist for each session. The order of the videos in the playlist was randomized, with the constraint that videos generated from a reference video were separated by at least one video generated from another reference video. The randomized ordering of the videos reduced the possibility of visual memory effects or any bias caused by playing the videos in a particular order. Each human study session involved rating 100 videos, and required approximately $38-40$ minutes of each participant's time.

\subsection{Subject Screening and Training}
Seventy-two human student volunteers were recruited from various majors at The University of Texas at Austin to take part in the study. \textcolor{black}{The pool of subjects had little/no experience in image and video quality assessment}. Each subject participated in two sessions separated by at least 24 hours to avoid fatigue. \\ 
\indent At the beginning of a volunteer's first session, we administered the Snellen and Ishihara tests to validate each subject’s vision. Two subjects were found to have a color deficiency, while three volunteers had $20/30$ visual acuity. These tests were performed to ensure there was no abnormally high percentage of deficient subjects. All subjects, regardless of their vision deficiencies, were allowed to participate in the study, following our standard goal of designing more realistic psychometric video quality databases \cite{liveprac}. \textcolor{black}{In Section \ref{ssec:procscores}, we study impact of participants having imperfect vision on the study, by analysing the individual bias and consistency scores obtained using the maximum likelihood estimation algorithm described in \cite{DBLP:journals/corr/LiB16c}.}\\
\indent We explained the study objectives to each volunteer before they engaged in the experiment.  Volunteers were instructed to rate the gaming videos only on quality, and not on the appeal of the content, such as how boring or exciting the game content was or how well or poorly the player had performed on the recorded gaming video they were rating. Additionally, we demonstrated how the setup could be used to view and rate gaming videos. At the beginning of each test session, volunteers were shown three versions of a same video, which were of perceptually separated qualities to familiarize themselves with the system and to experience the ranges of video quality they would be rating. The scores subjects gave the training videos were not included in the psychometric database. 
\subsection{Post Study Questionnaire}
The subjects were asked to fill out a questionnaire at the end of each video quality rating session. The data were collected to ensure the reliability of the subjective ratings collected during the human study sessions. Within this sub-section, we present a summary of answers to those questions and demographic information about the subjects. \\
\indent In Section \ref{subsection:source}, we deliberated on how to determine the optimal duration of each video in our database. To reinforce the result from our pre-study trial (that $20$ seconds was long enough to comfortably rate the perceptual quality of each video), we asked every volunteer, as part of the post-study questionnaire, whether the duration of the videos was long enough. Out of the $144$ sessions ($72$ subjects, with $2$ sessions per subject) we conducted, in $97.9\%$ ($141/144$) of the sessions, the human subjects felt that the $20$-second duration was adequate to subjectively judge the video quality. \textcolor{black}{Furthermore, we investigated observer bias and consistency among the three volunteers who deemed the allocated 20 seconds to be inadequate to evaluate subjective video quality in Section \ref{ssec:procscores}.} Section \ref{appendix:additionalanalysis} summarizes the answers given to the questions regarding the difficulty of rating the videos, and any uneasiness/dizziness induced during the rating process. It also includes the demographic data of the human subjects.
\iftrue
\begin{figure*}[htbp]
    \centering
    \subfloat[Recovered quality scores.]{{\includegraphics[width=\textwidth]{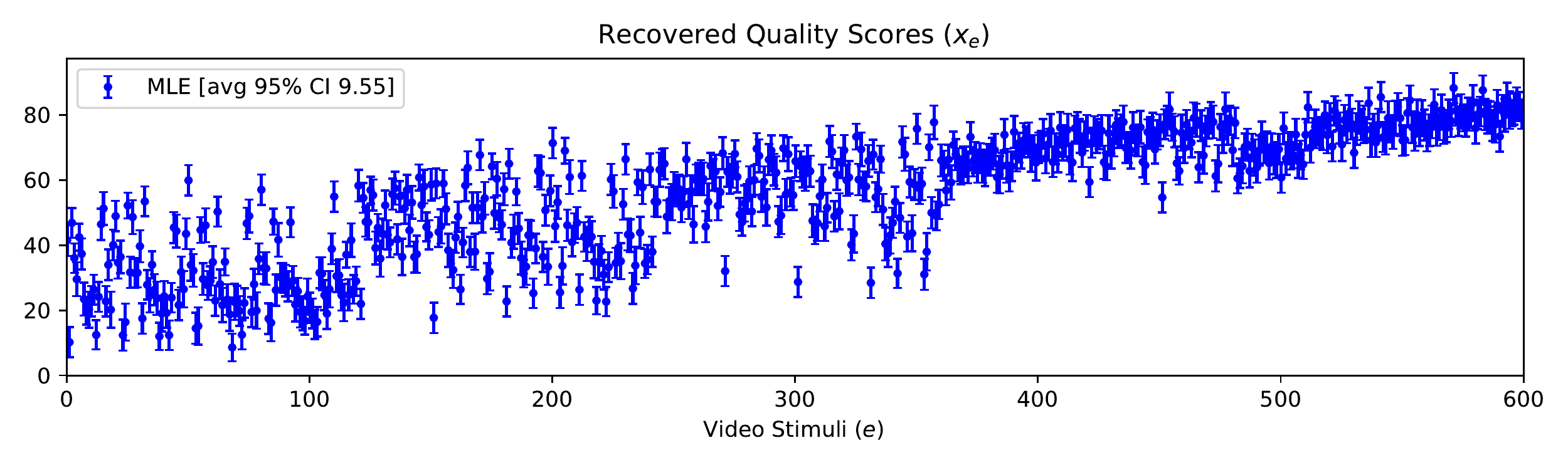}}\label{fig:outcomesa}}

    \subfloat[Subject bias and inconsistency.]{{\includegraphics[width=0.49\textwidth]{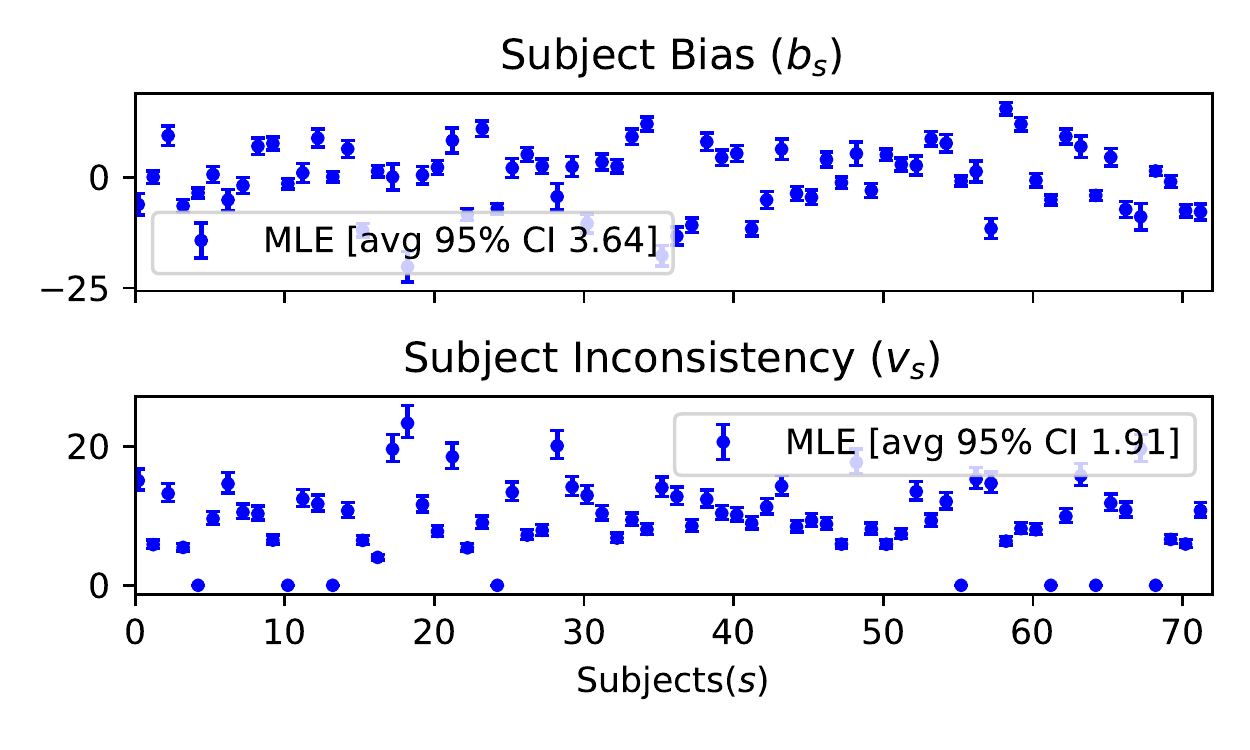}}\label{fig:outcomesb}}
    \hfill
    \subfloat[Content ambiguity.]{{\includegraphics[width=0.49\textwidth]{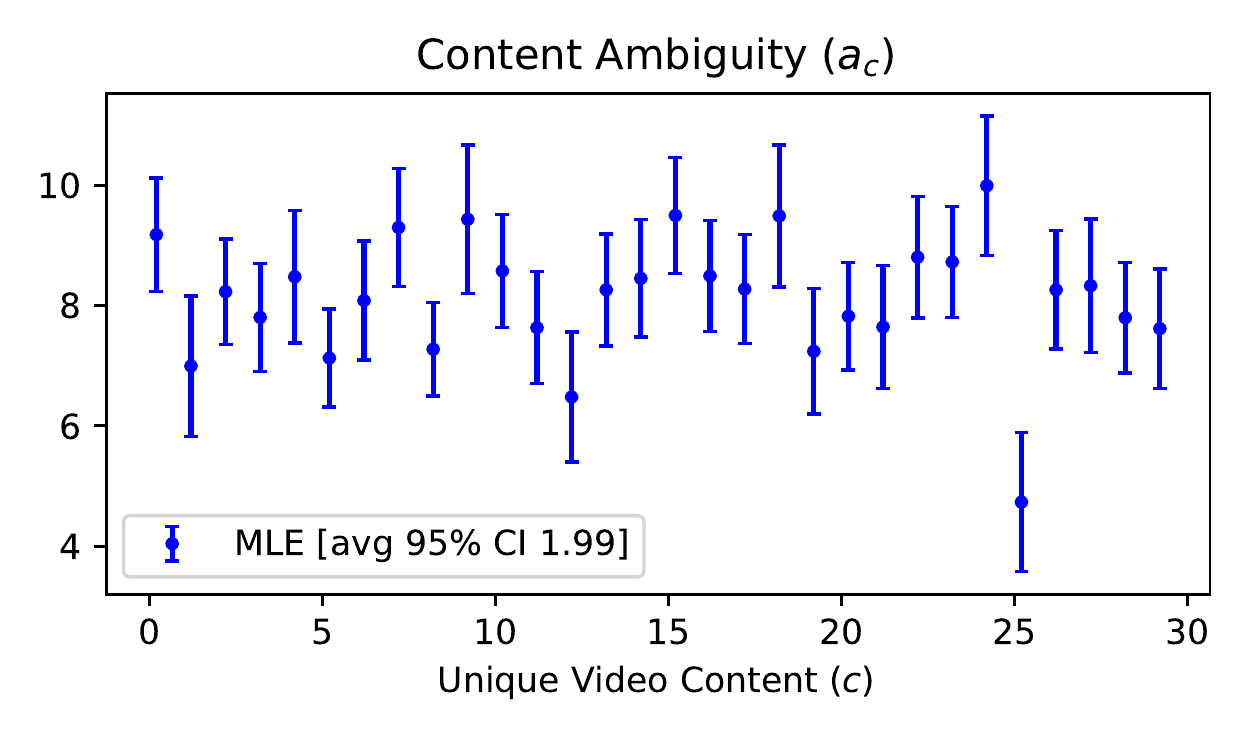}}\label{fig:outcomesc}}
    \caption{The result of the MLE formulation to estimate final opinion scores and associated information about subjects and contents.  Both the estimated parameters and their 95\% confidence intervals are shown.}%
    \label{fig:outcomes}
\end{figure*}
\fi
\subsection{Processing of Subjective Scores}
\label{ssec:procscores}
To ensure the reliability of the subjective data acquisition process, we first examined the inter-subject and intra-subject consistency of the data using the raw video quality ratings obtained from the human subjects. As explained earlier, we divided the $72$ subjects into $6$ groups as shown in Table \ref{tab:protocol_table}. We report the inter-subject consistency scores for each group. In order to determine inter-subject consistency, we randomly grouped the scores received for the videos rated by each subject group into two equal but disjoint subgroups, and computed the correlations of the mean opinion scores between the two sub-groups. The random groupings were performed over 100 trials and the medians of both the Spearman's Rank Order Correlation Coefficient (SROCC) and the Pearson Linear Correlation Coefficient (PLCC) between the two sub-groups were computed for each of the subject groups and are listed in Table  \ref{tab:consistency} in the Appendix Section \ref{sec:appendix}. Overall, the average SROCC and PLCC for inter-subject consistency across all subject groups was 0.912 and 0.929, respectively. Furthermore, we calculated intra-subject consistency measurements which provide insight into the behavior of individual subjects \cite{article} on the videos they rated. To do this, we measured the SROCC and PLCC between the individual opinion scores and MOS calculated using all the subjects within each subject group. This process was repeated for every human subject within all the subject groups. The medians for each of the subject groups for both SROCC and PLCC are listed in Table \ref{tab:consistency} in the Appendix Section \ref{sec:appendix}. The average SROCC and PLCC over all subject groups was respectively 0.848 and 0.860. These high correlation scores from the above analysis indicate that we can assign a high degree of confidence to the obtained opinion scores.\\
\indent We employed the method described in \cite{DBLP:journals/corr/LiB16c} to compute the final subjective quality scores on the videos using the raw subjective scores acquired from the human participants. The authors of \cite{DBLP:journals/corr/LiB16c} demonstrate that a maximum likelihood estimate (MLE) method of computing MOS offers advantages to traditional methods, by combining Z-score transformations and subject rejections \cite{itutpic}. The MLE method is less susceptible to subject corruption, provides tighter confidence intervals, better handles missing data, and can provide information on test subjects and video contents. \\
\indent In \cite{DBLP:journals/corr/LiB16c}, the raw opinion scores of the videos  are modeled as random variables ${\{X_{e,s}\}}$.  Decompose every rating of a video in the following way :
\begin{gather}
    \label{eq:sureal}
    X_{e,s} = x_e + B_{e,s} + A_{e,s},  \\   
    B_{e,s} \sim \mathcal{N}(b_s,v_s^2), \notag\\
    A_{e,s} \sim \mathcal{N}(0,a_{c:c(e)=c}^2), \notag
\end{gather}
where $e=1,2,3,...,600$ refer to the indices of the videos in the database and $s=1,2,3,...,72$ refers to the unique human participants. In the above model, $x_e$ represents the quality of the video $e$ as perceived by a hypothetical unbiased and consistent viewer. $B_{e,s}$ are i.i.d gaussian variables representing the human subject $s$ parameterized by a bias (i.e., mean) $b_s$ and inconsistency (i.e., variance) $v_s^2$. The human subject bias and inconsistency are assumed to remain constant across all the videos rated by the subject $s$. $A_{e,s}$ are i.i.d gaussian variables representing a particular video content parameterized by the ambiguity (i.e., variance) $a^2_c$ of the content $c$, and $c=1,2,...30$ indexes the unique source sequences in the database. All of the distorted versions of a reference video are presumed to contain the same level of ambiguity, and the video content ambiguity is assumed to be consistent across all users. In this formulation, the parameters $\theta=(\{x_e\},\{b_s\},\{v_s\},\{a_c\})$ denote the variables of the model. To estimate the parameters $\theta$ using MLE, the log likelihood function $L$ is defined as :
\begin{equation}
    L = \log P(\{x_{e,s}\}|\theta) 
\end{equation} 
Using the data obtained from the psychometric study, we derive a solution for $\hat\theta = \arg \max_{\theta} L$ using the Belief Propagation algorithm, as shown in \cite{DBLP:journals/corr/LiB16c}.

\indent Fig. \ref{fig:outcomes} shows a visual representation of the estimated parameters describing the recovered scores, the subject bias, and the inconsistency and content ambiguity. Fig. \ref{fig:outcomesa} shows the recovered quality scores for the $600$ videos in the database. The video files are indexed by increasing bitrate values, and further sorted by resolution within each bitrate group. The order of the presented video content is consistent across all resolutions and bitrates. According to our expectations, the average predicted quality scores of videos generally increased as bitrate was increased. Fig. \ref{fig:outcomesa} roughly identifies five clusters of videos based on predicted quality scores corresponding to the five bitrate values. Based on the parameter estimates obtained, the lowest bias value $b_s=-20.21$ was found for subject \#19, whereas the highest bias value $b_s=15.43$ was found for subject \#59, indicating subject \#19's quality scores were, on average, on the low side, while those of subject \#59 were, on average, on the high side, as compared to the other human subjects. \textcolor{black}{The median bias value obtained was 0.77.} Subject \#65 exhibited the greatest variability $v_s = 23.33$ when assigning quality judgements as indicated by the inconsistency estimates $v_s$, while subject \#19 exhibited the lowest level of variability $v_s = 2.06e^{-51}$. \textcolor{black}{The median of the inconsistency estimates was 9.49.} Fig. \ref{fig:outcomesc} shows the ambiguity in the 30 source videos. A source video from the State of Survival game had the lowest ambiguity $a_c = 4.73$, while a source video from the Sonic game had the highest ambiguity $a_c = 9.99$ among the 30 source videos. We denote the final opinion scores recovered using the above parameters as MLE-MOS. \\
\indent \textcolor{black}{We analysed both observer bias and inconsistency among individuals having imperfect vision.  We first consider observer bias. Earlier in this section, we reported that the minimum, median, and maximum of observer bias values across all subjects were $-20.21, 0.77,$ and $15.43$, respectively. The two subjects, \#32 and \#49, having color deficiencies, had estimated observer biases of $3.43$ and $5.30$, respectively, while the three subjects, \#29, \#58, and \#64, with 20/30 Snellen acuity had estimated observer bias values of $-11.59$, $6.90$, and $-4.39$, respectively. Since these bias values were not extrema, it is difficult to conclude that visual deficiencies had any impact on the subjective ratings. The minimum, median, and maximum subject inconsistencies across all subjects were estimated to be $2.06e^{-51}, 9.49, $ and $23.33$, respectively. The observer inconsistencies for \#32 and \#49 were estimated to be $10.35$ and $17.67$, respectively, while those for \#29, \#58, and \#64 were estimated to be $14.68$, $15.78$, and $20.06$, respectively. Although some inconsistency values were notably higher than the median, they were not extrema across all the subjects. Thus, we could not conclude that there was any induced observer inconsistency. A more detailed study, with subjects equally sampled with and without visual deficiencies, could better help reveal any impacts of color deficiencies and of slightly reduced visual acuity on video quality ratings. A similar analysis of observer bias and consistency was conducted for subjects \#2, \#47 and \#60, who deemed the 20-second duration insufficient to rate video quality in one of their sessions. The estimated observer bias values for these subjects were $0.01$, $3.96$, and $11.96$, respectively, and their estimated observer inconsistency values were $5.85$, $8.80$, and $8.16$, respectively. Again, the observer bias and inconsistency values for this group of individuals were not the highest or lowest values among all the subjects in our study. Hence, we could not make any significant conclusions or derive any notable insights from the analysis.  } \\
\indent MLE-MOS or MOS in general, is a reliable representation of subjective video quality and is required for the development and evaluation of No-Reference (NR) VQA algorithms, because reference undistorted videos are not available. The Difference MOS (DMOS) is more commonly used in the development and evaluation of Full Reference (FR) VQA algorithms because it allows the reduction of content-dependent quality labels. As discussed earlier, we use the 50 mbps encoded versions of the source videos at 720p resolution as the proxy reference videos when calculating the DMOS scores. The DMOS score of the $i^{th}$ video in the dataset is :
\begin{equation} \label{eq:dmos}
    DMOS(i) = 100 - (MOS(ref(i)) - MOS(i)),
\end{equation}
where MOS($i$) refers to the MLE-MOS of the $i^{th}$ distorted video obtained using the MLE formulation, and ref($i$) refers to the proxy reference video generated from the same source video sequence as the distorted video.


\begin{figure}[htbp]
    \centering
    \subfloat[Histogram of MLE-MOS of the human subjects using 20 equally spaced bins.\label{fig:histfirst}]{{\includegraphics[width=6.5cm]{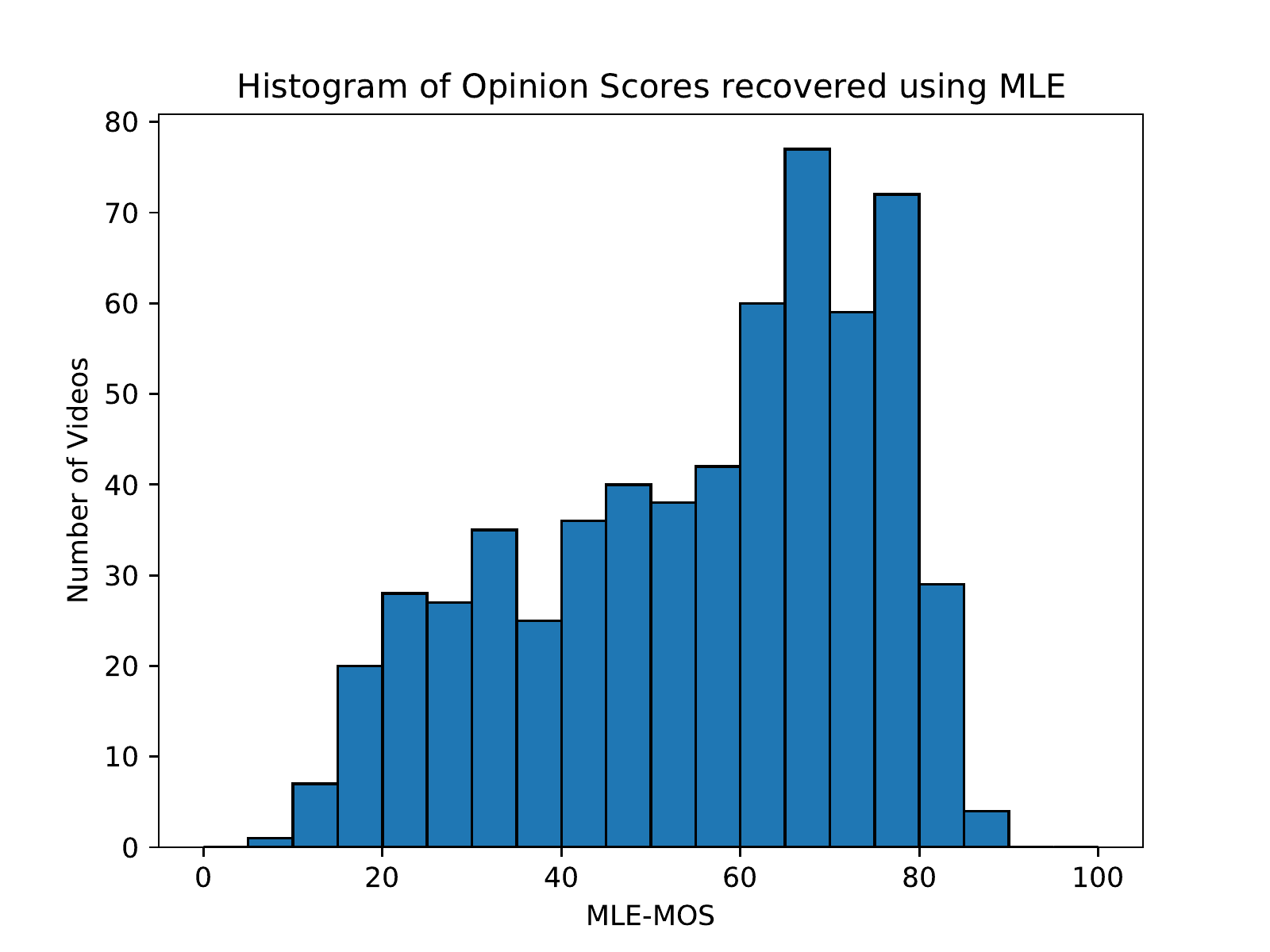}}}
  
    \subfloat[Histogram of DMOS of the human subjects using 20 equally spaced bins.\label{fig:histsecond}]{{\includegraphics[width=6.5cm]{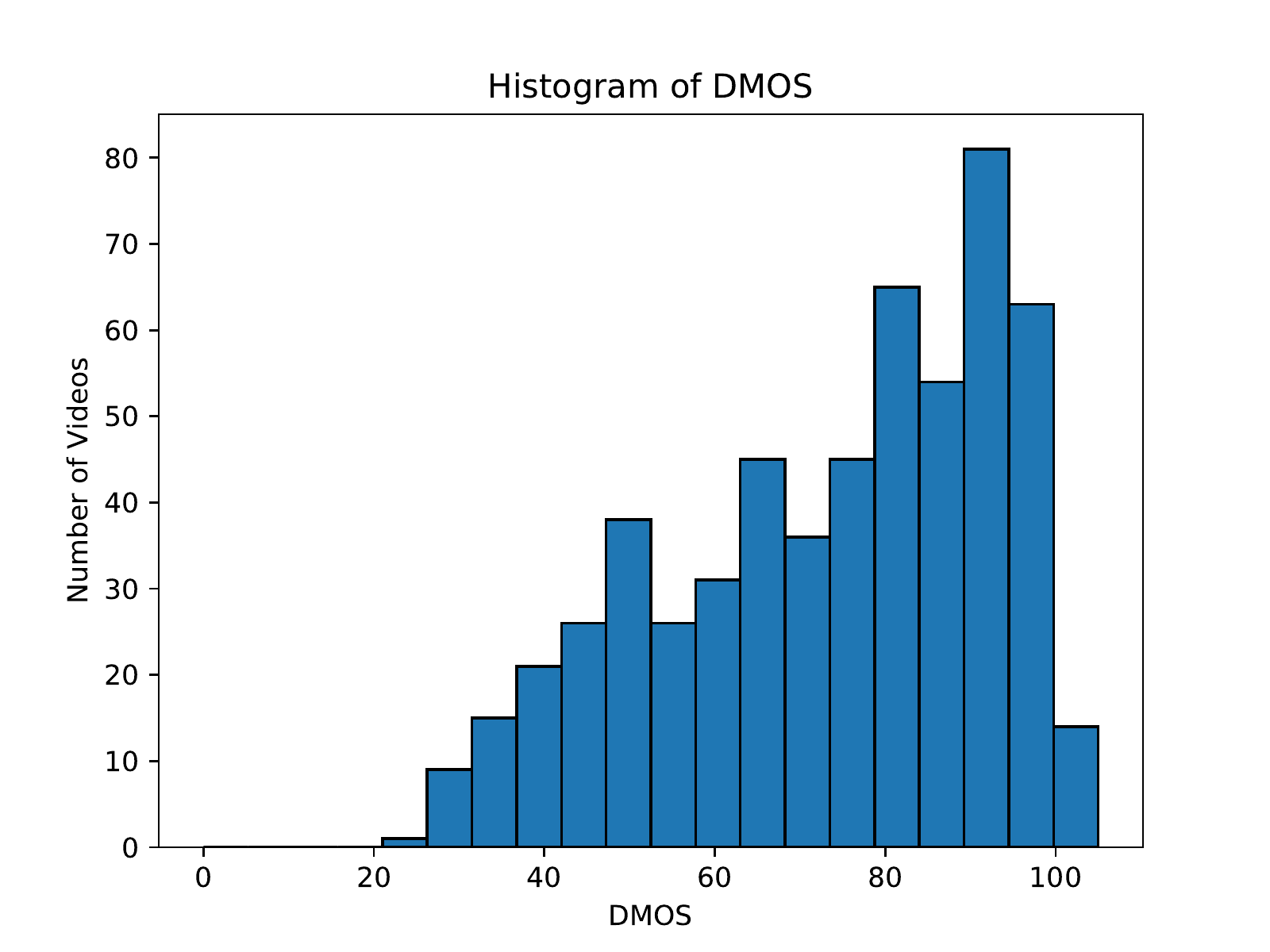}}}
        
    \caption{(a) MLE-MOS (b) DMOS for the LIVE-Meta Mobile Cloud Gaming Database. }%
    \label{fig:hist}
\end{figure}

\subsection{Analysis and Visualization of the Opinion Scores}
\label{ssec:mosvis}
\begin{figure}[htbp]
    \centering
    \subfloat[Average MLE-MOS vs Bitrate for Portrait and Landscape Videos.\label{fig:firstt}]{{\includegraphics[width=6.5cm]{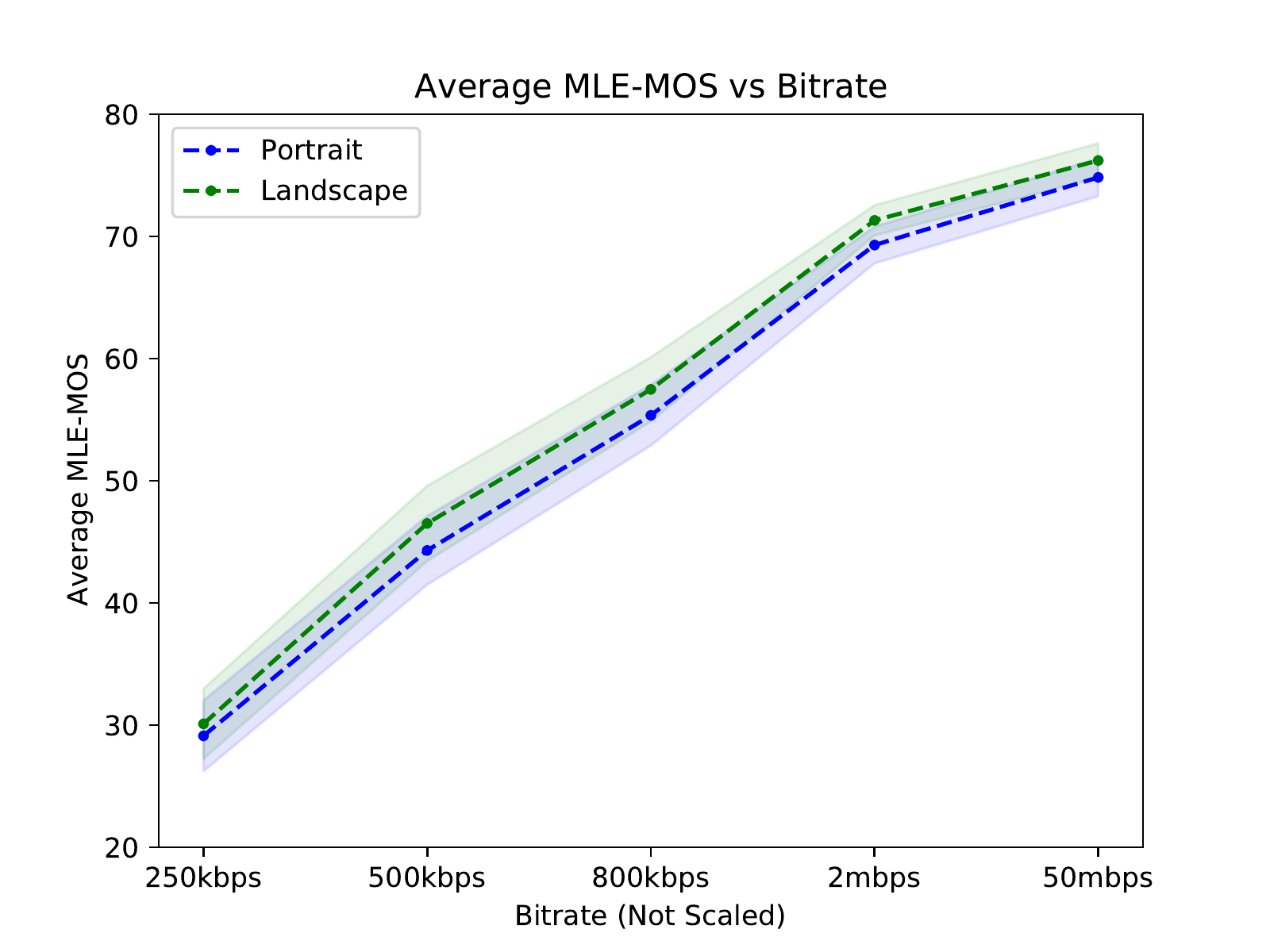}}}
  
    \subfloat[Average MLE-MOS vs Resolution for Portrait and Landscape Videos.\label{fig:secondd}]{{\includegraphics[width=6.5cm]{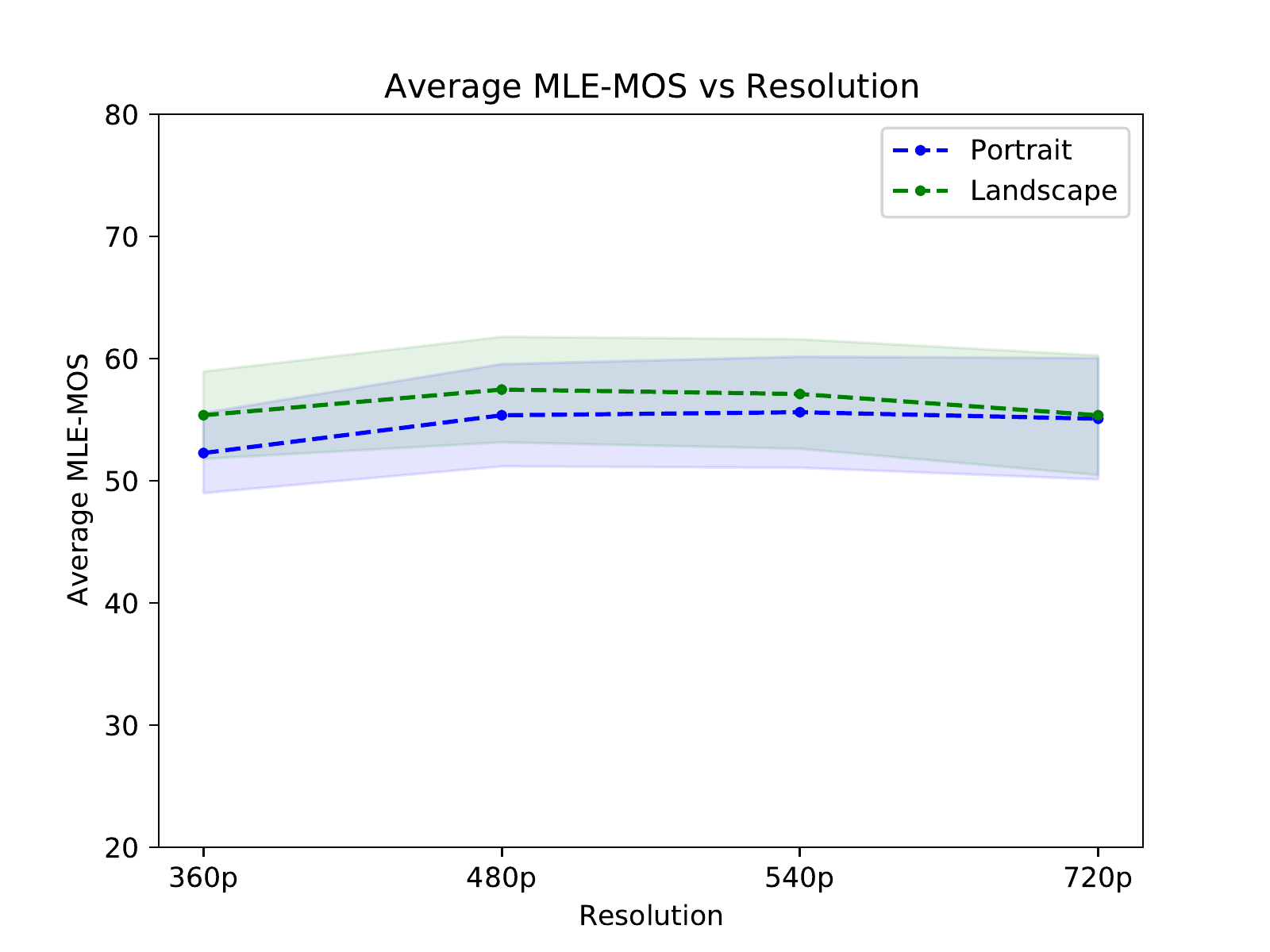}}}
        
    \caption{Comparison of the effect of Bitrate and Resolution on MLE-MOS for Landscape and Portrait Videos.}%
    \label{fig:portraitland}
\end{figure}

Fig. \ref{fig:histfirst} plots a histogram of the mean opinion scores recovered using the maximum likelihood estimator. The MLE-MOS of the videos in the database ranged from $[8.558,88.29]$. The MLE-MOS distribution shown in Fig. \ref{fig:histfirst} is slightly right-skewed, typical of other VQA databases. \textcolor{black}{Fig. \ref{fig:histsecond} plots the histogram of DMOS computed using equation \ref{eq:dmos}. The DMOS of the videos in our database ranged from $[21.94,104.04]$. The distribution of DMOS has a strong resemblance to that of MLE-MOS, with the only difference being a slight shift to the right.} \\
\indent Since our new dataset contains videos in both of the common display orientations (portrait and landscape), we also examined the statistics of the MLE-MOS on each of these two video categories. While the average MLE-MOS rating on all videos was $55.45$, it dropped to $54.578$ on the portrait videos, and rose to $56.322$ on the landscape video. Before reaching any conclusions, we conducted a two-sample one-sided t-test at the $95\%$ confidence interval, to determine whether the differences in the population means of the two video categories were statistically significant. The outcome of the test led us to conclude that the ratings on the two categories of oriented videos were statistically equivalent. We also plotted the average MLE-MOS scores as function of bitrate and resolution after partitioning the videos by orientation category in Fig. \ref{fig:portraitland}. Fig. \ref{fig:firstt} plots the average MLE-MOS for portrait and landscape videos against bitrate. Although the curve for landscape videos is slightly elevated above the one for portrait videos across all bitrates, applying a two sample one-sided t-tests at each bitrate concluded that the differences between were statistically insignificant. We observed that the average MLE-MOS increased monotonically against bitrate, as expected. A similar analysis was done on the average MLE-MOS of the portrait and landscape videos against resolution, as shown in Fig. \ref{fig:secondd}. Again, the plot of average MLE-MOS for landscape videos was higher than that of portrait videos across all resolutions, with the separation decreasing with increased resolution. Again, the differences were statistically insignificant across all resolutions. 

\textcolor{black}{The standard deviations of the estimated MLE-MOS were in the range $[2.023,2.917]$ with an average of 2.435. The corresponding $95\%$ confidence intervals of MLE-MOS estimates were in the range $[7.93, 11.433]$ with an average of $9.546$. We also separately computed the mean of 95\% confidence intervals of the MLE-MOS estimates for the portrait and landscape videos. The $95\%$ confidence intervals for the portrait videos were found to fall in the range $[8.421,11.433]$  with an average of $9.843$, while the landscape videos confidence intervals were in the range $[7.93,10.011]$  with an average of $9.25$. We verified that differences in the means of the $95\%$ confidence intervals of the MLE-MOS estimates between the portrait and landscape videos were statistically significant, by conducting a two-sample one-sided t-test. We also observed that the six source contents contributing to the highest magnitudes of the $95\%$ confidence interval in MLE-MOS estimates were all portrait videos. Based on this evidence, it may be hypothesized that landscape videos provide a more immersive experience than portrait videos, thanks to the horizontal alignment of the eyes. This may contribute to the tighter confidence intervals when measuring video quality.}

\begin{figure}[h!]
  \centering
  \centerline{\includegraphics[width=6.5cm]{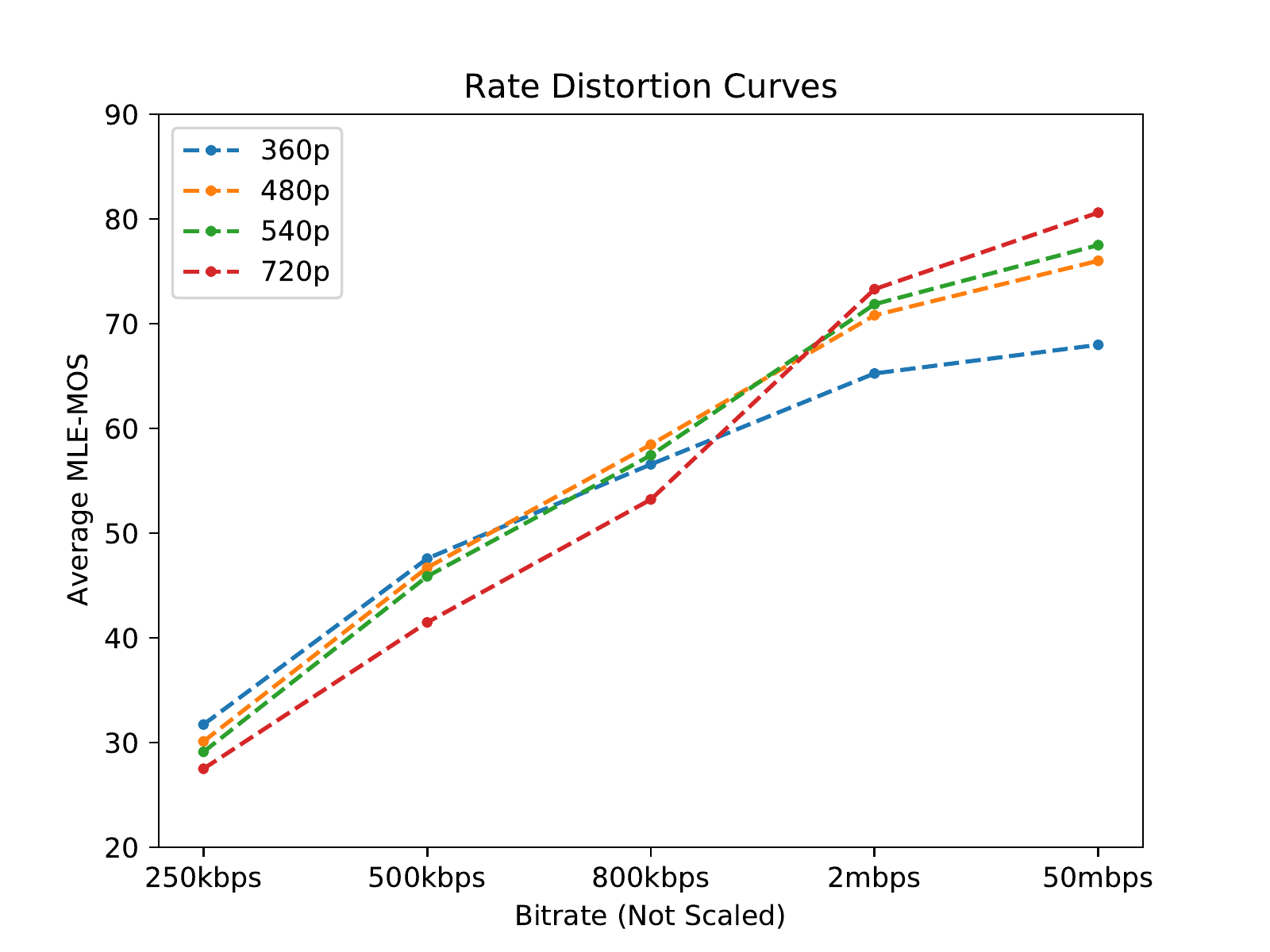}}
  \caption{Rate distortion curves at fixed resolutions.} \label{fig:rd}
\end{figure}
Fig. \ref{fig:rd} plots rate-distortion curves for all four resolutions of videos in the dataset. A plot of this type can supply clues regarding the selection of optimal streaming video resolutions as a function of bandwidth. We observed considerable overlap among the rate-distortion curves around the middle of the bitrate range  ($500$ kbps to $2$ mbps). Towards both lower and higher bitrates, the amount of overlap reduced, with $360$p being the most preferred resolution at bandwidths of $500$ kbps or less, and 720p the preferred resolution at $2$ mbps or higher. We provide additional analysis of the mean opinion scores in Section \ref{appendix:additionalanalysis} of the Appendix.

\begin{table*}[]
\centering
\caption{Median Srocc, Krcc, Plcc, and Rmse on the LIVE-META Mobile Cloud Gaming Database of NR-VQA Algorithms over $1000$ Train-Test Splits (Subjective MLE-MOS VS Predicted MLE-MOS). Standard Deviations are shown in parentheses. The best performing Algorithm is  Bold-Faced }
\label{tab:nrvqa}
\begin{tabular}{|c|c|c|c|c|}
\hline
Metrics      & SROCC$(\uparrow)$ & KRCC$(\uparrow)$ & PLCC$(\uparrow)$ & RMSE$(\downarrow)$ \\ \hline
NIQE         & -0.3900 (0.1816)   & -0.2795 (0.1366) & 0.4581 (0.2165)  & 16.5475 (1.9996)   \\ \hline
BRISQUE      & 0.7319 (0.1358)   & 0.5395 (0.1154)  & 0.7394 (0.1285)  & 12.5618 (2.5135)   \\ \hline
TLVQM        & 0.6553 (0.1428)   & 0.4777 (0.1166)  & 0.6889 (0.1464)  & 13.5413 (2.6724)   \\ \hline
VIDEVAL      & 0.7621 (0.1061)   & 0.5756 (0.0982)  & 0.7763 (0.1105)  & 11.7520 (2.2783)   \\ \hline
RAPIQUE      & 0.8740 (0.0673)   & 0.6964 (0.0759)  & 0.9039 (0.0565)  & 8.0242 (1.6755)    \\ \hline
GAME-VQP     & 0.8709 (0.0616)   & 0.6885 (0.0714)  & 0.8882 (0.0560)  & 8.5960 (1.7621)    \\ \hline
NDNet-Gaming & 0.8382 (0.1227)   & 0.6485 (0.1009)  & 0.8200 (0.1227)  & 10.5757 (3.0354)   \\ \hline
VSFA         & 0.9143 (0.0435)     & 0.7484 (0.0572)    & 0.9264 (0.0380)    & 7.1316 (1.6082)      \\ \hline
\textbf{GAMIVAL}      & \textbf{0.9441 (0.0281)}     & \textbf{0.7964 (0.0474)}   & \textbf{0.9524 (0.0290)}   & \textbf{5.7683 (1.429)}     \\ \hline
\end{tabular}
\end{table*}

\begin{table*}[]
\centering
\caption{Results of One-Sided T-Test Performed Using the 1000 (Srocc, Plcc) values of the compared NR-VQA Algorithms computed on the LIVE-META MCG Database. Each Cell contains 2 Symbols: the first symbol corresponds to the T-Test done using the Srocc values, and the second corresponds to the T-TEST done using the Plcc values. When a symbol `1' appears, it denotes that the algorithm on the row was statistically superior to that on the column, whereas `0' indicates that the algorithm on the column was statistically superior. A `-' symbol indicates that the column and row algorithms performed equally well }
\label{tab:stateq}
\begin{tabular}{|c|c|c|c|c|c|c|c|c|c|}
\hline
ALGORITHM    & NIQE & BRISQUE & TLVQM & VIDEVAL & RAPIQUE & GAME-VQP & NDNet-Gaming & VSFA & GAMIVAL \\ \hline
NIQE         & (-,-)   & (0,0)      & (0,0)    & (0,0)      & (0,0)      & (0,0)       & (0,0)           & (0,0)   & (0,0)      \\ \hline
BRISQUE      & (1,1)   & (-,-)      & (1,1)    & (0,0)      & (0,0)      & (0,0)       & (0,0)           & (0,0)   & (0,0)      \\ \hline
TLVQM        & (1,1)   & (0,0)      & (-,-)    & (0,0)      & (0,0)      & (0,0)       & (0,0)           & (0,0)   & (0,0)      \\ \hline
VIDEVAL      & (1,1)   & (1,1)      & (1,1)    & (-,-)      & (0,0)      & (0,0)       & (0,0)           & (0,0)   & (0,0)      \\ \hline
RAPIQUE      & (1,1)   & (1,1)      & (1,1)    & (1,1)      & (-,-)      & (-,1)       & (1,1)           & (0,0)   & (0,0)      \\ \hline
GAME-VQP     & (1,1)   & (1,1)      & (1,1)    & (1,1)      & (-,0)      & (-,-)       & (1,1)           & (0,0)   & (0,0)      \\ \hline
NDNet-Gaming & (1,1)   & (1,1)     & (1,1)    & (1,1)      & (0,0)      & (0,0)       & (-,-)           & (0,0)   & (0,0)      \\ \hline
VSFA         & (1,1)   & (1,1)      & (1,1)    & (1,1)      & (1,1)      & (1,1)       & (1,1)           & (-,-)   & (0,0)      \\ \hline
GAMIVAL      & (1,1)   & (1,1)      & (1,1)    & (1,1)      & (1,1)      & (1,1)       & (1,1)           & (1,1)   & (-,-)      \\ \hline
\end{tabular}
\end{table*}

\section{Benchmarking Objective NR-VQA Algorithms}
\label{sec:bench}
To demonstrate the usefulness of the new data resource, we evaluated a number of publicly available No-Reference (NR-VQA) algorithms on the LIVE-Meta MCG database. We selected six well-known general-purpose NR-VQA models to test :  NIQE \cite{6353522}, BRISQUE \cite{6272356}, TLVQM \cite{8742797}, VIDEVAL \cite{9405420}, RAPIQUE \cite{DBLP:journals/corr/abs-2101-10955}, and VSFA \cite{DBLP:journals/corr/abs-1908-00375}, as well as three NR-VQA models that were specifically developed for gaming video quality assessment tasks : NDNet-Gaming \cite{NDNetgaming}, GAME-VQP \cite{https://doi.org/10.48550/arxiv.2203.12824} and GAMIVAL \cite{gamival}. NIQE and BRISQUE are frame-based, and operate by extracting quality-aware features on each frame, then average pooling them to obtain quality feature representations. For the unsupervised, training-free model NIQE, the predicted frame quality scores were directly pooled, yielding the final video quality scores. For the supervised methods (BRISQUE, TLVQM, VIDEVAL, RAPIQUE, GAME-VQP and GAMIVAL),  we used a support vector regressor (SVR) with the radial basis function kernel to learn mappings from the pooled quality-aware features to the ground truth MLE-MOS. VSFA uses a Resnet-$50$ \cite{DBLP:journals/corr/HeZRS15} deep learning backbone to obtain quality-aware features, followed by a single layer Artificial Neural Network (ANN) and Gated Rectified Unit (GRU) \cite{DBLP:journals/corr/ChoMGBSB14} to map features to MLE-MOS. The NDNet-Gaming model however, regressed the video quality scores directly using a Densenet-$121$ \cite{DBLP:journals/corr/HuangLW16a} deep learning backbone. GAMIVAL modifies RAPIQUE's natural scene statistics model and replaces its Imagenet \cite{5206848} pretrained Resnet-50 CNN feature extractor with the Densenet-121 backbone used in NDNet-Gaming \\
\indent We evaluated the performance of the objective NR-VQA algorithms using the following metrics: Spearman's Rank Order Correlation Coefficient (SROCC), Kendall Rank Correlation Coefficient (KRCC), Pearson's Linear Correlation Coefficient (PLCC), and Root Mean Square Error (RMSE). The metrics SROCC and KRCC measure the monotonicity of the objective model prediction with respect to human scores, while the metrics PLCC and RMSE measure prediction accuracy. As stated earlier for the PLCC and RMSE measures, the predicted quality scores were passed through a logistic non-linearity function \cite{5404314} to further linearize the objective predictions and to place them on the same scale as MLE-MOS :
$$
f(x)=\beta_{2}+\frac{\beta_{1}-\beta_{2}}{1+\exp \left(-x+\beta_{3} /\left|\beta_{4}\right|\right)}
$$
We tested the algorithms mentioned above on $1000$ random train-test splits using the four metrics. For each split, the training and validation set consisted of videos randomly selected from $80\%$ of the contents, while videos from the remaining $20\%$ constituted the test set. We also ensured that the contents of the training and validation sets were always mutually disjoint. We separated the contents in the training, validation, and test sets to ensure that the content of the videos would not influence the performance of the NR-VQA algorithms. Other than NIQE and NDNet-Gaming, all of the algorithms were trained on one part of the dataset, then tested using the other, using the aforementioned train-test dataset split. Since NIQE is an unsupervised model, we evaluated its performance on all $1000$ test sets, without any training. We also evaluated NDNet-Gaming using the available pre-trained model on all of the 1000 tests sets, since training code was not available from the authors. We applied five-fold cross-validation to the training and validation sets of BRISQUE, TLVQM, VIDEVAL, RAPIQUE, GAME-VQP and GAMIVAL to find the optimal parameters of the SVRs they were built on. When testing VSFA, for each of the 1000 splits, the train and validation videos were used to select the best performing ANN-GRU model weights on the validation set.

\begin{table*}[]
\centering
\caption{Median Srocc, Krcc, Plcc, and Rmse of the compared NR-VQA Models on the LIVE-META Mobile Cloud Gaming Database, divided by display orientations, over $400$ train-test splits. Standard Deviations are shown in parentheses. The Best Performing Algorithm is Bold-Faced}
\label{tab:portland}
\begin{tabular}{|c|cccl|cccl|}
\hline
 &
  \multicolumn{4}{c|}{Landscape Videos} &
  \multicolumn{4}{c|}{Portrait Videos} 
   \\ \hline
Metrics &
  \multicolumn{1}{c|}{RAPIQUE} &
  \multicolumn{1}{c|}{GAME-VQP} &
  \multicolumn{1}{c|}{VSFA} &
  \textbf{GAMIVAL} &
  \multicolumn{1}{c|}{RAPIQUE} &
  \multicolumn{1}{c|}{GAME-VQP} &
  \multicolumn{1}{c|}{\textbf{VSFA}} &
  GAMIVAL \\ \hline
SROCC$(\uparrow)$ &
  \multicolumn{1}{c|}{0.876 (0.120)} &
  \multicolumn{1}{c|}{0.885 (0.087)} &
  \multicolumn{1}{c|}{0.927 (0.084)} &
  \textbf{0.955 (0.035)} &
  \multicolumn{1}{c|}{0.851 (0.122)} &
  \multicolumn{1}{c|}{0.850 (0.111)} &
  \multicolumn{1}{c|}{\textbf{0.903 (0.076)}} &
  0.900 (0.062) \\ \hline
KRCC$(\uparrow)$ &
  \multicolumn{1}{c|}{0.701 (0.117)} &
  \multicolumn{1}{c|}{0.715 (0.093)} &
  \multicolumn{1}{c|}{0.774 (0.090)} &
  \textbf{0.829 (0.056)} &
  \multicolumn{1}{c|}{0.680 (0.124)} &
  \multicolumn{1}{c|}{0.673 (0.109)} &
\multicolumn{1}{c|}{0.732 (0.087)} &
  \textbf{0.735 (0.083)} \\ \hline
PLCC$(\uparrow)$ &
  \multicolumn{1}{c|}{0.919 (0.103)} &
  \multicolumn{1}{c|}{0.912 (0.069)} &
  \multicolumn{1}{c|}{0.946 (0.071)} &
  \textbf{0.969 (0.023)} &
  \multicolumn{1}{c|}{0.882 (0.122)} &
  \multicolumn{1}{c|}{0.876 (0.103)} &
  \multicolumn{1}{c|}{\textbf{0.916 (0.075)}} &
  0.912 (0.068) \\ \hline
RMSE$(\downarrow)$ &
  \multicolumn{1}{c|}{7.294 (2.811)} &
  \multicolumn{1}{c|}{7.470 (2.630)} &
  \multicolumn{1}{c|}{5.873 (2.226)} &
  \textbf{4.547 (1.525)} &
  \multicolumn{1}{c|}{8.723 (2.632)} &
  \multicolumn{1}{c|}{8.706 (2.504)} &
  \multicolumn{1}{c|}{\textbf{7.371 (2.822)}} &
  7.417 (2.576) \\ \hline
\end{tabular}
\end{table*}

\begin{table}[]
\centering
\caption{Computation Complexity expressed in terms of Time and  Floating Point Operations (FLOPS) on $600$ Frames of a $360\times720$ Video upscaled to $1080\times2160$ Frames from the LIVE-META MCG Database}
\label{tab:complexity}
\begin{tabular}{|c|c|c|c|}
\hline
ALGORITHM & Platform &  \begin{tabular}[c]{@{}c@{}}Time \\ (seconds)\end{tabular} & \begin{tabular}[c]{@{}c@{}}FLOPS\\ ($\times10^9$)\end{tabular} \\ \hline
NIQE       & MATLAB  &  728  & 1965 \\ \hline
BRISQUE    & MATLAB  & 205  & 241 \\ \hline
TLVQM      & MATLAB  & 588  & 283 \\ \hline
VIDEVAL    & MATLAB  & 959  & 2334 \\ \hline
RAPIQUE    & MATLAB   & 103  & 322 \\ \hline
GAME-VQP   & MATLAB   & 2053  & 11627 \\ \hline
NDNet-Gaming & Python, Tensorflow   & 779 & 126704 \\ \hline
VSFA       & Python, Pytorch  & 2385  & 229079 \\ \hline
GAMIVAL       & \begin{tabular}[c]{@{}c@{}}Python, Tensorflow, \\ MATLAB\end{tabular}  & 201  & 8683 \\ \hline
\end{tabular}
\end{table}

\subsection{Performance of NR-VQA Models}
\label{ssec:sed}
\begin{figure}[h!]
  \centering
  \centerline{\includegraphics[width=6.5cm]{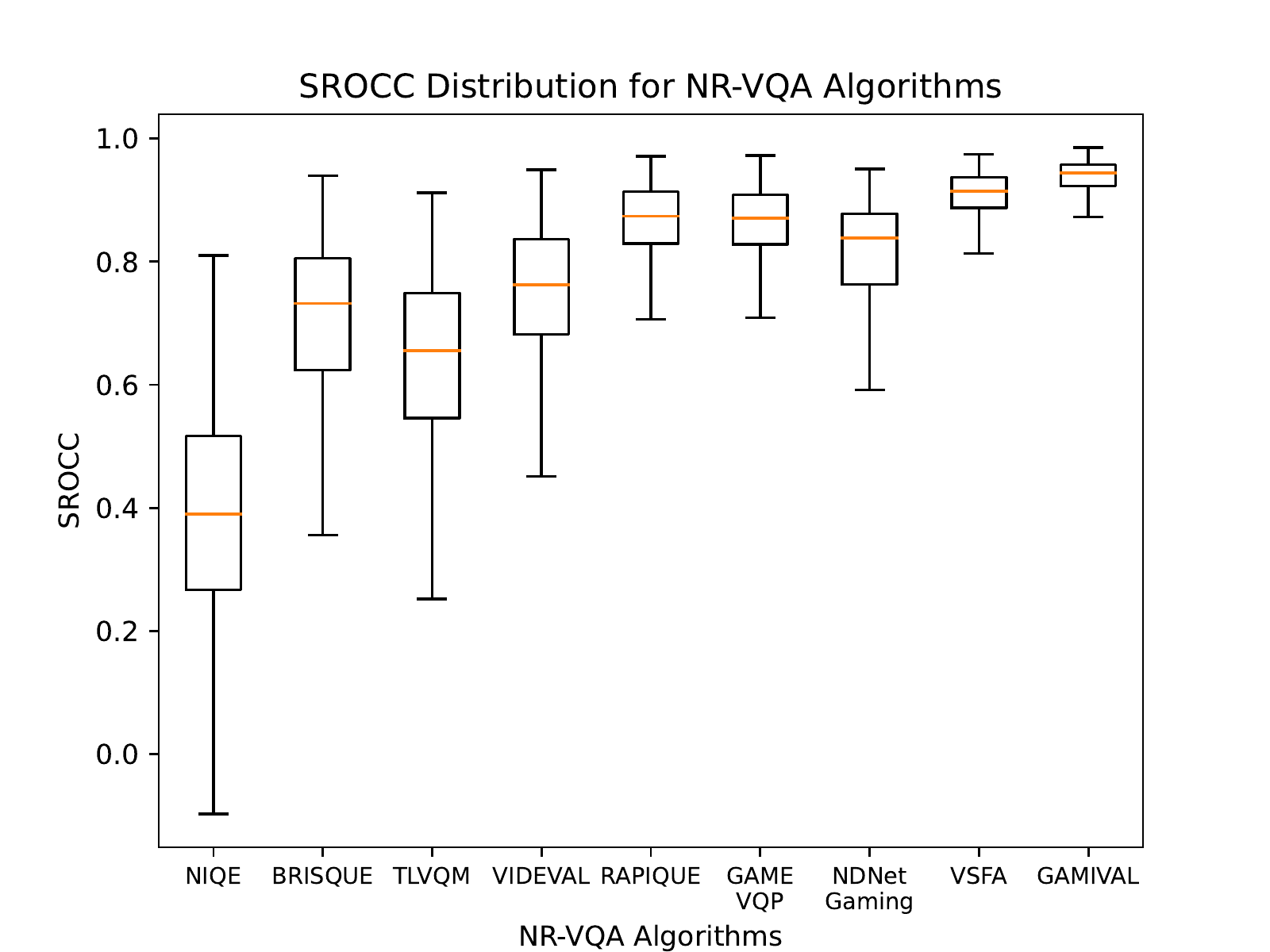}}
  \caption{Boxplots of SROCC distributions of the compared NR-VQA
algorithms.} \label{fig:boxplot}
\end{figure}

Table \ref{tab:nrvqa} lists the performances of the aforementioned NR-VQA algorithms on the LIVE-Meta Mobile Cloud Gaming database. In addition, we used the 1000 SROCC and PLCC scores produced by the NR VQA models to run one-sided t-tests, using the $95\%$ confidence level, to determine whether one VQA algorithm was statistically superior to another. Each entry in Table \ref{tab:stateq} consists of two symbols, where the first symbol corresponds to the t-test done using the SROCC values, and the second symbol corresponds to the t-test done using the PLCC values. We found that NIQE performed poorly, which is unsurprising since it was developed using natural images, while gaming videos are rendered synthetically and have different statistical structures. However, the performance of the same NIQE features improved when we extracted them and used an SVR to regress from the features to the MLE-MOS in the BRISQUE algorithm. The gap in performance between NIQE and BRISQUE points to the differences in the statistics of camera-captured videos of the real world as compared to graphical rendered synthetic gaming video scenes. However, BRISQUE was able to adapt to these synthetic scene statistics. The performance of TLVQM was average, probably because that model uses many hand-tuned hyper-parameters that were selected to optimize the prediction of video quality on general purpose content and do not generalize well to gaming videos. A similar scenario occurs with VIDEVAL. Although VIDEVAL had slightly boosted performance relative to BRISQUE, its performance may be limited since it uses $60$ features selected from more than 700 to maximize performance on in-the-wild UGC videos. Models that use deep learning like VSFA and NDNet-Gaming, and others that use hybrids of deep-learning-based features and handcrafted perceptual features, like RAPIQUE, GAME-VQP and GAMIVAL exhibit considerably improved performance, showing that they are able to capture the statistical structure of synthetically generated gaming videos, suggesting their potential as VQA algorithms targetting Cloud Gaming applications. The NR-VQA algorithms GAME-VQP and RAPIQUE use a combination of traditional NSS and deep-learning features to considerably improve performance relative to BRISQUE, VIDEVAL, and TLVQM on the LIVE-Meta MCG database. The superior performance of the VSFA model over GAME-VQP and RAPIQUE using only deep-learning features might indicate a reduced relevance of NSS features in the context of NR-VQA for cloud gaming. However, the GAMIVAL model, which uses adaptations of traditional NSS features, similar to the use of neural noise models in \cite{DBLP:journals/corr/abs-2106-13328}, along with deep-learning features, produced superior performance on synthetic gaming video content, suggesting the relevance of appropriately modified NSS features for synthetic rendered content. Fig. \ref{fig:boxplot} shows boxplots of the SROCC values computed on the predictions produced by each NR-VQA models, visually illustrating the results reported in Table \ref{tab:nrvqa}. The two top-performing algorithms VSFA and GAMIVAL exhibit very low variances of SROCC values, suggesting the reliability of these algorithms across multiple train-test splits. 

\subsection{Effects of Display Orientation on VQA Performance}
The new LIVE-Meta MCG database contains both portrait and landscape videos, allowing us to test the performances of NR-VQA algorithms on different display orientations. We tested the performance of the top-performing algorithms RAPIQUE, GAME-VQP, VSFA, and GAMIVAL on videos of both orientations over 400 train-test splits each. We may conclude from the results shown in Table \ref{tab:portland} that the NR-VQA algorithms performed slightly better when trained on landscape videos, than on portrait videos. Further, we performed one-sided t-tests using the 400 SROCC and PLCC scores used to report the results in Table \ref{tab:portland}. We were able to conclude from the results of the tests that the performances of the NR-VQA algorithms were statistically superior when trained on landscape videos than on portrait videos. \textcolor{black}{This could be attributed to the tighter $95\%$ confidence intervals of the MLE-MOS estimates obtained on landscape videos as compared to portrait videos,  as discussed in Sec. \ref{ssec:mosvis}}. From Tables \ref{tab:nrvqa} and \ref{tab:portland}, one may observe that  although overall GAMIVAL is the best performing algorithm on the LIVE-Meta MCG database, VSFA delivered slightly superior performance on the portrait gaming videos. 
\subsection{Comparison of Computational Requirements and Runtime}
\begin{figure}[h!]
  \centering
  \centerline{\includegraphics[width=6.5cm]{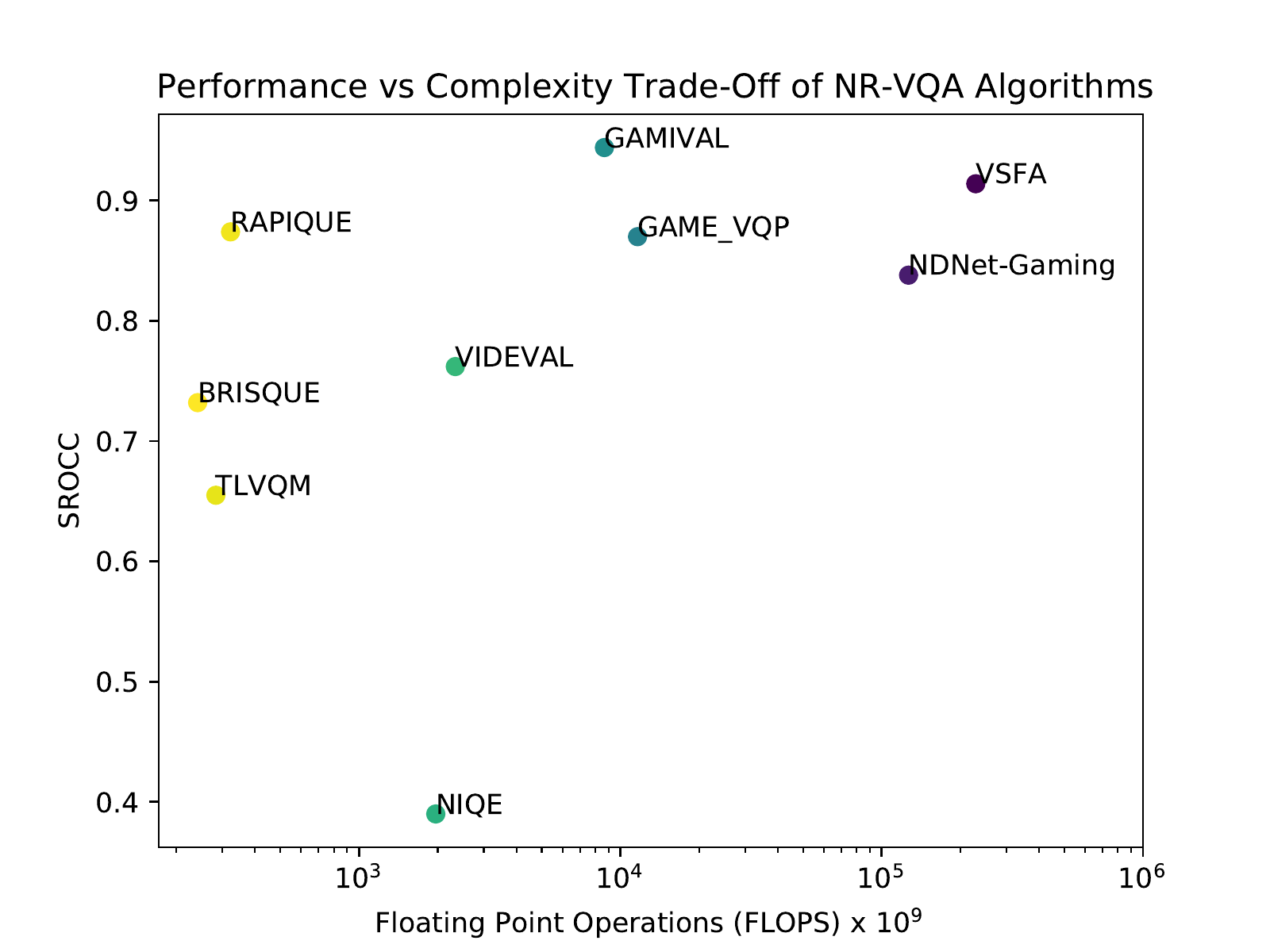}}
  \caption{Comparison of Performance vs Computational Requirement of NR-VQA Algorithms. FLOPs are shown in GigaFlops and shown in log scale.} \label{fig:tradeoff}
\end{figure}

This section analyzes the performance vs. complexity trade-off of the NR-VQA algorithms studied in Section \ref{ssec:sed}. All of the algorithms were run on a standalone computer equipped with an Intel Xeon E5-2620 v4 CPU running at a maximum frequency of 3 GHz. We used one of the videos from the LIVE-Meta MCG database of 360x720 resolution, upscaled it to the display resolution (1080x2160), and applied the algorithms on it. We report the execution time and the floating-point operations used by each algorithm in Table \ref{tab:complexity}. The algorithms VSFA and NDNet-Gaming were implemented in Python, GAMIVAL was implemented partly in MATLAB and partly in Python, while all the other algorithms were implemented in MATLAB. During the evaluation of deep NR-VQA algorithms, we ensured that the GPU was not used for fair comparison against other algorithms implemented on the CPU. From the results reported in Table \ref{tab:complexity}, none of the tested algorithms implemented in high level prototyping languages like MATLAB/Python run in real-time in their current implementations, however, they may be optimized for specific hardware using low-level languages like C/C++ by effectively exploiting their parallel processing capabilities in an application-specific setup. Based on the arguments presented above, we plotted the performance versus complexity trade-off (SROCC versus FLOPS) for each of the algorithms in Fig. \ref{fig:tradeoff}. Different orders of magnitude of FLOPS of the NR-VQA algorithms are indicated by distinct colors. The figure shows that the top four algorithms, RAPIQUE, GAME-VQP, VSFA and GAMIVAL, are computationally complex in varying degrees, with RAPIQUE having the lowest computational complexity and VSFA the highest. In addition to being the top-performing algorithm, GAMIVAL is also computationally efficient compared to VSFA and NDNet-Gaming, making it a viable option when evaluating the video quality of Mobile Cloud Gaming.

\begin{table*}[]
\centering
\caption{Median Srocc, Krcc, Plcc, and Rmse of FR-VQA Algorithms on the LIVE-META Mobile Cloud Gaming Database over $1000$ Train-Test Splits (Subjective DMOS VS Predicted DMOS). Standard Deviations are shown in parentheses. The best performing Algorithm is  Bold-Faced }
\label{tab:frvqa}
\begin{tabular}{|c|c|c|c|c|}
\hline
Metrics      & SROCC$(\uparrow)$ & KRCC$(\uparrow)$ & PLCC$(\uparrow)$ & RMSE$(\downarrow)$ \\ \hline
PSNR         & 0.7093 (0.0681)   & 0.5329 (0.0616) & 0.7172 (0.0676)  & 13.1194 (1.2216)   \\ \hline
SSIM      & 0.9235 (0.0301)   & 0.7647 (0.0435)  & 0.9332 (0.0313)  & 6.7599 (1.5737)   \\ \hline
MS-SSIM        & 0.9069 (0.0360)   & 0.7396 (0.0495)  & 0.9115 (0.0357)  & 7.7878 (1.5813)   \\ \hline
ST-RRED      & -0.8840 (0.0406)   & -0.7071 (0.0508)  & 0.9012 (0.1028)  & 8.2752 (2.1837)   \\ \hline
SpEED-QA      & -0.9171 (0.0283)   & -0.7528 (0.0389)  & 0.9070 (0.3196)  & 8.0244 (4.3767)   \\ \hline
ST-GREED      & 0.8573 (0.0556)   & 0.6642 (0.0667)  & 0.8776 (0.0514)  & 8.9718 (1.8265)   \\ \hline
VMAF (v0.6.1)      & 0.9347 (0.0210)   & 0.7773 (0.0328)  & 0.9362 (0.0261)  & 6.6705 (1.3785)    \\ \hline
\textbf{Gaming VMAF}      & \textbf{0.9410 (0.0407)}   & \textbf{0.7913 (0.0544)}  & \textbf{0.9428 (0.0420)}  & \textbf{6.2562 (1.9643)}    \\ \hline
\end{tabular}
\end{table*}

\begin{table*}[]
\centering
\caption{\textcolor{black}{Results of One-Sided T-Test Performed Using the 1000 (Srocc, Plcc) values of the compared FR-VQA Algorithms computed on the LIVE-META MCG Database. Each Cell contains 2 Symbols: the first symbol corresponds to the T-Test done using the Srocc values, and the second corresponds to the T-TEST done using the Plcc values. When a symbol `1' appears, it denotes that the algorithm on the row was statistically superior to that on the column, whereas `0' indicates that the algorithm on the column was statistically superior. A `-' symbol indicates that the column and row algorithms performed equally well} }
\label{tab:frvqattests}
\begin{tabular}{|c|c|c|c|c|c|c|c|c|c|}
\hline
ALGORITHM    & PSNR & SSIM & MS-SSIM & ST-RRED & SpEED-QA & ST-GREED & VMAF (v0.6.1) & Gaming VMAF \\ \hline
PSNR         & (-,-)   & (0,0)      & (0,0)    & (0,0)      & (0,0)      & (0,0)       & (0,0)           & (0,0)        \\ \hline
SSIM      & (1,1)   & (-,-)      & (1,1)    & (1,1)      & (1,1)      & (1,1)       & (0,-)           & (0,0)        \\ \hline
MS-SSIM        & (1,1)   & (0,0)      & (-,-)    & (1,1)      & (0,1)      & (1,1)       & (0,0)           & (0,0)       \\ \hline
ST-RRED      & (1,1)   & (0,0)      & (0,0)    & (-,-)      & (0,0)      & (1,1)       & (0,0)           & (0,0)       \\ \hline
SpEED-QA      & (1,1)   & (0,0)      & (1,0)    & (1,1)      & (-,-)      & (1,1)       & (0,0)           & (0,0)       \\ \hline
ST-GREED     & (1,1)   & (0,0)      & (0,0)    & (0,0)      & (0,0)      & (-,-)       & (0,0)           & (0,0)      \\ \hline
VMAF (v0.6.1) & (1,1)   & (1,-)     & (1,1)    & (1,1)      & (1,1)      & (1,1)       & (-,-)           & (-,0)      \\ \hline
Gaming VMAF         & (1,1)   & (1,1)      & (1,1)    & (1,1)      & (1,1)      & (1,1)       & (-,1)           & (-,-)         \\ \hline
\end{tabular}
\end{table*}

\section{Performance of FR-VQA Algorithms}
\label{sec:proxymos}
\textcolor{black}{In this section, we examine the performances of various Full Reference (FR) VQA models originally developed for natural videos on our proposed database. Our goal is to assess whether they can be utilized as suitable replacements for mean-opinion scores, or serve as pre-training targets when developing deep NR-VQA models for Mobile Cloud Gaming. Deep learning-based algorithms proposed in \cite{DBLP:journals/corr/abs-2101-10955,DBLP:journals/corr/abs-1908-00375, gotz2021konvid, wu2022fast, hvs5m, DBLP:journals/corr/abs-2011-13544} have been successfully used for generic No-Reference Video Quality tasks. Most of these deep learning backbones are pre-trained on one of the large natural image and video classification databases like ImageNet, Imagenet-22K \cite{5206848}, Kinetics-400 \cite{ DBLP:journals/corr/KayCSZHVVGBNSZ17} or benefit from dedicated large databases as in \cite{DBLP:journals/corr/abs-1908-00375}. Developing dedicated deep learning-based models similar to those that involve pre-training on a classification database is complicated in niche VQA sub-domains like Cloud Gaming, due to the absence of large-scale classification datasets comprising rendered gaming content. Furthermore, existing Cloud Gaming VQA databases are too small to support the training of deep learning backbones. To overcome these challenges, researchers working in the Cloud Gaming VQA domain have frequently employed Full Reference VQA algorithms originally developed for generic VQA tasks as substitutes for MOS scores when pre-training complex deep networks for NR-VQA \cite{Zadtootaghaj2018NRGVQMAN,9287080,NDNetgaming}. They achieve this by selecting a popular VQA metric, like VMAF, using it to predict the FR-VQA scores using a pristine gaming video and a synthetically distorted version of the pristine video. The low expense of producing synthetically distorted videos and estimating proxy MOS scores in the form of FR-VQA outputs makes it feasible to create large databases for pre-training deep networks. Once a deep network backbone is pre-trained, most authors \cite{9287080,NDNetgaming} fine-tune the pre-trained backbone with a small amount of human-annotated data to achieve better performance than traditional handcrafted feature-based models on the Cloud Gaming NR-VQA task. It is worth noting that using deep learning backbones pre-trained on natural images and videos may not lead to optimal performance on Cloud Gaming NR-VQA task. This is because the visual content generated by computer graphics, as in Cloud Gaming videos, typically has fewer details and is smoother than naturalistic videos or images, which alters the bandpass statistics of Cloud Gaming videos relative to those of naturalistic videos \cite{gamival}. \\
\indent Cloud Gaming  NR-VQA algorithms \cite{Zadtootaghaj2018NRGVQMAN,9287080,NDNetgaming} usually employ VMAF scores as their pre-training targets. Here, we comprehensively compare the performances of seven FR-VQA algorithms: PSNR, SSIM \cite{wang2004image}, MS-SSIM \cite{wang2003multiscale}, ST-RRED \cite{soundararajan2012video}, SpEED-QA \cite{7979533}, ST-GREED \cite{DBLP:journals/corr/abs-2010-13715}, and VMAF on the LIVE-Meta Mobile Cloud Gaming database to explore for their suitabilities as Proxy-MOS or intermediate pre-training targets for the development of NR-VQA models focused on Mobile Cloud Gaming. We calculated the DMOS using equation \eqref{eq:dmos} and the proxy reference videos in our database were used as reference videos when computing the FR-VQA scores. To ensure consistency, we utilized the same 1000 train-test split used for the NR-VQA algorithms in our evaluation of FR-VQA algorithms. \\
\indent PSNR, SSIM, and MS-SSIM are computed per-frame between the reference and distorted videos, then averaged across all frames. The FR-VQA algorithms PSNR, SSIM, MS-SSIM, ST-RRED, and SpEED-QA algorithms do not require training, and therefore, were directly evaluated on the 1000 test sets. ST-GREED features were obtained from the proxy reference and distorted videos in the training and test sets. The features from the training set and the corresponding DMOS were then used to train an SVR similar to the NR-VQA algorithms. Once the SVR model was obtained, the features from the test set and the corresponding DMOS scores were used to obtain the performance of the overall algorithm. We also present two versions of VMAF: VMAF (v0.6.1), the pre-trained open source version widely used for generic VQA tasks, and our version of VMAF which we call Gaming VMAF, which uses the same features as VMAF (v0.6.1) but with the SVR trained on the LIVE-Meta MCG database using the same evaluation strategy as ST-GREED. Table \ref{tab:frvqa} summarizes the results obtained for all the FR-VQA algorithms. It may be observed that the VMAF models outperformed the other models, while the computationally less expensive SSIM model also demonstrated competitive performance. \textcolor{black}{Similar to the evaluation of NR-VQA algorithms, we used the 1000 SROCC and PLCC scores produced by the FR VQA models to run one-sided t-tests, using the $95\%$ confidence level to determine whether the performance of one FR-VQA algorithm was statistically superior to another. Each entry in Table \ref{tab:frvqattests} consists of two symbols, corresponding to the t-tests conducted using the SROCC and PLCC values. Based on the results, we conclude that when comparing the two VMAF models, the use of SROCC as a performance metric did not show statistically significant differences. However, using PLCC revealed statistically significant differences, with Gaming VMAF exhibiting slightly better performance. It may also be concluded that a statistically significant difference exists between the performances of the Gaming VMAF and SSIM models when evaluated using both performance metrics.} \\
\indent The high correlations obtained on the VMAF models suggest that the VMAF models could be reasonably used as proxy-MOS scores or as pre-training targets for MCG NR-VQA models. By pre-training a deep learning model on VMAF scores, a model could potentially learn to extract useful ``gaming quality-aware" features on a small human-annotated database like ours, potentially improving performance on the MCG NR-VQA task. However, it is important to note that while pre-training can be beneficial, it may not always result in improved performance. Therefore, it is crucial to exercise caution when selecting a pre-training dataset, the synthetic distortions applied, and the proxy FR-VQA algorithm to ensure that pre-training boosts the performance of the target MCG NR-VQA task. Furthermore, relying on pre-training using a single FR-VQA model presents the potential danger of NR-VQA models adopting the strengths and limitations of that FR-VQA model, leading to reduced NR-VQA generalization. One possible solution would be to convert the pre-training to a Multi-Task Learning problem \cite{DBLP:journals/corr/abs-2009-09796}, using multiple FR-VQA algorithms as different tasks. For example, in case of Mobile Cloud Gaming, a combination of VMAF, SSIM and SpEED-QA could be used as multiple tasks to pre-train the deep network backbone. This approach could enable more generalized ``quality-aware" representations, which might further enhance performance on the MCG NR-VQA task.}

\section{Conclusion and Future Work}
\label{sec:conclusion}
In this work, we have introduced a new psychometric database that we call the LIVE-Meta Mobile Cloud Gaming (LIVE-Meta MCG) video quality database. It is our hope that this resource helps advance the development of No Reference VQA algorithms directed towards Mobile Cloud Gaming. The new database will be made publicly available to the research community at \href{https://live.ece.utexas.edu/research/LIVE-Meta-Mobile-Cloud-Gaming/index.html}{https://live.ece.utexas.edu/research/LIVE-Meta-Mobile-Cloud-Gaming/index.html}. We have also demonstrated the usability of the database for comparing, benchmarking and designing NR VQA algorithms. As a next step, algorithms based on traditional natural scene statistics (NSS) models and/or deep-learning methods could be developed to further improve the accuracy of NR-VQA algorithms. In addition, since cloud gaming applications require real-time video quality prediction capability, it is also of utmost interest to develop algorithms capable of running at least in real-time.\\
\indent \textcolor{black}{We also demonstrated that tighter 95\% confidence intervals were obtained on the MLE-MOS estimates of landscape videos than those of portrait videos. A possible research direction could be to explore this dichotomy in further detail.} Future work could also focus on development of ``Quality of Experience" (QoE) databases comprised of subjective QoE responses to various designs dimensions such as changing bitrates, content-adaptive encoding, network conditions and video content which would further help in the development of perceptually-optimized cloud video streaming strategies, leading to improved mobile cloud gaming experiences.

\section{Appendix}
\label{sec:appendix}

\subsection{Gaming Video Contents in LIVE-META Mobile Cloud Gaming Database}
\label{appendix:gamenames}
\begin{figure*}[htbp]
    \centering
    \subfloat[GamingVideoSET : Contrast vs Brightness]{{\includegraphics[width=0.32\textwidth]{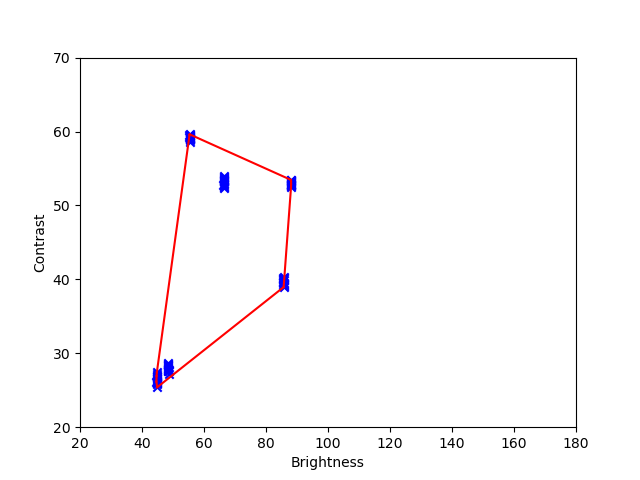}}}
    \hfill
    \subfloat[GamingVideoSET : Sharpness vs Colourfulness]{{\includegraphics[width=0.32\textwidth]{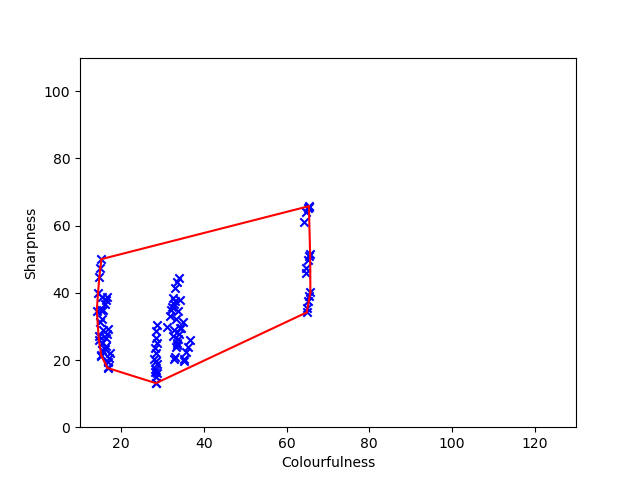}}}
    \hfill
    \subfloat[GamingVideoSET : TI vs SI]{{\includegraphics[width=0.32\textwidth]{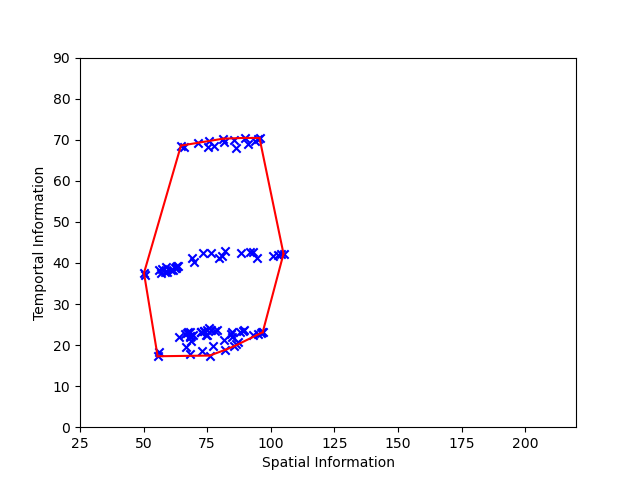}}}

    \subfloat[KUGVD : Contrast vs Brightness]{{\includegraphics[width=0.32\textwidth]{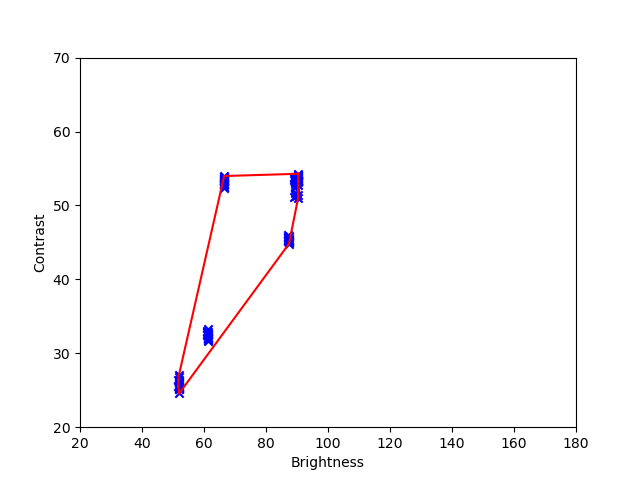}}}
    \hfill
    \subfloat[KUGVD : Sharpness vs Colourfulness]{{\includegraphics[width=0.32\textwidth]{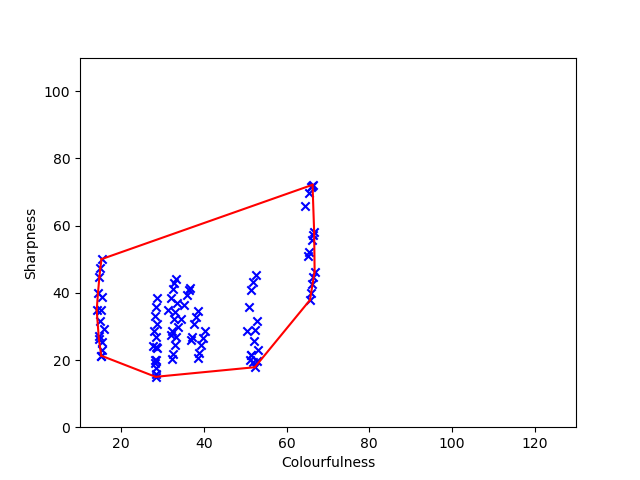}}}
    \hfill
    \subfloat[KUGVD : TI vs SI]{{\includegraphics[width=0.32\textwidth]{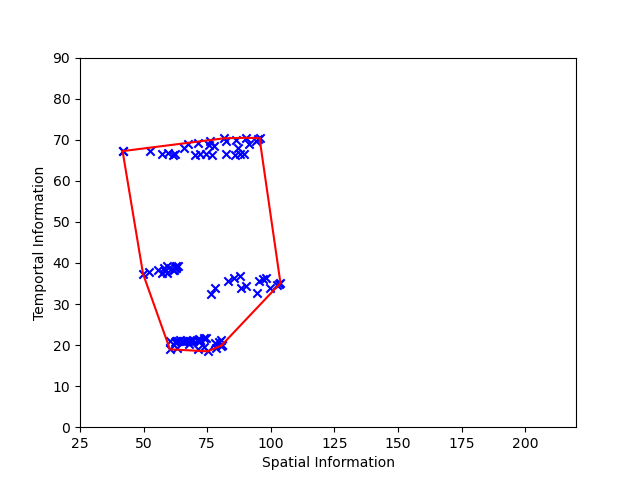}}}

    \subfloat[CGVDS : Contrast vs Brightness]{{\includegraphics[width=0.32\textwidth]{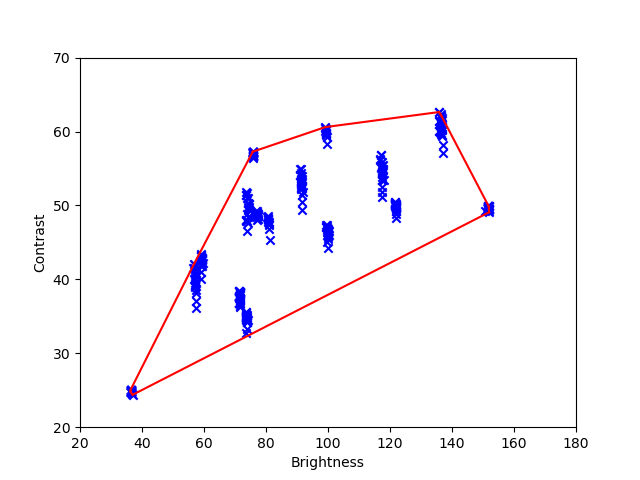}}}
    \hfill
    \subfloat[CGVDS : Sharpness vs Colourfulness]{{\includegraphics[width=0.32\textwidth]{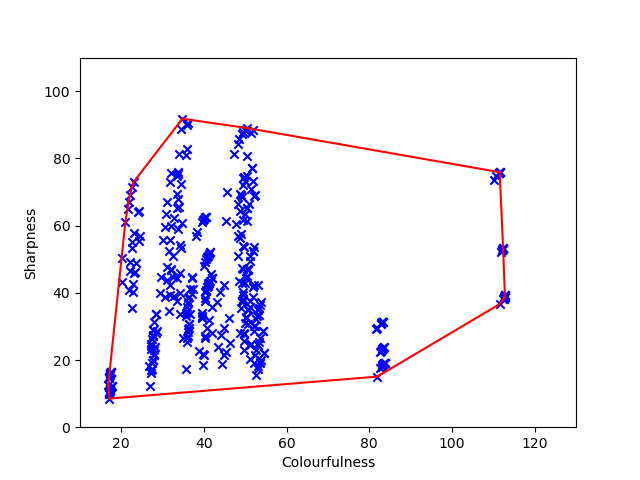}}}
    \hfill
    \subfloat[CGVDS : TI vs SI]{{\includegraphics[width=0.32\textwidth]{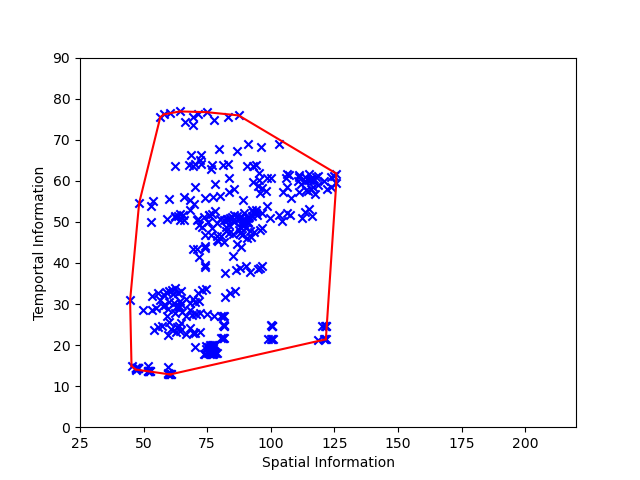}}}

    \subfloat[LIVE-Meta MCG : Contrast vs Brightness]{{\includegraphics[width=0.32\textwidth]{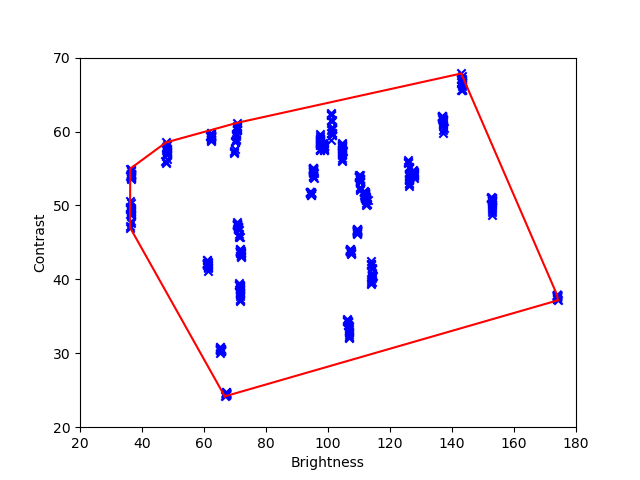}}}
    \hfill
    \subfloat[LIVE-Meta MCG : Sharpness vs Colourfulness]{{\includegraphics[width=0.32\textwidth]{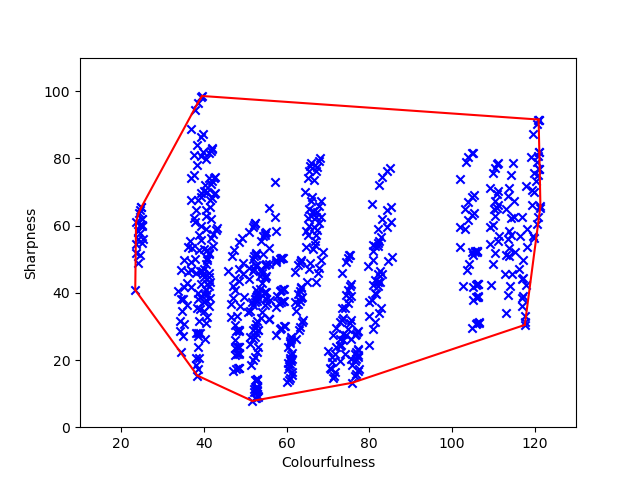}}}
    \hfill
    \subfloat[LIVE-Meta MCG : TI vs SI]{{\includegraphics[width=0.32\textwidth]{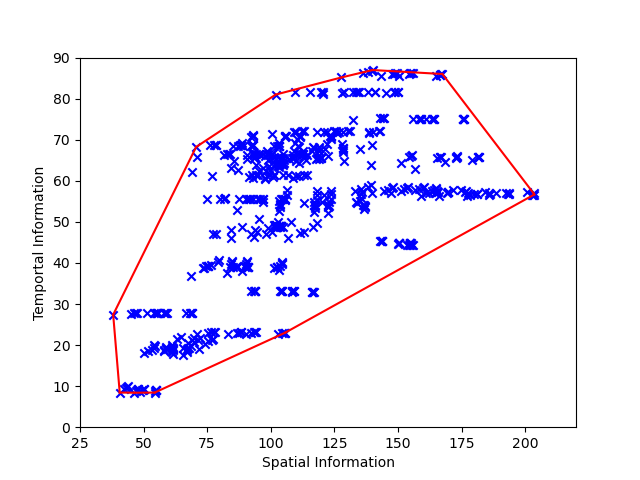}}}
    
    \captionsetup{justification=centering}
    \caption{Source content (blue ‘x’) distribution in paired feature space with
corresponding convex hulls (red boundaries). Left column: Contrast x Brightness,
middle column: Sharpness x Colourfulness, right column: Temporal Information (TI) vs Spatial Information (SI) across four Cloud Gaming Databases.}%
    
    \label{fig:objfeats}
\end{figure*}

Table \ref{tab:game_name} lists the games  present in the dataset along with their original resolutions as rendered by the Cloud Game engine. \textcolor{black}{Fig. \ref{fig:objfeats} compares the coverage of a number of objective features, including contrast, brightness, sharpness, colorfulness, spatial information, and temporal information of the videos in our database against the same features computed from other existing Cloud Gaming databases. The content distribution in the paired feature space shows that the coverage of our proposed database is significantly better than all the other three existing cloud gaming databases. }
\begin{table}[]
\caption{Details of Games Present in the Proposed LIVE-META Mobile Cloud Gaming (LIVE-META MCG) Database}
\centering
\begin{tabular}{|c|c|c|}
\hline
Cloud Games              & Original Resolution & Display Orientation \\ \hline
Asphalt                  & 1664 x 720          & Landscape           \\ \hline
Bejwelled                & 720 x 1280          & Portrait            \\ \hline
Bowling Club             & 720 x 1440          & Portrait            \\ \hline
Design Island            & 1664 x 720          & Landscape           \\ \hline
Dirt Bike                & 720 x 1440          & Portrait            \\ \hline
Dragon Mania Legends     & 1440 x 720          & Landscape           \\ \hline
Hungry Dragon            & 1512 x 720          & Landscape           \\ \hline
Mobile Legends Adventure & 1440 x 720          & Landscape           \\ \hline
Monument Valley 2        & 720 x 1280          & Portrait            \\ \hline
Mystery Manor            & 1728 x 720          & Landscape           \\ \hline
PGA Golf Tour            & 720 x 1280          & Portrait            \\ \hline
Plants vs Zombies        & 1280 x 720          & Landscape           \\ \hline
Solitaire                & 1664 x 720          & Landscape           \\ \hline
Sonic                    & 720 x 1280          & Portrait            \\ \hline
State of Survival        & 1664 x 720          & Landscape           \\ \hline
WWE                      & 720 x 1440          & Portrait            \\ \hline
\end{tabular}
\label{tab:game_name}
\end{table}

\begin{figure*}[htbp]
    \centering
    \subfloat[Game Video Playback]{{\includegraphics[width=0.32\textwidth]{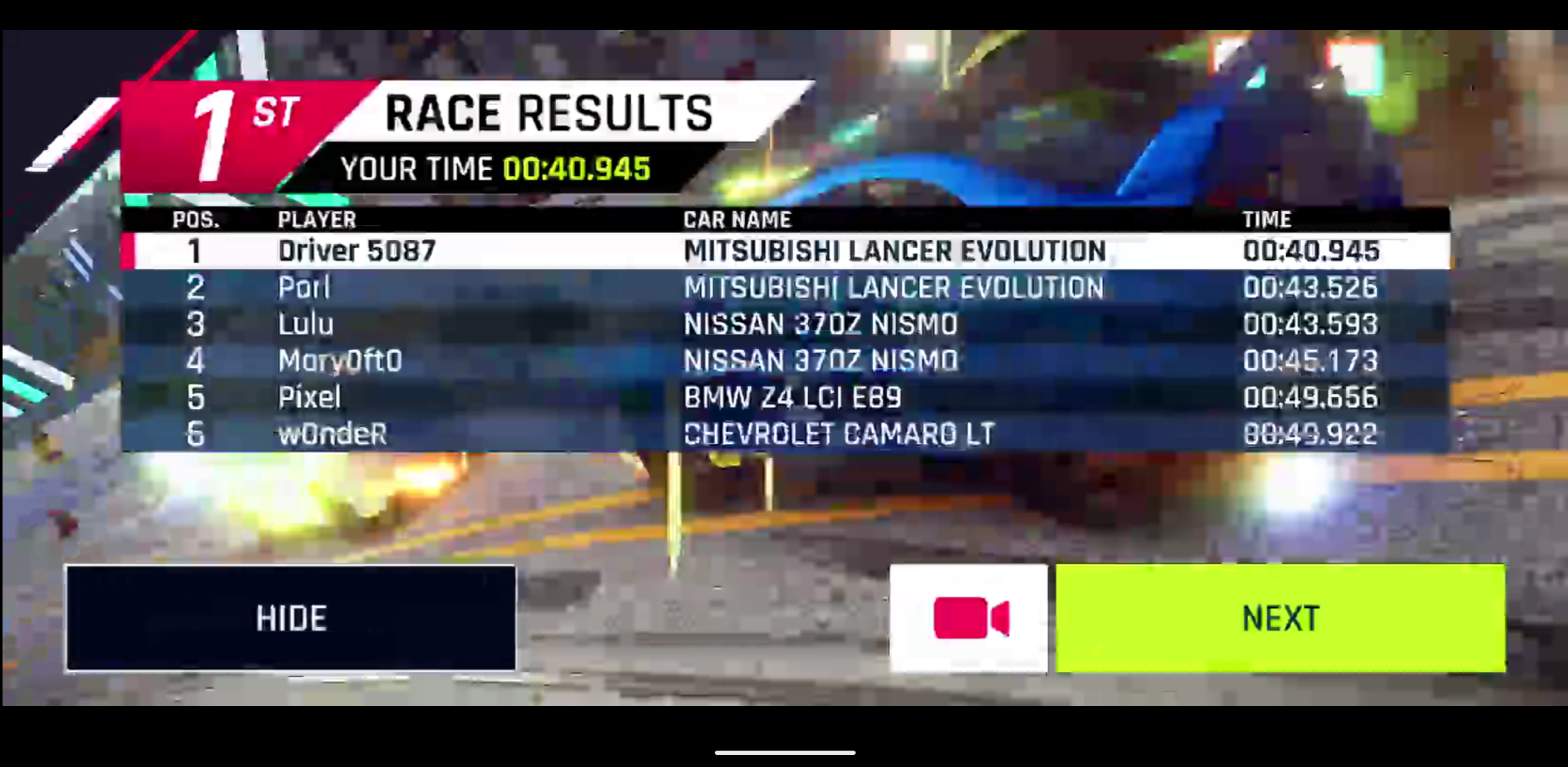}}}
    \hfill
    \subfloat[Initial State of Rating bar]{{\includegraphics[width=0.32\textwidth]{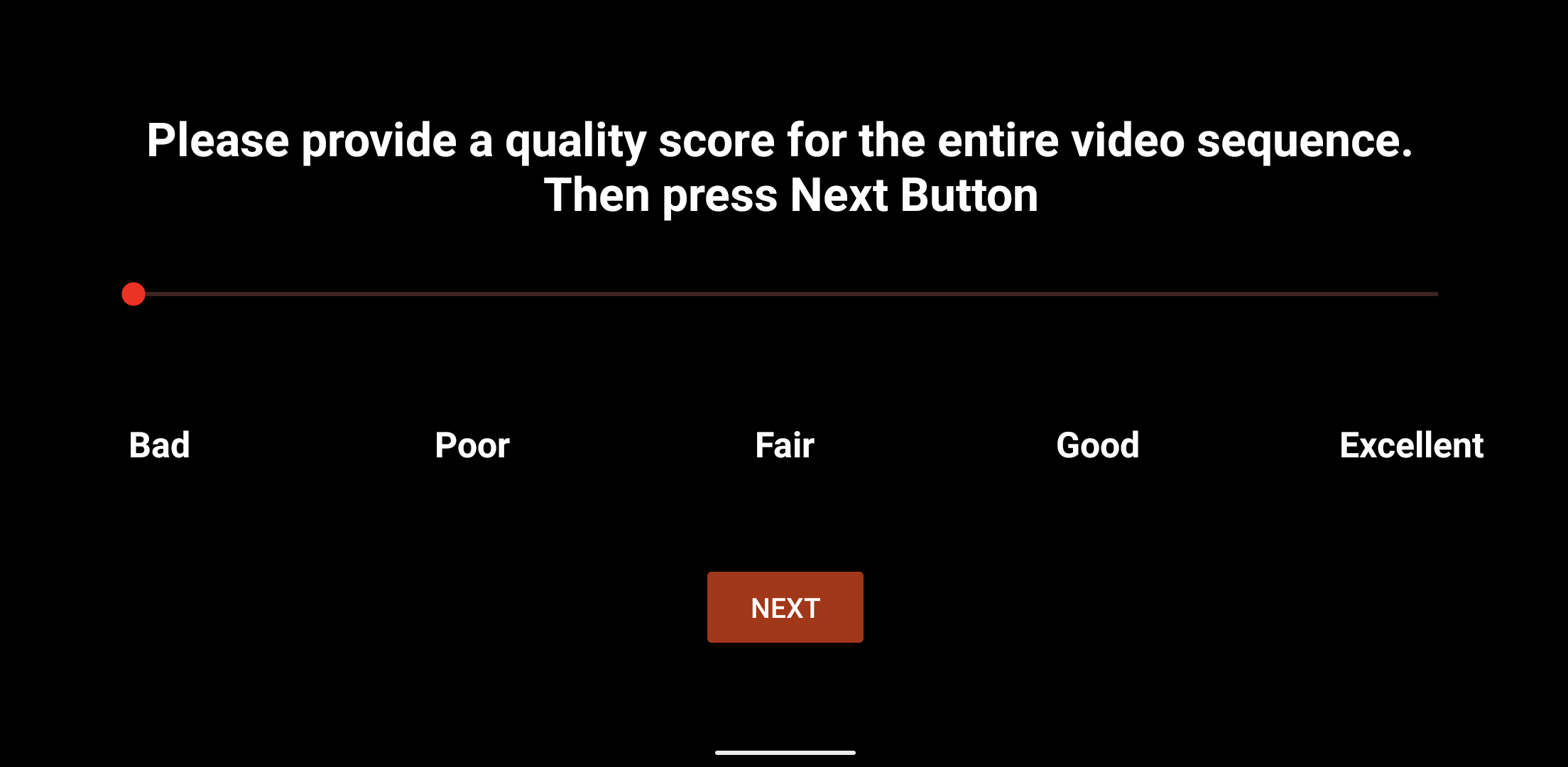}}}
    \hfill
    \subfloat[Final State of the Rating bar]{{\includegraphics[width=0.32\textwidth]{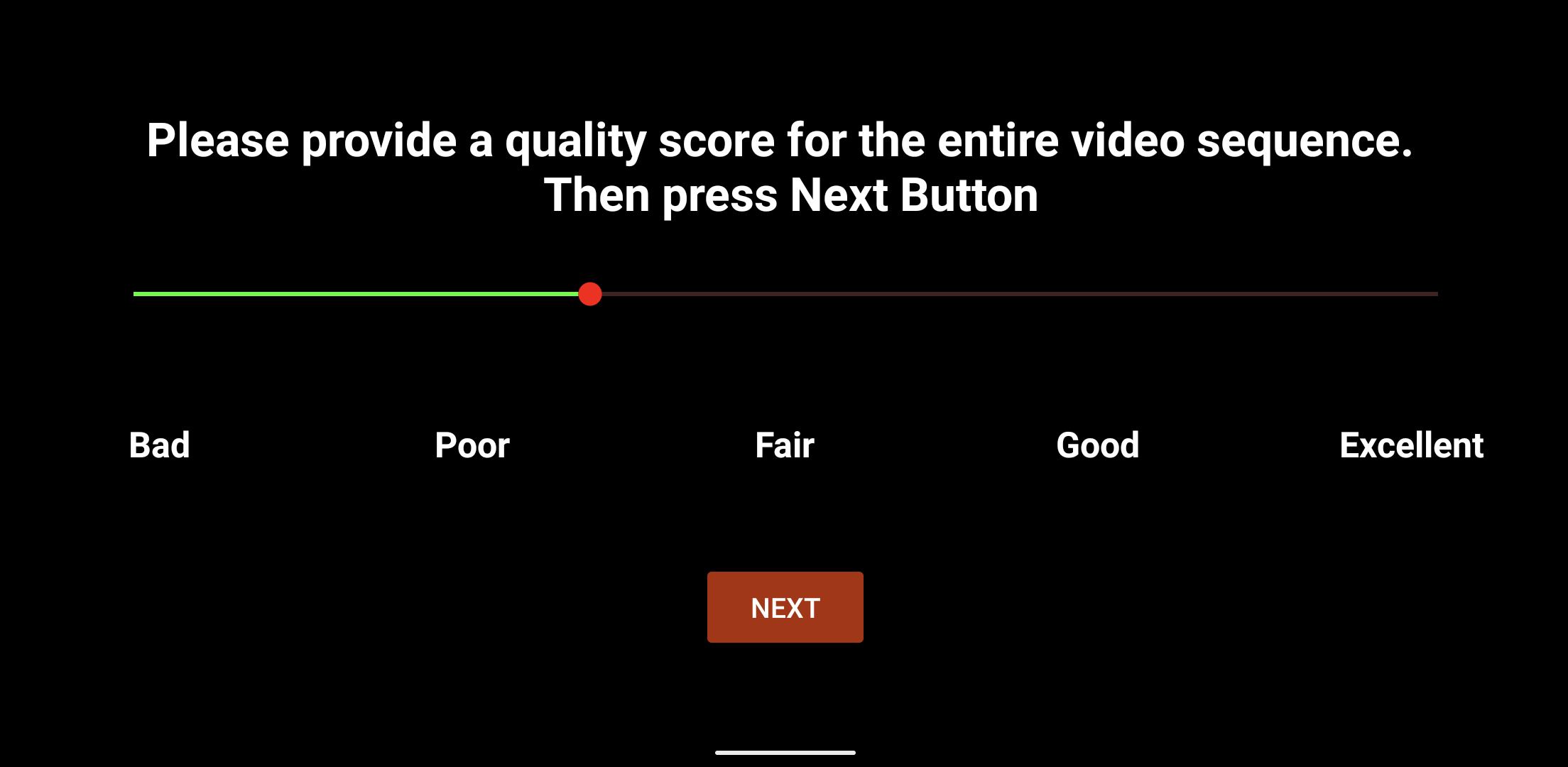}}}

    \caption{Video Quality Rating process in our custom-developed Android Application. Left column: A game video playback of duration 20 seconds, Middle Column: Initial state of the rating bar initialized to extreme left, Right Column: Exemplar final state of the rating bar when the user records their final score.}%
    \label{fig:appscreen}
\end{figure*}

\subsection{Android Application}
\label{appendix:android}
We used a custom developed Android Application to conduct the in-lab subjective study for the development of the LIVE-Meta MCG database. The code will be made publicly available at \url{https://github.com/avinabsaha/LIVE-Meta-MCG-SubjectiveStudySetup}. Fig. \ref{fig:appscreen} demonstrates the steps involved in the video quality rating process in the Android application.

\subsection{Additional Post Study Questionnaire \& Demographics}
\label{appendix:poststudy}
As a part of the post-study questionnaire, we also asked the human subjects about the distribution of videos, the difficulty of rating the videos, and whether they experienced any sort of dizziness or uneasiness while viewing and rating the videos. In the end, in $74.3\%$ $(107/144)$ sessions, the subjects felt that the distribution of quality was uniform with an equal number of good, intermediate and bad quality videos. In the other sessions, the subjects felt that the majority of the videos were either of very good or very bad quality, and few, if any of the videos were of intermediate quality. On a scale from $0$ to $100$, we asked the subjects to rate the difficulty of judging the perceptual quality of the video after each session, with $0$ being very difficult and $100$ being reasonably easy to judge. All of the subjects were able to provide subjective quality ratings without much difficulty, as reflected by the mean and median scores of difficulty, which were $72.1$ and $77.5$, respectively. The human subjects reported that they felt slight dizziness or uneasiness in approximately $11\%$ of the sessions, however the percentage of dizziness or uneasiness inducing videos was much lower. More detailed results from the survey regarding dizziness and uneasiness can be found in  Table \ref{tab:dizz}.\\
\indent The demographic data of age and gender were collected only at the end of the first session. The mean, median, and standard deviation of the ages of the participants were found to be 23.57, 23.0, and 3.04. We summarize the gender distribution among the participants in Table \ref{tab:gender}.

\begin{table}[h]
\centering
\caption{Opinions of Study Participants regarding the percentage of Gaming Videos that induced Dizziness/Uneasiness}
\label{tab:dizz}
\begin{tabular}{|c|c|c|c|c|c|}
\hline
\begin{tabular}[c]{@{}c@{}}\% of Gaming \\ videos inducing\\ dizziness/\\ uneasiness\end{tabular} &
  None &
  \textless{}10\% &
  10-20\% &
  20-40\% &
  \textgreater{}40\% \\ \hline
\# of sessions &
  \begin{tabular}[c]{@{}c@{}}128 \\ (88.89\%)\end{tabular} &
  \begin{tabular}[c]{@{}c@{}}6\\ (4.16\%)\end{tabular} &
  \begin{tabular}[c]{@{}c@{}}7\\ (4.86\%)\end{tabular} &
  \begin{tabular}[c]{@{}c@{}}3\\ (2.08\%)\end{tabular} &
  \begin{tabular}[c]{@{}c@{}}0\\ (0\%)\end{tabular} \\ \hline
\end{tabular}
\end{table}

\begin{table}[]
\centering
\caption{Demographics of Human Study Participants based on Gender}
\label{tab:gender}
\begin{tabular}{|c|c|c|c|c|}
\hline
Gender    & Male        & Female      & Others    & Prefer Not to Say \\ \hline
Count(\%) & 58(80.55\%) & 11(15.27\%) & 2(2.72\%) & 1(1.36\%)         \\ \hline
\end{tabular}%

\end{table}

\subsection{Group-wise Inter-Subject and Intra-Subject Consistency}
\label{appendix:intersubject}
We report the inter-subject and intra-subject consistency scores for each of the subject groups in Table \ref{tab:consistency} using the methodology described in Section IV-G of the main paper. Across subject groups, the SROCC scores for inter-subject consistency ranged from 0.900 to 0.936 with an average of 0.912, while PLCC scores ranged from 0.915 to 0.949 with an average of 0.929. The SROCC scores for intra-subject consistency ranged from 0.827 to 0.866 with an average of 0.848, while PLCC scores ranged from 0.844 to 0.870 with an average of 0.860. These scores reflect the consistency of our data acquisition process across all the subject groups.
\begin{table}[]                                                                                                                                                                                                 
\centering
\caption{Subject Consistency }
\label{tab:consistency}
\begin{tabular}{|c|cc|cc|}
\hline
\multicolumn{1}{|l|}{} & \multicolumn{2}{l|}{Inter-Subject Consistency} & \multicolumn{2}{l|}{Intra-Subject Consistency} \\ \hline
Subject Group          & \multicolumn{1}{c|}{SROCC}        & PLCC       & \multicolumn{1}{c|}{SROCC}        & PLCC       \\ \hline
1 & \multicolumn{1}{c|}{0.901} & 0.915 & \multicolumn{1}{c|}{0.850} & 0.870 \\ \hline
2 & \multicolumn{1}{c|}{0.900} & 0.917 & \multicolumn{1}{c|}{0.840} & 0.854 \\ \hline
3 & \multicolumn{1}{c|}{0.905} & 0.920 & \multicolumn{1}{c|}{0.849} & 0.870 \\ \hline
4 & \multicolumn{1}{c|}{0.913} & 0.941 & \multicolumn{1}{c|}{0.827} & 0.844 \\ \hline
5 & \multicolumn{1}{c|}{0.916} & 0.933 & \multicolumn{1}{c|}{0.866} & 0.859 \\ \hline
6 & \multicolumn{1}{c|}{0.936} & 0.949 & \multicolumn{1}{c|}{0.854} & 0.865 \\ \hline
\end{tabular}
\end{table}

\subsection{Additional Analysis and Visualization of Opinion Scores}
\label{appendix:additionalanalysis}

\begin{figure}[h!]
  \centering
  \centerline{\includegraphics[width=6.5cm]{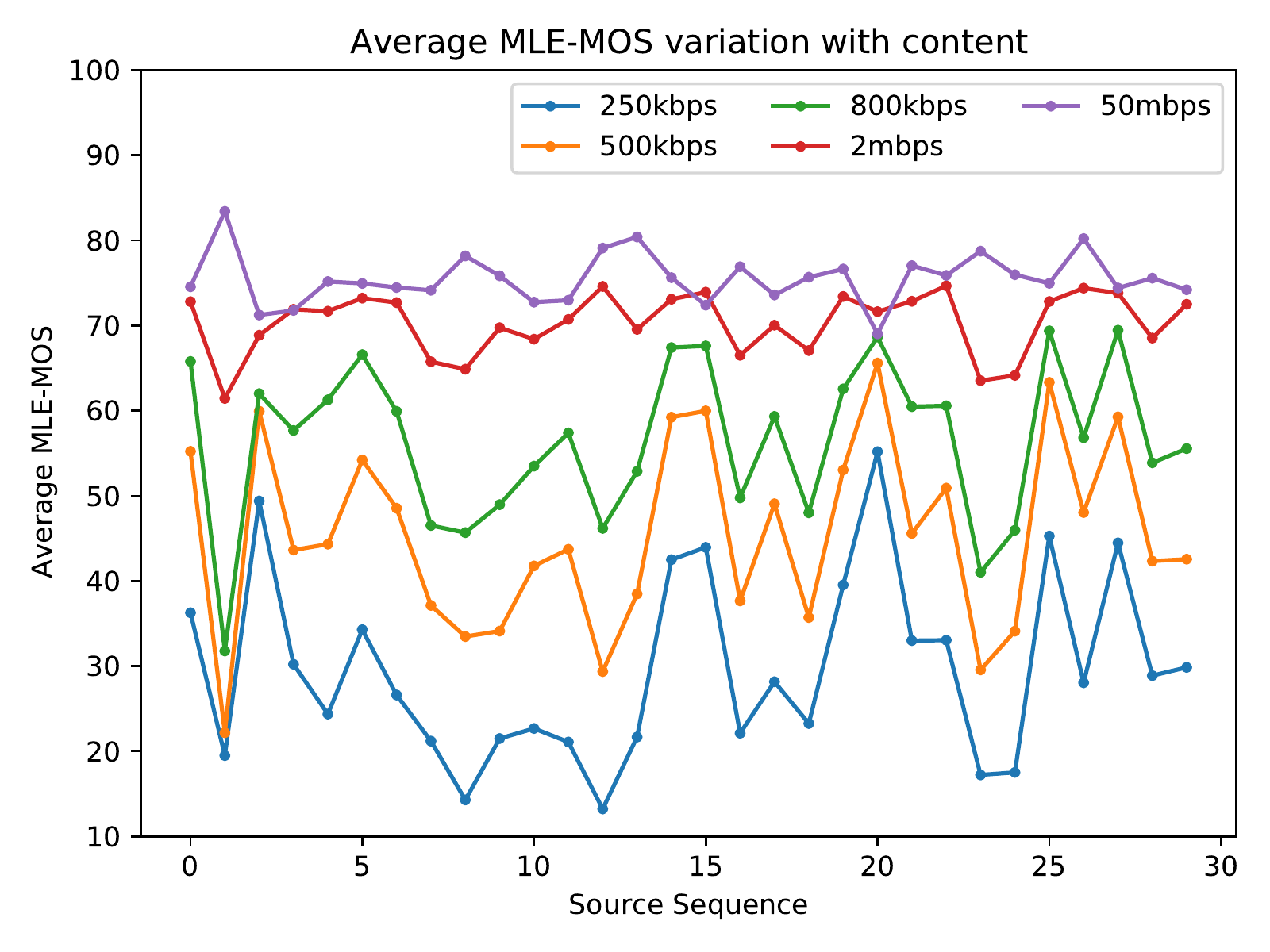}}
  \caption{Variation of average MLE-MOS against content for five fixed bitrates.} \label{fig:content}
\end{figure}

\begin{figure}[h!]
  \centering
  \centerline{\includegraphics[width=6.5cm]{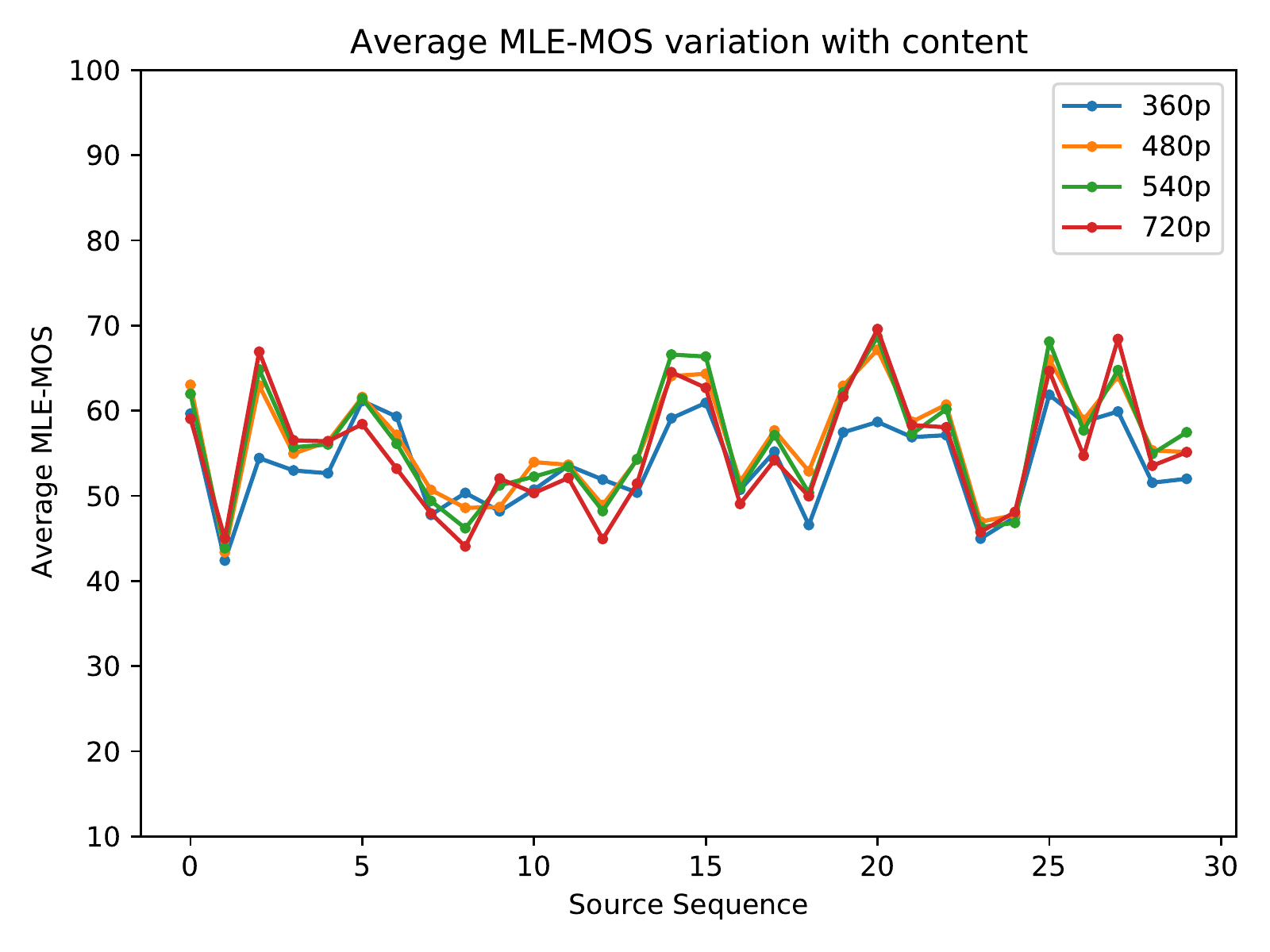}}
  \caption{Variation of average MLE-MOS against content for four fixed resolutions.} \label{fig:content1}
\end{figure}

Fig. \ref{fig:content}, examines the interplay of source video content and bitrate and how these together affect MLE-MOS. To obtain the plot, we separately calculated the average MLE-MOS ratings of each of the 30 source sequences on a per-bitrate basis across all available resolutions. Fig. \ref{fig:content} shows a clear separations between the MLE-MOS curves of all the contents, except at very high bitrates. Across contents, however, the curves are commingled, which is a good illustration of the difficulty of the VQA problem (it is not just about bitrate). The variation of MLE-MOS for all contents was greatly reduced at bitrates of $2$ mbps or higher as compared to lower bitrates. Clearly, as shown in prior studies, the effect of video compression induced distortions on perceptual video quality is highly content-dependent because of perceptual masking and similar processes. \\
\indent Fig. \ref{fig:content1} shows the effects of video source content on MLE-MOS, across all bitrates for each of the fixed four resolutions. Specifically, we plotted the average MLE-MOS scores of the encoded videos over the five different bitrates associated with each resolution in the database. As may be observed, there was no strong separation between the MLE-MOS curves, although the content did cause notable differences in the reported video qualities. A salient takeaway from these two analyses is that video compression has a heavier impact on the visual perception of video quality than does resizing, at least on gaming videos. This further suggests the efficacy of resizing to achieve data efficiencies with little perceptual loss in the context of mobile gaming video streaming.

\section*{Acknowledgment}

The authors would thank all the volunteers who took
part in the human study. The authors also acknowledge the Texas Advanced Computing Center (TACC), at the University of Texas at Austin for providing HPC, visualization, database, and grid resources that have contributed to the research results reported in this paper. URL: \href{http://www.tacc.utexas.edu}{http://www.tacc.utexas.edu}

\section*{Change Log}
\begin{itemize}
    \item v1 Uploaded to Arxiv on 26th May, 2023.
\end{itemize}

\ifCLASSOPTIONcaptionsoff
  \newpage
\fi



\bibliographystyle{IEEEtran}
\bibliography{refs}

\begin{thebibliography}{10}
\providecommand{\url}[1]{#1}
\csname url@samestyle\endcsname
\providecommand{\newblock}{\relax}
\providecommand{\bibinfo}[2]{#2}
\providecommand{\BIBentrySTDinterwordspacing}{\spaceskip=0pt\relax}
\providecommand{\BIBentryALTinterwordstretchfactor}{4}
\providecommand{\BIBentryALTinterwordspacing}{\spaceskip=\fontdimen2\font plus
\BIBentryALTinterwordstretchfactor\fontdimen3\font minus
  \fontdimen4\font\relax}
\providecommand{\BIBforeignlanguage}[2]{{%
\expandafter\ifx\csname l@#1\endcsname\relax
\typeout{** WARNING: IEEEtran.bst: No hyphenation pattern has been}%
\typeout{** loaded for the language `#1'. Using the pattern for}%
\typeout{** the default language instead.}%
\else
\language=\csname l@#1\endcsname
\fi
#2}}
\providecommand{\BIBdecl}{\relax}
\BIBdecl

\bibitem{Allied}
``{Cloud Gaming Market by offering (Infrastructure and Gaming Platform
  Service), Device Type (Smartphones, Tablets, Gaming Consoles, PCs \& Laptops,
  Smart TVs, and HMDs), and Solution (File streaming and video streaming):
  Global Opportunity Analysis and Industry Forecast, 2021–2030.}''
  \url{https://www.alliedmarketresearch.com/cloud-gaming-market-A07461}, 2021,
  [Online; accessed 30-January-2022].

\bibitem{Barman2018GamingVideoSETAD}
N.~Barman, S.~Zadtootaghaj, S.~Schmidt, M.~G. Martini, and S.~M{\"o}ller,
  ``Gamingvideoset: A dataset for gaming video streaming applications,''
  \emph{2018 16th Annual Workshop on Network and Systems Support for Games
  (NetGames)}, pp. 1--6, 2018.

\bibitem{8727887}
N.~Barman, E.~Jammeh, S.~A. Ghorashi, and M.~G. Martini, ``No-reference video
  quality estimation based on machine learning for passive gaming video
  streaming applications,'' \emph{IEEE Access}, vol.~7, pp. 74\,511--74\,527,
  2019.

\bibitem{10.1145/3339825.3391872}
\BIBentryALTinterwordspacing
S.~Zadtootaghaj, S.~Schmidt, S.~S. Sabet, S.~M\"{o}ller, and C.~Griwodz,
  ``Quality estimation models for gaming video streaming services using
  perceptual video quality dimensions,'' in \emph{Proceedings of the 11th ACM
  Multimedia Systems Conference}, ser. MMSys '20.\hskip 1em plus 0.5em minus
  0.4em\relax New York, NY, USA: Association for Computing Machinery, 2020, p.
  213–224. [Online]. Available: \url{https://doi.org/10.1145/3339825.3391872}
\BIBentrySTDinterwordspacing

\bibitem{Wen2021SubjectiveAO}
S.~Wen, S.~Ling, J.~Wang, X.~Chen, L.~Fang, Y.~Jing, and P.~L. Callet,
  ``Subjective and objective quality assessment of mobile gaming video,''
  \emph{ArXiv}, vol. abs/2103.05099, 2021.

\bibitem{yu2022subjective}
X.~Yu, Z.~Tu, Z.~Ying, A.~C. Bovik, N.~Birkbeck, Y.~Wang, and B.~Adsumilli,
  ``Subjective quality assessment of user-generated content gaming videos,'' in
  \emph{Proceedings of the IEEE/CVF Winter Conference on Applications of
  Computer Vision}, 2022, pp. 74--83.

\bibitem{Zadtootaghaj2018NRGVQMAN}
S.~Zadtootaghaj, N.~Barman, S.~Schmidt, M.~G. Martini, and S.~M{\"o}ller,
  ``Nr-gvqm: A no reference gaming video quality metric,'' \emph{2018 IEEE
  International Symposium on Multimedia (ISM)}, pp. 131--134, 2018.

\bibitem{vmaf}
Z.~Li, A.~Aaron, I.~Katsavounidis, A.~Moorthy, and M.~Manohara, ``Toward a
  practical perceptual video quality metric,'' vol.~6, 2016, p.~2.

\bibitem{inproceedingsnofu}
S.~Göring, R.~R. Ramachandra~Rao, and A.~Raake, ``nofu -a lightweight
  no-reference pixel based video quality model for gaming content,'' 06 2019.

\bibitem{NDNetgaming}
M.~Utke, S.~Zadtootaghaj, S.~Schmidt, S.~Bosse, and S.~Moeller, ``{NDNetGaming
  - Development of a No-Reference Deep CNN for Gaming Video Quality
  Prediction},'' in \emph{Multimedia Tools and Applications}.\hskip 1em plus
  0.5em minus 0.4em\relax Springer, 2020.

\bibitem{9287080}
S.~Zadtootaghaj, N.~Barman, R.~R.~R. Rao, S.~Göring, M.~G. Martini, A.~Raake,
  and S.~Möller, ``Demi: Deep video quality estimation model using perceptual
  video quality dimensions,'' in \emph{2020 IEEE 22nd International Workshop on
  Multimedia Signal Processing (MMSP)}, 2020, pp. 1--6.

\bibitem{gamival}
Y.-C. Chen, A.~Saha, C.~Davis, B.~Qiu, X.~Wang, R.~Gowda, I.~Katsavounidis, and
  A.~C. Bovik, ``Gamival: Video quality prediction on mobile cloud gaming
  content,'' \emph{IEEE Signal Processing Letters}, vol.~30, pp. 324--328,
  2023.

\bibitem{DBLP:journals/corr/HuangLW16a}
\BIBentryALTinterwordspacing
G.~Huang, Z.~Liu, and K.~Q. Weinberger, ``Densely connected convolutional
  networks,'' \emph{CoRR}, vol. abs/1608.06993, 2016. [Online]. Available:
  \url{http://arxiv.org/abs/1608.06993}
\BIBentrySTDinterwordspacing

\bibitem{recogaming}
\emph{Opinion model predicting gaming quality of experience for cloud gaming
  services}, document ITU-T recommendation G.1072, 2020.

\bibitem{8463417}
S.~Schmidt, S.~Möller, and S.~Zadtootaghaj, ``A comparison of interactive and
  passive quality assessment for gaming research,'' in \emph{2018 Tenth
  International Conference on Quality of Multimedia Experience (QoMEX)}, 2018,
  pp. 1--6.

\bibitem{8093636}
D.~Ghadiyaram, J.~Pan, and A.~C. Bovik, ``A subjective and objective study of
  stalling events in mobile streaming videos,'' \emph{IEEE Transactions on
  Circuits and Systems for Video Technology}, vol.~29, no.~1, pp. 183--197,
  2019.

\bibitem{7987076}
C.~G. Bampis, Z.~Li, A.~K. Moorthy, I.~Katsavounidis, A.~Aaron, and A.~C.
  Bovik, ``Study of temporal effects on subjective video quality of
  experience,'' \emph{IEEE Transactions on Image Processing}, vol.~26, no.~11,
  pp. 5217--5231, 2017.

\bibitem{https://doi.org/10.48550/arxiv.1808.03898}
\BIBentryALTinterwordspacing
C.~G. Bampis, Z.~Li, I.~Katsavounidis, T.-Y. Huang, C.~Ekanadham, and A.~C.
  Bovik, ``Towards perceptually optimized end-to-end adaptive video
  streaming,'' 2018. [Online]. Available:
  \url{https://arxiv.org/abs/1808.03898}
\BIBentrySTDinterwordspacing

\bibitem{itut}
\emph{Subjective evaluation methods for gaming quality}, document ITU-T
  Recommendation P.809, 2018.

\bibitem{10.1117/12.477378}
\BIBentryALTinterwordspacing
D.~Hasler and S.~E. Suesstrunk, ``{Measuring colorfulness in natural images},''
  in \emph{Human Vision and Electronic Imaging VIII}, B.~E. Rogowitz and T.~N.
  Pappas, Eds., vol. 5007, International Society for Optics and
  Photonics.\hskip 1em plus 0.5em minus 0.4em\relax SPIE, 2003, pp. 87 -- 95.
  [Online]. Available: \url{https://doi.org/10.1117/12.477378}
\BIBentrySTDinterwordspacing

\bibitem{6280595}
S.~Winkler, ``Analysis of public image and video databases for quality
  assessment,'' \emph{IEEE Journal of Selected Topics in Signal Processing},
  vol.~6, no.~6, pp. 616--625, 2012.

\bibitem{itutww}
\emph{Subjective video quality assessment methods for multimedia applications},
  document ITU-T recommendation P.910, 2008.

\bibitem{Nvenc}
``{NVENC Video Encoder API Programming Guide},''
  \url{https://docs.nvidia.com/video-technologies/video-codec-sdk/nvenc-video-encoder-api-prog-guide/},
  2021, [Online; accessed 30-January-2022].

\bibitem{xda}
``{Google Pixel 5 Display Review: Worthy of a Flagship},''
  \url{https://www.xda-developers.com/google-pixel-5-display-review/#color_accuracy},
  2021, [Online; accessed 19-February-2023].

\bibitem{itutpic}
\emph{Methodology for the Subjective Assessment of the Quality of Television
  Pictures}, document ITU-R Recommendation BT. 500-13, 2012.

\bibitem{liveprac}
``{Visual Screening, Laboratory of Image and Video Engineering},''
  \url{https://live.ece.utexas.edu/research/Quality/visualScreening.htm},
  [Online; accessed 30-January-2022].

\bibitem{DBLP:journals/corr/LiB16c}
\BIBentryALTinterwordspacing
Z.~Li and C.~G. Bampis, ``Recover subjective quality scores from noisy
  measurements,'' \emph{CoRR}, vol. abs/1611.01715, 2016. [Online]. Available:
  \url{http://arxiv.org/abs/1611.01715}
\BIBentrySTDinterwordspacing

\bibitem{article}
T.~Hossfeld, C.~Keimel, M.~Hirth, B.~Gardlo, J.~Habigt, K.~Dieopold, and
  P.~Tran-Gia, ``Best practices for qoe crowdtesting: Qoe assessment with
  crowdsourcing,'' \emph{Multimedia, IEEE Transactions on}, vol.~16, pp.
  541--558, 02 2014.

\bibitem{6353522}
A.~Mittal, R.~Soundararajan, and A.~C. Bovik, ``Making a “completely blind”
  image quality analyzer,'' \emph{IEEE Signal Processing Letters}, vol.~20,
  no.~3, pp. 209--212, 2013.

\bibitem{6272356}
A.~Mittal, A.~K. Moorthy, and A.~C. Bovik, ``No-reference image quality
  assessment in the spatial domain,'' \emph{IEEE Transactions on Image
  Processing}, vol.~21, no.~12, pp. 4695--4708, 2012.

\bibitem{8742797}
J.~Korhonen, ``Two-level approach for no-reference consumer video quality
  assessment,'' \emph{IEEE Transactions on Image Processing}, vol.~28, no.~12,
  pp. 5923--5938, 2019.

\bibitem{9405420}
Z.~Tu, Y.~Wang, N.~Birkbeck, B.~Adsumilli, and A.~C. Bovik, ``Ugc-vqa:
  Benchmarking blind video quality assessment for user generated content,''
  \emph{IEEE Transactions on Image Processing}, vol.~30, pp. 4449--4464, 2021.

\bibitem{DBLP:journals/corr/abs-2101-10955}
\BIBentryALTinterwordspacing
Z.~Tu, X.~Yu, Y.~Wang, N.~Birkbeck, B.~Adsumilli, and A.~C. Bovik, ``{RAPIQUE:}
  rapid and accurate video quality prediction of user generated content,''
  \emph{CoRR}, vol. abs/2101.10955, 2021. [Online]. Available:
  \url{https://arxiv.org/abs/2101.10955}
\BIBentrySTDinterwordspacing

\bibitem{DBLP:journals/corr/abs-1908-00375}
\BIBentryALTinterwordspacing
D.~Li, T.~Jiang, and M.~Jiang, ``Quality assessment of in-the-wild videos,''
  \emph{CoRR}, vol. abs/1908.00375, 2019. [Online]. Available:
  \url{http://arxiv.org/abs/1908.00375}
\BIBentrySTDinterwordspacing

\bibitem{https://doi.org/10.48550/arxiv.2203.12824}
\BIBentryALTinterwordspacing
X.~Yu, Z.~Ying, N.~Birkbeck, Y.~Wang, B.~Adsumilli, and A.~C. Bovik,
  ``Subjective and objective analysis of streamed gaming videos,'' 2022.
  [Online]. Available: \url{https://arxiv.org/abs/2203.12824}
\BIBentrySTDinterwordspacing

\bibitem{DBLP:journals/corr/HeZRS15}
\BIBentryALTinterwordspacing
K.~He, X.~Zhang, S.~Ren, and J.~Sun, ``Deep residual learning for image
  recognition,'' \emph{CoRR}, vol. abs/1512.03385, 2015. [Online]. Available:
  \url{http://arxiv.org/abs/1512.03385}
\BIBentrySTDinterwordspacing

\bibitem{DBLP:journals/corr/ChoMGBSB14}
\BIBentryALTinterwordspacing
K.~Cho, B.~van Merrienboer, {\c{C}}.~G{\"{u}}l{\c{c}}ehre, F.~Bougares,
  H.~Schwenk, and Y.~Bengio, ``Learning phrase representations using {RNN}
  encoder-decoder for statistical machine translation,'' \emph{CoRR}, vol.
  abs/1406.1078, 2014. [Online]. Available:
  \url{http://arxiv.org/abs/1406.1078}
\BIBentrySTDinterwordspacing

\bibitem{5206848}
J.~Deng, W.~Dong, R.~Socher, L.-J. Li, K.~Li, and L.~Fei-Fei, ``Imagenet: A
  large-scale hierarchical image database,'' in \emph{2009 IEEE Conference on
  Computer Vision and Pattern Recognition}, 2009, pp. 248--255.

\bibitem{5404314}
K.~Seshadrinathan, R.~Soundararajan, A.~C. Bovik, and L.~K. Cormack, ``Study of
  subjective and objective quality assessment of video,'' \emph{IEEE
  Transactions on Image Processing}, vol.~19, no.~6, pp. 1427--1441, 2010.

\bibitem{DBLP:journals/corr/abs-2106-13328}
\BIBentryALTinterwordspacing
Y.~Jin, A.~Patney, R.~Webb, and A.~C. Bovik, ``{FOVQA:} blind foveated video
  quality assessment,'' \emph{CoRR}, vol. abs/2106.13328, 2021. [Online].
  Available: \url{https://arxiv.org/abs/2106.13328}
\BIBentrySTDinterwordspacing

\bibitem{gotz2021konvid}
F.~G{\"o}tz-Hahn, V.~Hosu, H.~Lin, and D.~Saupe, ``Konvid-150k: A dataset for
  no-reference video quality assessment of videos in-the-wild,'' \emph{IEEE
  Access}, vol.~9, pp. 72\,139--72\,160, 2021.

\bibitem{wu2022fast}
H.~Wu, C.~Chen, J.~Hou, L.~Liao, A.~Wang, W.~Sun, Q.~Yan, and W.~Lin,
  ``Fast-vqa: Efficient end-to-end video quality assessment with fragment
  sampling,'' in \emph{Computer Vision--ECCV 2022: 17th European Conference,
  Tel Aviv, Israel, October 23--27, 2022, Proceedings, Part VI}.\hskip 1em plus
  0.5em minus 0.4em\relax Springer, 2022, pp. 538--554.

\bibitem{hvs5m}
\BIBentryALTinterwordspacing
A.-X. Zhang, Y.-G. Wang, W.~Tang, L.~Li, and S.~Kwong, ``Hvs revisited: A
  comprehensive video quality assessment framework,'' 2022. [Online].
  Available: \url{https://arxiv.org/abs/2210.04158}
\BIBentrySTDinterwordspacing

\bibitem{DBLP:journals/corr/abs-2011-13544}
\BIBentryALTinterwordspacing
Z.~Ying, M.~Mandal, D.~Ghadiyaram, and A.~C. Bovik, ``Patch-vq: 'patching up'
  the video quality problem,'' \emph{CoRR}, vol. abs/2011.13544, 2020.
  [Online]. Available: \url{https://arxiv.org/abs/2011.13544}
\BIBentrySTDinterwordspacing

\bibitem{DBLP:journals/corr/KayCSZHVVGBNSZ17}
\BIBentryALTinterwordspacing
W.~Kay, J.~Carreira, K.~Simonyan, B.~Zhang, C.~Hillier, S.~Vijayanarasimhan,
  F.~Viola, T.~Green, T.~Back, P.~Natsev, M.~Suleyman, and A.~Zisserman, ``The
  kinetics human action video dataset,'' \emph{CoRR}, vol. abs/1705.06950,
  2017. [Online]. Available: \url{http://arxiv.org/abs/1705.06950}
\BIBentrySTDinterwordspacing

\bibitem{wang2004image}
Z.~Wang, A.~C. Bovik, H.~R. Sheikh, and E.~P. Simoncelli, ``Image quality
  assessment: from error visibility to structural similarity,'' \emph{IEEE
  transactions on image processing}, vol.~13, no.~4, pp. 600--612, 2004.

\bibitem{wang2003multiscale}
Z.~Wang, E.~P. Simoncelli, and A.~C. Bovik, ``Multiscale structural similarity
  for image quality assessment,'' in \emph{The Thrity-Seventh Asilomar
  Conference on Signals, Systems \& Computers, 2003}, vol.~2.\hskip 1em plus
  0.5em minus 0.4em\relax Ieee, 2003, pp. 1398--1402.

\bibitem{soundararajan2012video}
R.~Soundararajan and A.~C. Bovik, ``Video quality assessment by reduced
  reference spatio-temporal entropic differencing,'' \emph{IEEE Transactions on
  Circuits and Systems for Video Technology}, vol.~23, no.~4, pp. 684--694,
  2012.

\bibitem{7979533}
C.~G. Bampis, P.~Gupta, R.~Soundararajan, and A.~C. Bovik, ``Speed-qa: Spatial
  efficient entropic differencing for image and video quality,'' \emph{IEEE
  Signal Processing Letters}, vol.~24, no.~9, pp. 1333--1337, 2017.

\bibitem{DBLP:journals/corr/abs-2010-13715}
\BIBentryALTinterwordspacing
P.~C. Madhusudana, N.~Birkbeck, Y.~Wang, B.~Adsumilli, and A.~C. Bovik,
  ``{ST-GREED:} space-time generalized entropic differences for frame rate
  dependent video quality prediction,'' \emph{CoRR}, vol. abs/2010.13715, 2020.
  [Online]. Available: \url{https://arxiv.org/abs/2010.13715}
\BIBentrySTDinterwordspacing

\bibitem{DBLP:journals/corr/abs-2009-09796}
\BIBentryALTinterwordspacing
M.~Crawshaw, ``Multi-task learning with deep neural networks: {A} survey,''
  \emph{CoRR}, vol. abs/2009.09796, 2020. [Online]. Available:
  \url{https://arxiv.org/abs/2009.09796}
\BIBentrySTDinterwordspacing

\end{thebibliography}
\end{document}